%% file: main.tex
\documentclass[twoside]{article}
\usepackage[utf8]{inputenc} 
\usepackage[T1]{fontenc}    
\usepackage[british]{babel}
\usepackage{url}            
\usepackage{booktabs}       
\usepackage{amsfonts}       
\usepackage{nicefrac}       
\usepackage{microtype}      
\usepackage{placeins}
\usepackage{csquotes}
\usepackage{tabularx}
\usepackage{subcaption}
\usepackage{wrapfig}
\usepackage[dvipsnames]{xcolor}
\usepackage{hyperref}       
\hypersetup{
    colorlinks=true,
    linkcolor=red,
    filecolor=magenta,      
    urlcolor=PineGreen,
    citecolor=PineGreen,
    anchorcolor=PineGreen,
    }

\usepackage{listings}
\usepackage{mathtools} 
\usepackage{nccmath}
\usepackage{amsmath}
\usepackage{amsthm}
\usepackage{amssymb}
\usepackage{bm}
\usepackage[bb=dsserif]{mathalpha}

\usepackage[ruled]{algorithm2e}
\SetKwIF{If}{ElseIf}{Else}{if}{}{else if}{else}{end if}%

\usepackage{tikz} 
\usetikzlibrary{shapes.geometric, arrows, shadows, bayesnet}

\usetikzlibrary{arrows,positioning} 
\tikzset{>=stealth', punkt/.style={rectangle, rounded corners, draw=black, very thick, text width=6.5em, minimum height=2em, text centered},pil/.style={->,thick,shorten <=2pt,shorten >=2pt,}}

\usepackage{definitions}

\usepackage{natbib}

\usepackage[accepted]{aistats2025}
\usepackage[capitalize,noabbrev]{cleveref}

\theoremstyle{plain}
\newtheorem{theorem}{Theorem}[section]
\newtheorem{proposition}[theorem]{Proposition}

\theoremstyle{definition}

\theoremstyle{remark}

\begin{document}
%
\runningtitle{Effective Bayesian Causal Inference via Structural Marginalisation and Autoregressive Orders}

%

\twocolumn[
\aistatstitle{Effective Bayesian Causal Inference via\\ Structural Marginalisation and Autoregressive Orders}

\aistatsauthor{Christian Toth \And Christian Knoll \And  Franz Pernkopf \And Robert Peharz}

\aistatsaddress{TU Graz \And  Levata GmbH \And TU Graz \And TU Graz}
]

\begin{abstract}
\looseness-1
The traditional two-stage approach to causal inference first identifies a \emph{single} causal model (or equivalence class of models), which is then used to answer causal queries.
However, this neglects any epistemic model uncertainty. 
In contrast, \emph{Bayesian} causal inference does incorporate epistemic uncertainty into query estimates via Bayesian marginalisation (posterior averaging) over \emph{all} causal models.
While principled, this marginalisation over entire causal models, i.e., both causal structures (graphs) and mechanisms, poses a tremendous computational challenge. 
In this work, we address this challenge by decomposing structure marginalisation into the marginalisation over (i) causal orders and (ii) directed acyclic graphs (DAGs) given an order.
We can marginalise the latter in closed form by limiting the number of parents per variable and utilising Gaussian processes to model mechanisms.
To marginalise over orders, we use a sampling-based approximation, for which we devise a novel auto-regressive distribution over causal orders (ARCO).
Our method outperforms state-of-the-art in structure learning on simulated non-linear additive noise benchmarks, and yields competitive results on real-world data. Furthermore, we can accurately infer interventional distributions and average causal effects.
\end{abstract}

\input{01_introduction}

\input{02_related_work}
\input{03_background}

\input{04_bci_via_orders}

\input{05_experiments}
\input{06_discussion}

\subsubsection*{Acknowledgements}
This research was funded in whole or in part by the Austrian Science Fund (FWF) 10.55776/COE12.
This research has been conducted in whole or in part within the project VENTUS (project number FO999910263), which has received funding in the framework "AI4Green, Call 2023", a research and technology program of the Austrian Ministry of Climate Action and Energy, granted by the Austrian Research Promotion Agency (FFG, \url{www.ffg.at}).
The financial support by the Austrian Federal Ministry of Labour and Economy, the National Foundation for Research, Technology and Development and the Christian Doppler Research Association is gratefully acknowledged. 
The computational results presented have in part been achieved using the Vienna Scientific Cluster (VSC).

\FloatBarrier
\bibliography{references}
\bibliographystyle{apalike}

\input{A0_checklist}

\newpage
\appendix
\onecolumn
\aistatstitle{Appendix \\\vspace{1em} Effective Bayesian Causal Inference via\\ Structural Marginalisation and Autoregressive Orders}
\vspace{-30em}
\input{A1_identifiability}

\input{A2_details}
\input{A3_exp_setup}

\input{A4_ablations}
\input{A5_proofs}

\end{document}

%% file: 01_introduction.tex
\section{INTRODUCTION}\label{sec:intro}

Few topics in science and philosophy have been as controversial as the nature of causality.
Interestingly, the discussion becomes relatively benign, from a philosophical perspective, as soon as one agrees on a well-defined mathematical model of causality, such as a \emph{structural causal model} (SCM) \citep{Pearl2009}.
Assuming that the data comes from \emph{some} model within a considered class of SCMs, causal questions reduce, in principle, to \emph{epistemic} questions, i.e., questions about what and how much is known about the model. 

\looseness-1
In principle, knowing \enquote{just} the causal structure of the true model, e.g., provided in the form of a directed acyclic graph (DAG), 
already permits the identification and estimation of causal quantities.
When lacking such structural knowledge, one typically infers a \emph{single} causal structure, which is then used to estimate the desired causal quantities (e.g., an average treatment effect).
However, committing to a single model neglects any epistemic model uncertainty stemming from the finite amount and/or quality of available data.
This is problematic, as a mismatch between the inferred causal structure and the true structure may severely affect the quality and truthfulness of the subsequent causal estimates.
Reflecting the epistemic uncertainty about the underlying causal structure in downstream causal estimates is thus of central importance.

\looseness-1
These considerations naturally invite a Bayesian treatment, i.e., specifying a prior distribution over \emph{entire} SCMs (including causal structure, mechanisms and exogenous variables) and a likelihood model to infer the posterior over SCMs given collected data.
\emph{Bayesian causal inference} (BCI) then naturally incorporates epistemic uncertainty about the true causal model into downstream causal estimates via Bayesian marginalisation (posterior averaging) over \emph{all} causal models: the causal estimate of each model is weighted with the models' posterior score.

\blankfootnote{\hspace{-1.8em}Code available at:\newline\href{https://www.github.com/chritoth/bci-arco-gp}{https://www.github.com/chritoth/bci-arco-gp}}

\looseness-1
Although BCI is conceptually appealing and principled, in practice, it becomes computationally intractable even for small problem instances due to the prohibitive number of possible causal structures to marginalise over.
For example, a model with 20 variables would require to compute and marginalise over posterior SCMs with more than $10^{72}$ candidate causal DAGs \citep{OEIS2024}.
Therefore, any practical approach to BCI must rely on approximations and/or assumptions to limit the number of considered candidate structures.

\looseness-1
In this work, we improve upon existing BCI approaches by decomposing structure marginalisation into (i) the marginalisation over causal orders and (ii) the marginalisation over DAGs given a causal order.
We perform the latter by restricting the number of parents per variable to a fixed maximal number $K$, permitting \emph{exact} marginalisation over all possible DAGs for a given order over $d$ variables in polynomial ($\mathcal{O}(d^K)$) time.
This \emph{structural marginalisation} over DAGs given causal orders, in combination with the analytic treatment of the causal mechanisms via Gaussian processes (GPs) as in \citep{Toth2022}, reduces the inference problem to the marginalisation w.r.t. causal orders.
To this end, we propose a neural \emph{A}uto-\emph{R}egressive distribution over \emph{C}ausal \emph{O}rders (ARCO), that we utilise for a sampling-based approximation.%

\looseness-1
These techniques form a novel and effective approach to BCI, where our main contributions are:
\begin{itemize}
    \setlength\itemsep{0em}
    \item We propose ARCO, a neural auto-regressive distribution over causal orders, and devise a gradient-based learning scheme for it. 
    \item We combine ARCO and exact structural marginalisation over DAGs with GPs into \emph{ARCO-GP}, an effective BCI framework, reducing the intractable BCI problem to the marginalisation over causal orders.
    \item We demonstrate in experiments that ARCO-GP sets state-of-the-art in causal structure learning against a wide range of baselines when our model assumption are exactly matched. When the model assumptions are violated, we still outperform or are competitive with all baselines.
    \item We illustrate that our method accurately infers interventional distributions, which allows us to estimate posterior average causal effects and many other causal quantities of interest.
\end{itemize}

%% file: 02_related_work.tex
\section{RELATED WORK}\label{sec:related-work}

Rather than Bayesian \emph{causal inference}, most existing literature addresses Bayesian approaches to causal \emph{structure learning}, which dates back to work as early as \citep{Heckerman1995,Heckerman1997,Madigan1995,Murphy2001,Tong2001}. 
Often based on MCMC techniques, many works utilise (causal) orders, e.g., \citep{Koller2003,Koivisto2004,Teyssier2005,Ellis2008,Niinimaki2016,Kuipers2017,Viinikka2020,Giudice2023}, for a better exploration of the posterior space.
The technique of restricting the maximum number of parents per node is well-established in Markov chain Monte Carlo (MCMC)-based structure learning \citep{Koller2003, Koivisto2004,Viinikka2020}, but has---to the best of our knowledge---not been exploited in gradient-based structure learning so far. 
Additionally, MCMC inference comes with its own set of challenges, and none of these works implement non-linear mechanism models.
In contrast, our work focuses on non-linear additive noise models and utilises gradient-based learning of a generative model over causal orders.
Besides sampling based inference, orders can also facilitate exact optimization schemes, e.g.~\citep{cussens2010maximum,de2011efficient,peharz2012exact}.

In a different stream of work utilising gradient-based (Bayesian) DAG structure learning methods \citep{Zheng2018, Yu2019, Brouillard2020, Lachapelle2020,Lorch2021,Annadani2021,Tigas2022,Deleu2022,Wang2022a,Rittel2023}, inference via orders recently gained interest in the gradient-based causal structure learning community as a vehicle to sample DAGs without the need of utilising soft acyclicity constraints during optimisation \citep{Cundy2021,Charpentier2022,Wang2022a,Annadani2023,Rittel2023}. \citet{Wang2022a} utilise probabilistic circuits to enable tractable uncertainty estimates for causal structures.
In contrast to these works, we stress the importance of an expressive gradient-based model for causal orders and we utilise causal orders for structural marginalisation in Bayesian causal inference.

Finally, existing BCI approaches are mostly restricted to linear Gaussian models \citep{geiger1994learning,Viinikka2020,pensar2020bayesian,Horii2021} or binary variables \citep{moffa2017using,kuipers2019links}.
Only recently, works on a Bayesian treatment of entire \emph{non-linear} SCMs (i.e., including mechanisms and exogenous noise) have been proposed by \citet{Toth2022,Giudice2024}.
\citet{Toth2022} focus on an active learning scenario using DIBS \citep{Lorch2021} for inferring posterior causal graphs and GPs for mechanism inference.
Similarly, in a recent pre-print, \citet{Giudice2024} follow \cite{Toth2022} in using GPs for BCI but use MCMC for sampling posterior causal graphs.
None of these approaches features gradient-based inference of causal orders, nor utilises causal orders for structure marginalisation.  

%% file: 03_background.tex
\section{BACKGROUND}
\label{sec:background}

\paragraph{Structural Causal Models.}
An SCM~$\scm$ over
observed endogenous variables~$\Xb=\{X_1, \dots, X_d\}$ and unobserved exogenous variables~$\Ub=\{U_1, \dots, U_d\}$ with joint distribution $p(\Ub)$, consists of structural equations, or mechanisms,
\begin{equation}
\label{eq:structural_eqs}
    X_i:=f_i(\mathbf{Pa}_i, U_i), \qquad \mtext{for} i\in\{1,\dots, d\},
\end{equation}
which assign the value of each $X_i$ as a deterministic function $f_i$ of its direct causes, or causal parents, $\mathbf{Pa}_i\subseteq\Xb\setminus\{X_i\}$ and an exogenous variable $U_i$.
In this paper we assume that the exogenous variables are independent Gaussian and enter in additive fashion, i.e.~$f_i(\mathbf{Pa}_i, U_i) = f_i(\mathbf{Pa}_i) + U_i$, thus implying causal sufficiency.
Associated with each SCM is a directed graph $G$ induced by the set of parent sets $\Pa = \{\Pa_i\}_{i=1}^d$ with vertices $\Xb$ and edges $X_j\to X_i$ if and only if  $X_j\in\mathbf{Pa}_i$.
Any SCM with an \emph{acyclic} directed graph (DAG) then induces a unique observational distribution $p(\Xb\given \scm)$ over the endogenous variables~$\Xb$, which is obtained as the pushforward measure of $p(\bm{U})$ through the causal mechanisms in~\cref{eq:structural_eqs}.

A (hard) intervention $do(\Wb = \wb)$ on a set of endogenous variables $\Wb \subset \Xb$ replaces the targeted mechanisms with constants $\wb$, resulting in a modified SCM. 
The entailed modified causal graph lacks the incoming edges into any intervention target.
The pushforward through the modified SCM yields an interventional distribution $p(\Xb \given do(\Wb=\wb), \scm)$.

\paragraph{Causal Orders.}
A permutation $L = \abr{L_1, \dots, L_d}$ of the endogenous variables    
$\Xb = \bigcup_{i=1}^d \{L_i\}$, where 
$L_i \neq L_j$ for all $i\neq j$, 
entails a strict total order $L_1 \prec L_2 \prec \dots \prec L_d$ among the variables. Henceforth, we refer to such a permutation $L$ as a \emph{causal order}.
A causal order $L$ constrains the possible causal interactions between the variables, i.e., $X_i$ can be a (direct) cause of $X_j$ if, and only if, $X_i \prec X_j$ in $L$.
We define $L_{<k} = \abr{L_1, \dots , L_{k-1}}$ to be the first $k-1$ elements in $L$.
Finally, let $\lambda^L \colon \Xb \mapsto \{1, \dots, d\}$ be the bijective mapping between $\Xb$ and  indices in $L$, i.e.,~\mbox{$\lambda^L(X_i) = k \iff L_k = X_i$}.
We then denote by $Q^L\in\{0, 1\}^{d\times d}$ the \emph{permutation matrix} representing $L$, where $Q^L_{ij} = 1$ iff $\lambda^L(X_i) = j$.

%% file: 04_bci_via_orders.tex
\section{BAYESIAN CAUSAL INFERENCE VIA STRUCTURAL MARGINALISATION}
\label{sec:bci-via-cos}

\input{tikz/fig-scm}

Within the Bayesian causal inference (BCI) framework, we refer to the causal quantity of interest as the \emph{causal query} $Y$, which is a function of the SCM $\scm$ (see \citep{Toth2022}).
This causal query could be, for example, an endogenous variable under some intervention, (features of) the true causal graph, or even the entire SCM.
Since we are following a Bayesian approach, $\scm$ is a random variable equipped with a prior distribution $p(\scm)$, and hence also $Y$ is a random variable with distribution $p(Y\given \scm)$.
In this work, we focus our practical implementation on acyclic, non-linear additive Gaussian noise models, which informs our prior $p(\scm)$ and likelihood $p(\Dcal\given\scm)$ accordingly (see \cref{sec:arco,sec:psm,sec:gps}).
Under our assumptions, the true graph and causal effects are identifiable (see discussion in \cref{app:identifiability}).

Given a set of (observational) data $\Dcal = \{\xb_n \overset{\text{i.i.d}}{\sim} p(\Xb\given\scm^*)\}_{n=1}^N$ collected from the true underlying SCM $\scm^*$, BCI aims at inferring the posterior $p(Y\given \Dcal)$ of the causal query.
Since the causal query $Y$ is determined by the SCM, we obtain the query posterior
\begin{align}\label{eq:query_posterior}
p(Y \given \Dcal) &= \int p(\scm\given\Dcal)\cdot p(Y\given\scm) \,d\scm \notag\\
                  &= \EE_{\scm\given\Dcal}[\,p(Y\given\scm)],
\end{align}
by marginalising over posterior SCMs.
Simply put, to practically perform BCI, we first learn a generative model over SCMs $\scm = (G, \fb, \psib)$ which we parametrise by a causal graph $G$, causal mechanisms $\fb$ and parameters $\psib$ of a joint distribution over mechanisms and exogenous variables $p(\fb, \Ub\given \psib)$ (see description in \cref{fig:scm}).
This allows us to sample from and/or evaluate $p(Y\given\scm)$.
We then use our model to (approximately) marginalise over posterior SCMs as in \cref{eq:query_posterior}.

\subsection{Learning and Inference}
Our inference procedure, described in \cref{alg:bci}, divides into a parameter learning and an inference phase. 
In the learning phase (\cref{alg:bci}, lines~\ref{alg:bci-train-start}-\ref{alg:bci-train-end}), we infer posterior parameters $p(\thetab, \psib \given \Dcal)$ of our generative model outlined in \cref{fig:scm}.
In the inference phase (\cref{alg:bci}, lines~\ref{alg:bci-inference-start}-\ref{alg:bci-inference-end}), we use samples drawn from the learned generative model to approximate the query posterior.
We first provide an overview of the individual phases starting with the inference phase, as it motivates our learning objective.
We elaborate on the more technical details in \cref{sec:arco,sec:psm,sec:gps}.

\paragraph{Inference Phase.}
The query posterior in \cref{eq:query_posterior} can be written as the following importance weighted expectation w.r.t. our generative model in \cref{fig:scm} (for a derivation see \cref{app:scm-expectation}):
\begin{align}\label{eq:scm_expectation}
    &p(Y \given \Dcal) = \EE_{\scm\given\Dcal}\sbr{p(Y\given\scm)} \\\notag
    &= \EE_{\thetab,\psib\given\Dcal}\sbr{\EE_{L\given\thetab} \sbr{w^L\cdot\EE_{G\given L, \psib, \Dcal}\sbr{\EE_{\fb\given \psib, \Dcal}\sbr{p(Y\given\scm)}}}}
\end{align}
with importance weights 
\begin{align}\label{eq:importance-weight}
    &w^L:= \frac{\EE_{G\given L}\sbr{p(\Dcal\given\psib, G) \cdot p(\psib\given G)}}{\EE_{L'\given\thetab}\sbr{\EE_{G'\given L'}\sbr{p(\Dcal\given\psib, G') \cdot p(\psib\given G')}}}.
\end{align}

In the inference phase, we assume that we have already learned a set of posterior parameters $\thetab, \psib$ approximating $p(\thetab, \psib \given \Dcal)$ (see the learning phase).
On a high level, the query posterior $p(Y\given \Dcal)$ in \eqref{eq:scm_expectation} is then estimated by (i) sampling several candidate SCMs from the learned generative model in a nested manner given $\thetab, \psib$---first sampling an order $L$, then a graph $G$ (parent sets) conditional on $L$, and mechanisms $\fb$ conditional on $G$, (ii) sampling queries given the candidate SCM from $p(Y\given\scm)$, and (iii) weighting the sampled queries with their corresponding importance weight $w^L$. See \cref{app:query-estimation} for details.

\paragraph{Learning Phase.}
To estimate the query posterior in \cref{eq:scm_expectation} we need to approximate $p(\thetab, \psib \given \Dcal)$ (see the inference phase).
By ensuring that the generative distributions $p(L\given \thetab)$ and $p(\fb, \Ub\given \psib)$ of our model are sufficiently expressive, it suffices to infer a maximum a posteriori (MAP) estimate of the parameters $\thetab, \psib$ via gradient-based optimisation of (see \cref{app:grads} for a derivation)
\begin{align}\label{eq:posterior_grad}
    \nabla \log &p(\thetab, \psib\given\Dcal) =\nabla \log p(\thetab) \\\notag
    &+ \nabla \log \EE_{L\given \thetab} \sbr{\EE_{G\given L} \sbr{p(\Dcal\given \psib, G) \cdot p(\psib \given G)}}.
\end{align}

A major issue for the training of such a model is the quality of the estimated gradients in the face of the high-dimensional problem space and the coupling of the parameters $\thetab, \psib$.
Specifically, updating all model parameters simultaneously in a single gradient step is prone to yield very noisy gradients:
\begin{enumerate}
    \item The gradient w.r.t. $\thetab$ (causal order model) depends on $\psib$ (mechanism and noise parameters) through the quality of the estimated importance weights $w^L$ in \cref{eq:importance-weight,eq:arco-grad}. A bad estimate of $\psib$ will result in poor estimates of the importance weights and consequently the gradient w.r.t. $\thetab$.
    \item The gradient w.r.t. $\psib$ likewise depends on $\thetab$ via the quality of the sampled orders and their induced parent sets. In general, the more often a parent set occurs in the sampled orders relative to other parent sets, the stronger its gradient w.r.t. $\psib$.%
    \footnote{Arguably, this may be especially problematic when sharing parameters between mechanisms with different parent sets, as is the case, e.g.,  when modeling the mechanisms with a single, masked neural network.}
\end{enumerate}

To mitigate these issues, we propose a nested optimisation procedure as laid out in \cref{alg:bci} (lines~\ref{alg:bci-train-start}-\ref{alg:bci-train-end}).
In an outer loop, we learn the parameters $\thetab$ of our generative distribution $p(L\given \thetab)$ using gradient-based optimisation of \cref{eq:arco-grad} (see \cref{sec:arco}).
However, the gradient estimates w.r.t. $\thetab$ in \cref{eq:arco-grad}, depend on the mechanism and noise parameters $\psib$ needed to compute the importance weights $w^L$.
Therefore, \emph{before} computing the importance weights, we optimise the GP hyper-parameters in an inner loop (see \cref{sec:gps}) for each yet unseen parent set compatible with some sampled order (\cref{alg:bci} (lines~\ref{alg:compute-iw-start}-\ref{alg:compute-iw-end})).
This ensures truthful estimates of the importance weights for the sampled orders and, consequently, provides good gradient estimates.

\let\oldnl\nl
\newcommand{\nonl}{\renewcommand{\nl}{\let\nl\oldnl}}
{
\begin{algorithm}[!ht]
\LinesNumbered
\SetAlgoLined
\SetAlgoSkip{}
\SetSideCommentRight
\DontPrintSemicolon
\caption{BCI with ARCO-GP}
\label{alg:bci}
\SetKwComment{cmt}{$\vartriangleright$}{}
\KwIn{(Observational) data $\Dcal$.}
\KwOut{Posterior parameters $\thetab$, $\psib$. Estimated (posterior over the) causal query $Y$ and importance weights $\wb^L$.}
\SetKwFunction{computeiw}{ComputeIW}
\tcc{Learning Phase}\label{alg:bci-train-start}
\Repeat{\upshape convergence}{
sample causal orders $\Lb \leftarrow \cbr{L_m\sim p(L\given \thetab)}$ \\\nonl\hfill  \cmt*{\cref{sec:arco}}
$\wb^L, \psib \leftarrow$ \computeiw{$\Lb$, $\Dcal$}  \\\nonl\hfill\cmt*{\cref{eq:importance-weight}}\label{alg:bci-computeiw}
estimate the gradients for $\thetab$ and perform a gradient step \cmt*{\cref{eq:arco-grad}}
} \label{alg:bci-train-end}
\tcc{Inference Phase}
sample causal orders $\Lb \leftarrow \cbr{L_m\sim p(L\given \thetab)}$ \\\nonl\hfill\cmt*{\cref{sec:arco}} \label{alg:bci-inference-start}
$\wb^L, \psib \leftarrow$ \computeiw{$\Lb$, $\Dcal$}  \\\nonl\hfill\cmt*{\cref{eq:importance-weight}}
\emph{[optional]} sample graphs $\Gb \leftarrow \cbr{G_k\sim p(G\given L,\psib,\Dcal)}$ \cmt*{\cref{sec:psm}}
\emph{[optional]} sample mechanisms $\fb \leftarrow \cbr{\fb_j\sim p(\fb\given \psib,\Dcal)}$ \cmt*{\cref{sec:gps}}
sample candidate queries $\Yb \leftarrow \cbr{Y_i\sim p(Y\given \fb, \psib, G)}$ \cmt*{\cref{sec:gps}}
\Return $\thetab$, $\psib$, $\Yb$, $\wb^L$ \\ \label{alg:bci-inference-end}
\tcc{Subroutine: compute importance weights and update mechanisms.}
\SetKwProg{myproc}{Subroutine}{}{}
\myproc{\computeiw{$\Lb$, $\Dcal$}}{ \label{alg:compute-iw-start}
\ForEach{\upshape causal order $L_m \in \Lb$ }
{
    \ForEach{\upshape parent set $\Pa_i$ compatible with $L_m$}
    {
        Learn (or retrieve) the corresponding GP hyper-parameters $\psib_i$ \\\nonl\hfill\cmt*{\cref{eq:psi-grad}} \label{alg:bci-update-hp} 
    }
    Compute the importance weight $w^{L_m}$ \\\nonl\hfill\cmt*{\cref{eq:importance-weight}}
}
\Return $\wb^L, \psib$ \label{alg:compute-iw-end}
}
\end{algorithm}
}

\subsection{Marginalising over Causal Orders}\label{sec:arco}

Computing the expectation w.r.t.~causal orders in \cref{eq:scm_expectation} poses a hard combinatorial problem, as there are $d!$ possible causal orders over $d$ variables.
The involved distribution over causal orders $p(L\given \thetab)$ appearing in the BCI estimator in \cref{eq:scm_expectation} can be interpreted as proposal distribution with importance weights as defined in \cref{eq:importance-weight}, where the optimal proposal is the true posterior over causal orders.
Hence, we require an expressive representation over causal orders which can account for \emph{multi-modal distributions} over orders.
In particular, consider the example of a Markov equivalence class (MEC) including a chain graph.
Since the chain is contained in the MEC, also the reverse chain graph must be in the MEC, and thus, the proposal over causal orders must be able to represent both orders with equal probability.

A simple parameterisation of orders as proposed in~\citep{Charpentier2022} is not able to represent the true posterior over causal orders in this case.
Specifically, they sample orders using the Gumbel Top-k trick \citep{Kool2019a} by perturbing $d$ logits (corresponding to the $d$ variables) with Gumbel noise and sorting these perturbed logits, yielding an order over variables.
In essence, learning such a model boils down to ordering (and spreading) the $d$ logits on the real line.
Now, to sample a chain graph and a reverse chain graph, some variable is the first element in the causal order in one case and must thus have the highest (perturbed) logit, and it is the last element in the causal order in the other case where it must have the lowest (perturbed) logit, which is contradictory.
In practice, we observe that when trying to learn a multi-modal distribution over causal orders with this model, the logits cluster together, resulting approximately in a uniform distribution over causal orders.

We therefore propose an expressive, auto-regressive distribution \citep{larochelle2011neural}
$
    p(L\given\thetab) = p(L_1 \given \thetab) \cdot \prod_{k=2}^d p(L_k \given L_{<k}, \thetab)
$
over causal orders (ARCO) that is amenable to gradient-based optimisation and can represent multi-modal distributions over orders, avoiding the shortcoming described above.%
\footnote{ARCO can be understood as an auto-regressive distribution estimator \citep{larochelle2011neural} constrained to causal orders.}

\paragraph{Sampling Causal Orders.}
The ordering of variables naturally implies a sequential sampling procedure as listed in \cref{alg:arco}.
In each step of the sampling procedure, we sample the next variable in the order from a categorical distribution $p(L_k \given L_{<k}, \thetab)$ over the set of yet unassigned variables, conditional on all preceding variables in the order (\cref{alg:arco}, line~\ref{alg:arco-cat}).
To account for the dependence on the preceding order $L_{<k}$, we compute the logits of the categorical distribution using a differentiable function $g_{\thetab}: \RR^{d\times d} \mapsto \RR^d$ (\cref{alg:arco}, line~\ref{alg:arco-logits}) and re-normalise them to exclude the elements in $L_{<k}$ (\cref{alg:arco}, line~\ref{alg:arco-normalise}).
We implement $g_{\thetab}$ as feed-forward neural network, taking as input a suitable encoding of the so-far sampled order $L_{<k}$.
To this end, we encode $L_{<k}$ using its induced permutation matrix $Q^{L_{<k}}$ (see \cref{sec:background}) and mask the rows corresponding to elements $L_{>=k}$ with zeros.

\paragraph{Training ARCO.}
Training ARCO amounts to learning the parameters $\thetab$ of the neural network $g_{\thetab}$ by performing gradient ascent on \cref{eq:posterior_grad}.
As \cref{eq:posterior_grad} is not differentiable w.r.t. $\thetab$  because of the discrete nature of causal orders, we use the score-function estimator \citep{Williams1992} to estimate the gradients via (see \cref{app:scm-expectation,app:grads})
\begin{align}\label{eq:arco-grad}
    \nabla_{\thetab} \log p(\thetab, \psib \given \Dcal)& = \nabla_{\thetab} \log p(\thetab) \\\notag 
    &+ \EE_{L\given\thetab}\sbr{ w^L \cdot \nabla_{\thetab} \log p(L \given \thetab)},
\end{align}
with $w^L$ as defined in \cref{eq:importance-weight}.
To evaluate $\log p(L\given \thetab)$
for a given causal order $L$, we simply need to sum the log-probabilities of the categorical distributions $\log p(L_k \given L_{<k}, \thetab)$ for the respective elements $L_k$.
The necessary log-probabilities (i.e., the normalised logits) are computed as described in \cref{alg:arco}.
Note that, although we need to compute the logits sequentially in the sampling procedure, we can compute them in parallel during evaluation.

{
\begin{algorithm}[t]
\LinesNumbered
\SetAlgoLined
\SetAlgoSkip{}
\DontPrintSemicolon
\caption{Sample Causal Order (ARCO)}
\label{alg:arco}
\SetKwComment{cmt}{$\vartriangleright$}{}
\KwIn{Logit function $g_{\thetab}: \RR^{d\times d}\mapsto \RR^d$}
\KwOut{Causal order $L$ sampled from $p(L\given \thetab)$}
$\Rb \leftarrow \Xb$ \cmt*{set of unassigned elements}
$L \leftarrow \0$ \cmt*{causal order}
\For{$k=1\dots d$}{
 $\phib \leftarrow g_{\thetab}(Q^{L_{<k}})$ \cmt*{compute logits} \label{alg:arco-logits}
 $\phi_i \leftarrow \begin{cases}
     \phi_i - \log \sum_{j\given X_j \in \Rb} \exp \phi_j \mtext{if}  X_i\in \Rb \\
     -\infty\mtext{otherwise}
 \end{cases}$ \\\nonl\hfill\cmt*{normalise logits} \label{alg:arco-normalise}
 $l \sim \text{CATEGORICAL}(\phib)$ \\\nonl\hfill\cmt*{sample next element}\label{alg:arco-cat}
 $L \leftarrow L\cup \langle X_l \rangle$ \cmt*{update causal order}
 $\Rb \leftarrow \Rb\setminus \{X_l\}$  
}
\end{algorithm}
}

\subsection{Marginalising over Causal Graphs}\label{sec:psm}
The marginalisation w.r.t. causal graphs $G$ in \cref{eq:scm_expectation,eq:importance-weight,eq:posterior_grad} is in general intractable, as the number of DAGs consistent with any given (causal) order is $2^\frac{d\cdot (d-1)}{2}$. Although this is significantly smaller than the total number of DAGs with $d$ nodes (which grows super-exponentially in $d$, see e.g. \citep{OEIS2024}), an exhaustive enumeration is still infeasible.

In this work, we tackle this problem by restricting the number of parents per variable.
By restricting the maximum size of any admissible parent set to some integer $K$, the number of distinct parent sets consistent with any causal order $L$ is in $\mathcal{O}(d^K)$.
Although the exhaustive enumeration of all DAGs with restricted parent set size is still infeasible, it turns out that, given a causal order, the expectation w.r.t. graphs in \cref{eq:scm_expectation} can be tractably computed under certain assumptions.
Rewriting the relevant expectation yields
\begin{align}\label{eq:graph-expectation}
    \EE_{G\given L, \psib, \Dcal}\sbr{Y(G)} = \EE_{G\given L}[w(G) \cdot Y(G)]
\end{align}
by letting $Y(G) := \EE_{\fb\given \psib, \Dcal}\sbr{p(Y\given\scm)}$ to avoid clutter, and
\begin{align}\label{eq:graph-weights}
    w(G) = \frac{p(\Dcal\given\psib, G) \cdot p(\psib\given G)}{\EE_{G\given L}\sbr{p(\Dcal\given\psib, G) \cdot p(\psib\given G)}}.
\end{align}
For $w(G)$ and the prior over graphs $p(G \given L) = \prod_i p(\Pa_i\given L)$ factorising, we can compute queries $Y(G)$ that decompose over the parent sets by the following two propositions (proofs in \cref{app:proposition-proofs}).%
\footnote{In practice, we assume a uniform prior over parent sets consistent with a given causal order.}%
\footnote{These propositions generalise the results presented by \citet{Koller2003, Koivisto2004} on how to compute the posterior probabilities of edges or parent sets, which is a special case of \cref{prop:ex-factorising}, to our setting of Bayesian causal inference.}

\begin{proposition}\label{prop:ex-factorising}
    Let $Y(G) = \prod_i Y_i(\Pa_i^G)$ and $w(G) = \prod_i w(\Pa_i^G)$ be factorising over the parent sets, then
    \begin{align}
    \EE_{G\given L} \sbr{w(G) Y(G)} = \prod_{i=1}^d \sum_{\Pa_i} p(\Pa_i\given L) w_i(\Pa_i) Y_i(\Pa_i).
    \end{align}
\end{proposition}
\begin{proposition}\label{prop:ex-summing}
    Let $Y(G) = \sum_i Y_i(\Pa_i^G)$ be summing and $w(G) = \prod_i w(\Pa_i^G)$ be factorising over the parent sets, then
    \begin{align}
        \EE_{G\given L} &\sbr{w(G) Y(G)} = \\\notag 
        &\sum_{i=1}^d \rbr{\prod_{j\neq i} \alpha_j(L)}\cdot \sum_{\Pa_i} p(\Pa_i\given L) w_i(\Pa_i) Y_i(\Pa_i),
    \end{align}
    where
    \begin{align*}
        \alpha_j(L) = \sum_{\Pa_i} p(\Pa_i\given L) w_i(\Pa_i).
    \end{align*}
\end{proposition}

Furthermore, for any causal query that does not decompose over parent sets, we can compute a Monte-Carlo estimate of the query by sampling DAGs from the true posterior $p(G\given L, \psib, \Dcal) = w(G) \cdot p(G\given L)$ in \cref{eq:scm_expectation}.
Employing \cref{prop:ex-factorising}, we can compute $\EE_{G\given L}\sbr{p(\Dcal\given\psib, G) \cdot p(\psib\given G)}$ and consequently $w(G)$ in closed-form by choosing $w(G) = p(\Dcal\given\psib, G) \cdot p(\psib\given G)$ and $Y(G) = 1$.
Since $p(G\given L, \psib, \Dcal)$ factorises, we can sample DAGs by sampling parent sets individually for each node.

\begin{figure*}[t]
\def\figwidth{1.\textwidth}
     \centering 
     \begin{subfigure}[b]{\figwidth}
         \centering
         \includegraphics[width=\textwidth]{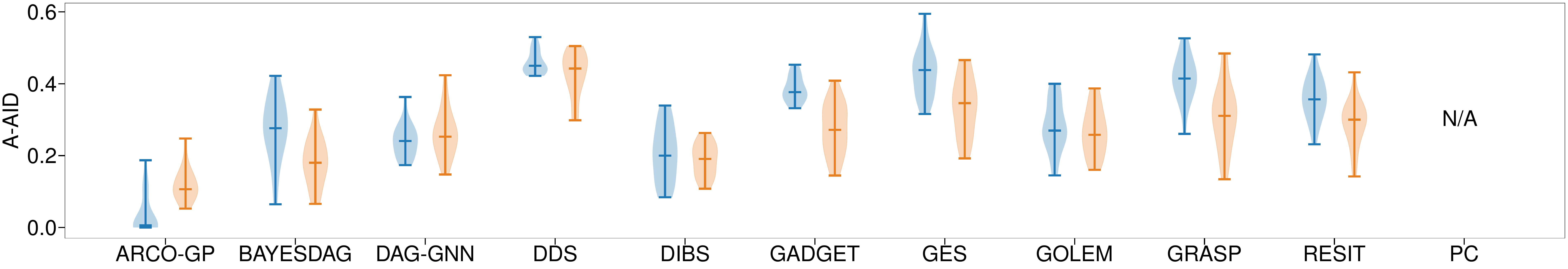}
     \end{subfigure}\\
     \vfill
     \begin{subfigure}[b]{\figwidth}
         \centering
         \includegraphics[width=\textwidth]{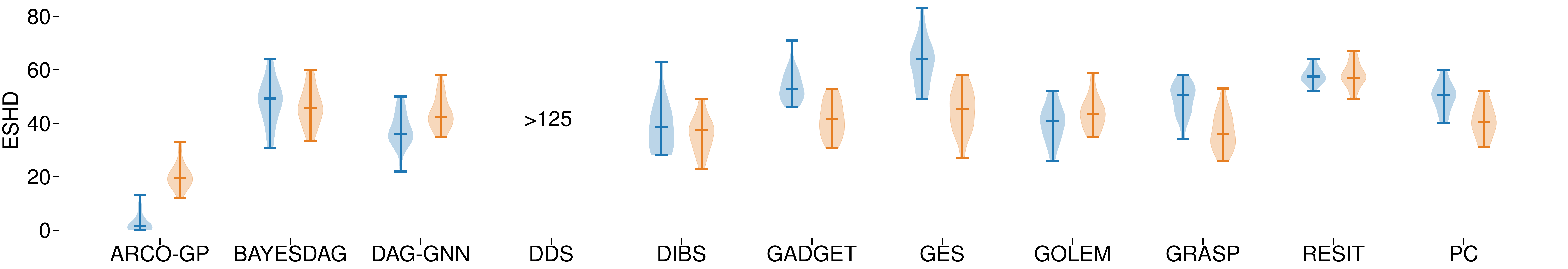}
     \end{subfigure}\\
    \caption{\textbf{Causal discovery on nonlinear additive noise models.}
    Structure learning results in terms of \emph{expected Hamming distance} (ESHD) and \emph{ancestor adjustment identification distance} (A-AID) on simulated non-linear models with scale-free (left, blue) and Erdös-Renyi (right, orange) graphs, each with $20$ nodes and $200$ data samples.
    Whiskers indicate maximum, minimum and median values across $20$ simulated ground truth instances.
    For both metrics lower is better.
    Range for ESHD is set for better readability, omitting the result for DDS ($>125$).
    }\label{fig:main_cd}
    \vspace{-1em}
\end{figure*}

\subsection{Mechanism Inference and Marginalisation}\label{sec:gps}
To compute the importance weights in \cref{eq:importance-weight,eq:graph-weights} we need to compute the marginal log-likelihood $p(\Dcal\given\psib, G) = \prod_i p(\Dcal_i\given\psib_i, \Pa_i^G)$ which is intractable for general models.%
\footnote{We abuse notation to denote by $p(\Dcal_i\given\psib_i, \Pa_i^G) := p(\xb_i\given\xb_{\Pa_i^G},\psib_i, \Pa_i^G)$}.
As we focus on non-linear additive noise models in this work, we follow \citep{VonKugelgen2019,Toth2022} and model each mechanism via a distinct GPs, assuming homoscedastic Gaussian noise and causal sufficiency.
Under these assumptions, we can compute the marginal likelihood
$
    p(\Dcal_i\given \psib_i, \Pa_i) = \EE_{\fb\given \psib_i}\sbr{p(\Dcal_i\given \fb_i,\psib_i, \Pa_i)} 
$
and the GP predictive posterior in closed form.
The GP hyper-parameter prior and likelihood factorise over parent sets, i.e.,
\begin{align}
    p(\Dcal\given\psib,G)\cdot p(\psib\given G) = \prod_i p(\Dcal_i\given \psib_i, G) \cdot p(\psib_i \given \Pa_i^G).
\end{align}
In practice, we thus use a separate GP for each unique parent set (in contrast to having a separate set of GPs for each individual graph), allowing for efficient computation by caching intermediate results.
For the individual GP we infer a MAP-Type II estimate of its hyper-parameters $\psib$ by performing gradient-ascent on%
\footnote{Importantly, in \cref{app:grads} we show that this not an ad-hoc choice, but a consequence of optimising \cref{eq:posterior_grad}.}
\begin{align}   \label{eq:psi-grad}
     \nabla_{\psib_i} \log &p(\psib_i \given \Dcal, G) = \\\notag 
     &\nabla_{\psib_i} \log p(\Dcal_i\given\psib_i,G) + \nabla_{\psib_i} \log p(\psib_i\given G).
\end{align}
For root nodes, i.e., nodes without parents, we place a conjugate normal-inverse-gamma prior on the mean and variance of that node, which also allows for closed-form inference.

%% file: tikz/fig-scm.tex
\begin{figure}[t]
    \def\xshift{4.0em}
    \def\yshift{-4.0em}
    \def\nodesize{\footnotesize}
    \footnotesize
    \centering
      \begin{tikzpicture}[node distance=1.5cm, auto,]
        \footnotesize
        \node (A) [latent, xshift=0*\xshift, yshift=0.*\yshift] {\nodesize $\thetab$};
        \node (B) [latent, xshift=1*\xshift, yshift=0.*\yshift] {\nodesize $L$};
        \node (E) [latent, xshift=2*\xshift, yshift=0*\yshift] {\nodesize $G$};
        \node (F) [latent, xshift=3*\xshift, yshift=0*\yshift] {\nodesize $\psib$};
        \node (G) [latent, xshift=4*\xshift, yshift=0*\yshift] {\nodesize $\fb$};
        \node (H) [latent, xshift=2.5*\xshift, yshift=1*\yshift] {\nodesize $\Dcal$};
        \node (I) [latent, xshift=3.5*\xshift, yshift=1*\yshift] {\nodesize $Y$};
        \edge{A}{B};
        \edge{B}{E};
        \edge{E}{F,H,I};
        \edge{F}{G,H,I};
        \edge{G}{H,I};
        \tikzset{plate caption/.style={caption, above right=0.2em and 6em of #1.north west}}
        \plate[inner sep=0.8em]{scm}{(E) (F) (G)}{SCM $\scm$};
      \end{tikzpicture}      
  \caption{\textbf{Generative model of ARCO-GP.}
  We characterise a Structural Causal Model (SCM) $\scm = (G, \fb, \psib)$ by a causal graph $G$, causal mechanisms $\fb$ and parameters $\psib$ of a joint distribution over mechanisms and exogenous variables $p(\fb, \Ub\given \psib)$.
  We model the mechanisms $\fb$ using Gaussian Processes (GPs) and $\Ub$ as additive Gaussian noise, implying that $\psib$ is a set of GP hyper-parameters.
  The SCM gives rise to the data-generating likelihood $p(\Dcal\given \fb, \psib, G)$ and determines the (distribution over the) causal query $Y$.
  To sample a SCM, we first sample a causal order $L$ from a neural auto-regressive distribution over causal orders (ARCO) $p(L\given \thetab)$ with parameters $\thetab$.
  Given a causal order and assuming a limited maximum cardinality of parent sets, we can then sample or marginalise causal graphs $G$ and mechanisms $\fb$ in closed form.
  }
  \label{fig:scm}
\end{figure}

%% file: 05_experiments.tex
\section{EXPERIMENTS}\label{sec:experiments}

As a truthful (posterior over) causal structure is paramount for downstream causal inferences, Task I evaluates structure learning.
To illustrate ARCO-GP's capability of performing integrated causal structure learning and reasoning, Task II evaluates the capability to infer interventional distributions from observational data.

\textbf{Task I: Causal Structure Learning.}
For our results in \cref{fig:main_cd}, we sample a fixed set of $200$ training samples from the observational distribution of non-linear additive noise ground truth SCMs with Erdös-Rényi (ER) \citep{Erdos1959} and scale-free (SF) \citep{Barabasi1999} graph structures with $20$ nodes.
The small sample size emulates the setting of significant uncertainty relevant to the Bayesian inference scenario.
We report the \emph{expected structural Hamming distance} (ESHD) as a standard structure learning metric.
Additionally, to asses the inferred structures in terms of causal implications, we report the \emph{ancestor adjustment identification distance} (A-AID) recently proposed by~\citet{Henckel2024} (see \cref{app:ablations} for details and additional metrics).

We compare our inference model (ARCO-GP) with maximum parent set size $K=2$ to a diverse set of ten different structure learning methods (see \cref{app:ablations} for descriptions): BAYESDAG \citep{Annadani2023}, DAG-GNN \citep{Yu2019}, DDS \citep{Charpentier2022}, DIBS-GP \citep{Toth2022,Lorch2021}, GADGET \citep{Viinikka2020}, GES \citep{Chickering2003}, GOLEM \citep{Ng2020}, GRASP \citep{Lam2022}, PC \citep{Spirtes2000} and RESIT \citep{Peters2014}.

The results in \cref{fig:main_cd} demonstrate that ARCO-GP is highly effective in terms of causal structure learning. 
In particular, for scale-free graphs, where the assumption of a restricted number of parents matches our modelling assumption, ARCO-GP clearly outperforms all baselines. 
Moreover, even when the assumptions are violated (in our case for Erdos-Renyi graphs), we demonstrate that ARCO-GP still outperforms the baselines, albeit with a smaller margin.
Additional experiments on real-world data from \citet{Sachs2005}, invertible non-linear models, and ablations varying the number of variables $d\in\cbr{11, 20, 50}$, sample sizes, as well as investigating the influence of the parent size restriction are provided in \cref{app:ablations}.


\begin{figure}[t]
    \def\figwidth{0.23\textwidth}
     \centering
     \begin{subfigure}[b]{\figwidth}
         \centering
         \includegraphics[width=\textwidth]{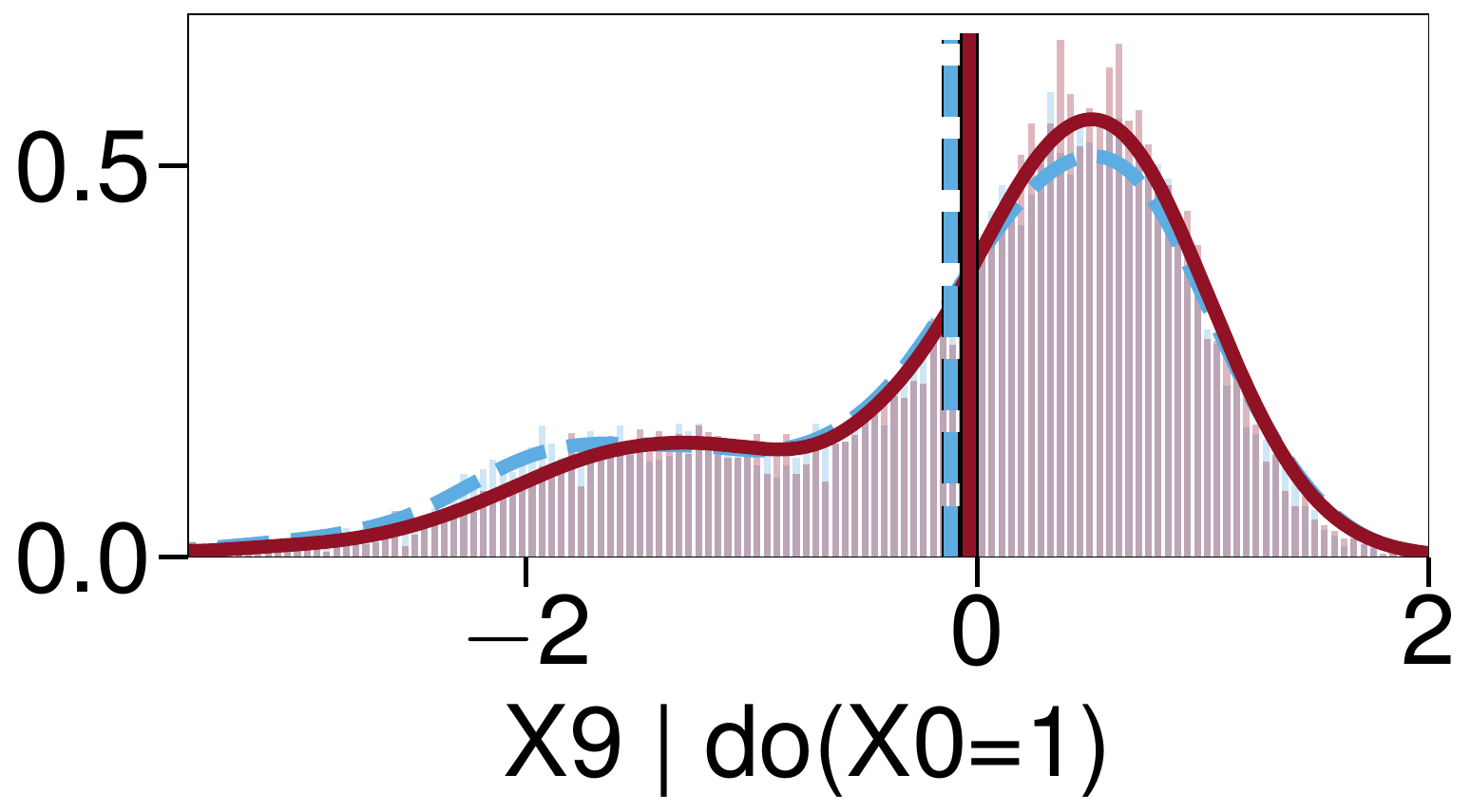}
     \end{subfigure}
     \hfill
     \begin{subfigure}[b]{\figwidth}
         \centering
         \includegraphics[width=\textwidth]{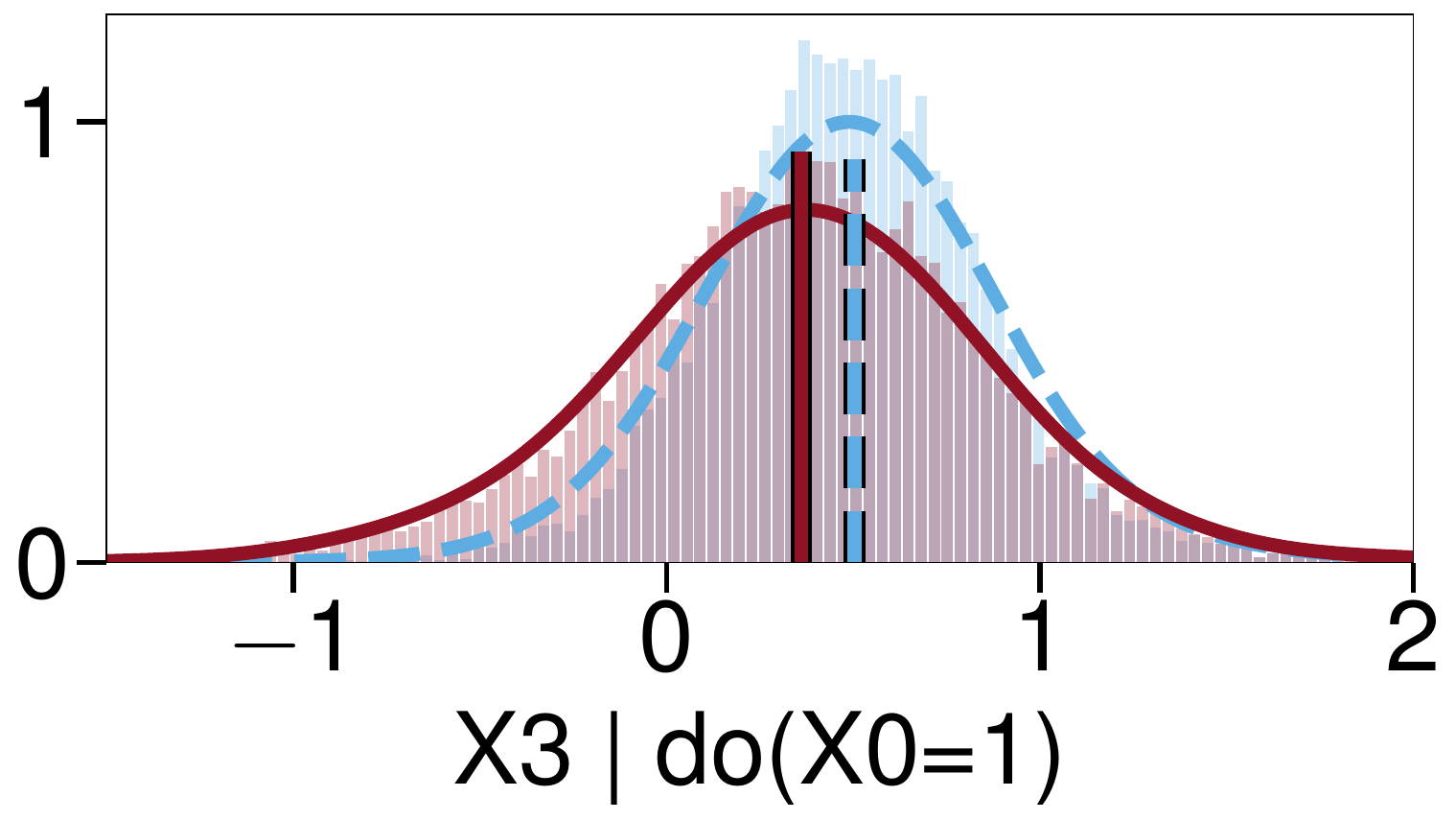}
     \end{subfigure}
     \hfill
     \\\vfill
     \begin{subfigure}[b]{\figwidth}
         \centering
         \includegraphics[width=\textwidth]{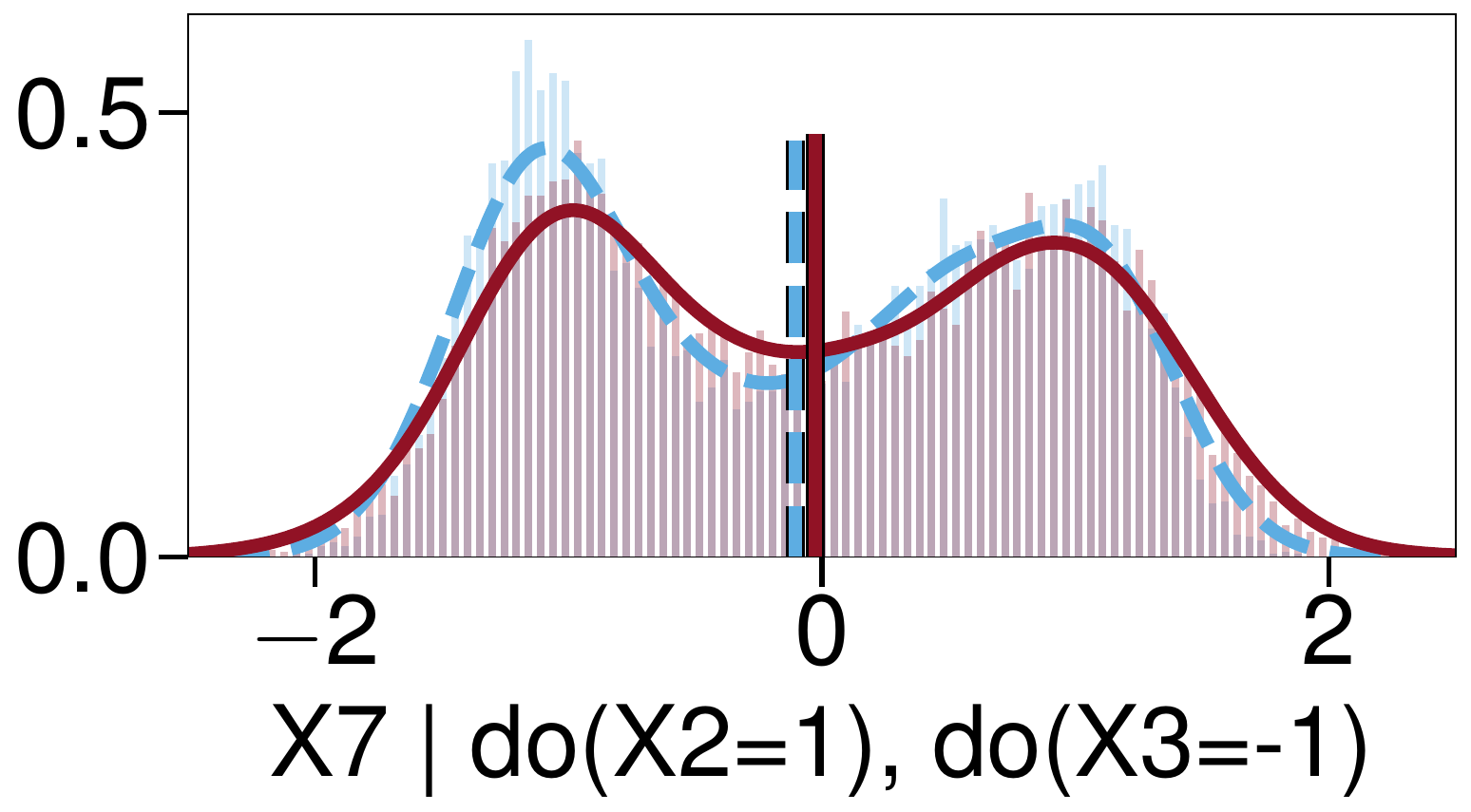}
     \end{subfigure}
     \hfill
     \begin{subfigure}[b]{\figwidth}
         \centering
         \includegraphics[width=\textwidth]{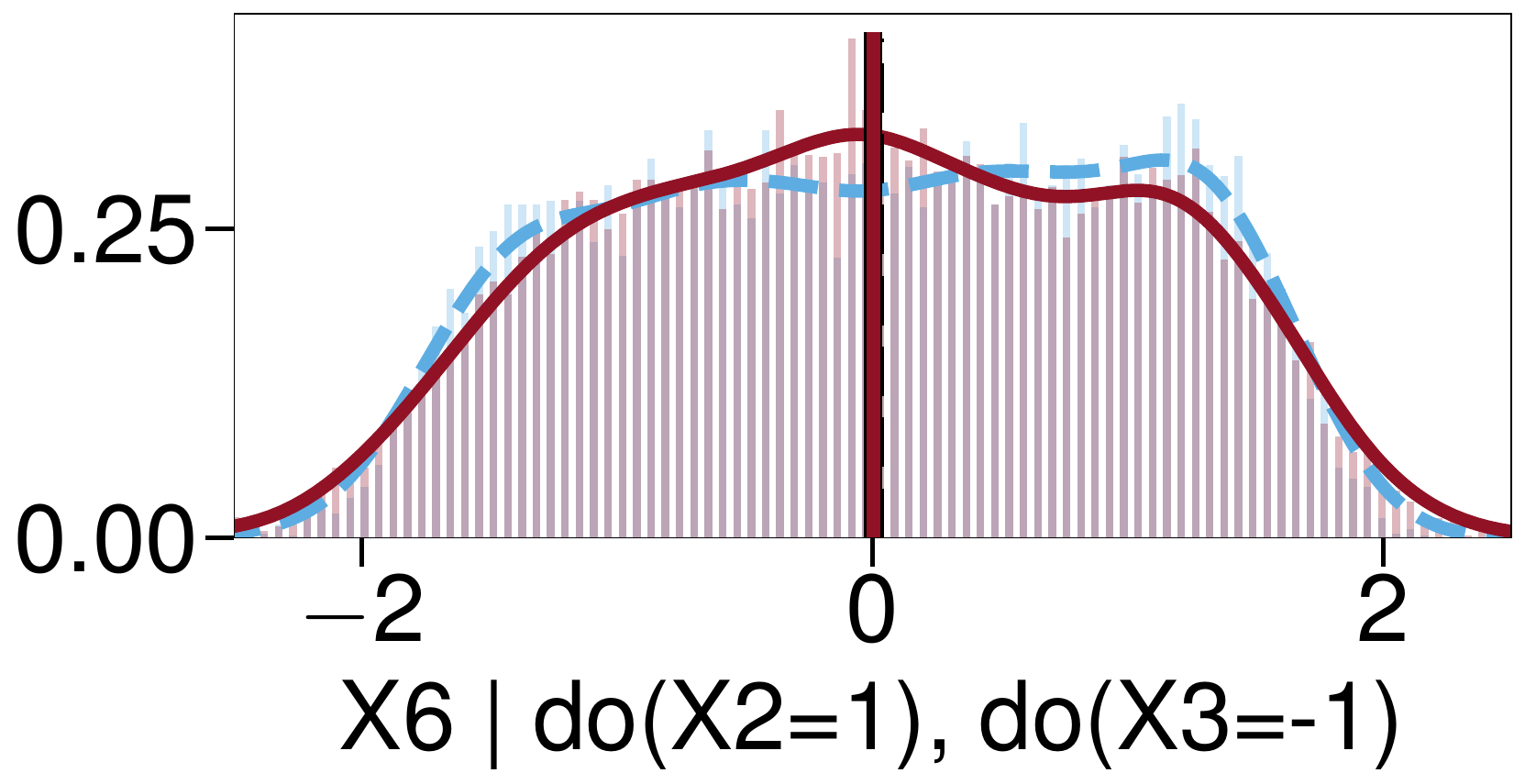}
     \end{subfigure}
     \hfill
    \caption{\textbf{Posterior interventional distributions.}
    Several interventional distributions as inferred by ARCO-GP (red, solid) and the corresponding ground truth (blue, dashed).
    Specifically, we sampled full SCMs (orders, graphs given orders, mechanisms, exogeneous variables) and performed the indicated intervention to produce a sample from the corresponding distribution, which effectively marginalises over the posterior over SCMs.
    Vertical lines indicate the estimated distribution means (average causal effects).
    See \cref{app:exp-setup} for details.
    }
    \label{fig:main_sachsgraph_idist}
\end{figure}

\textbf{Task II: Inferring Interventional Distributions and Average Causal Effects.}
We illustrate ARCO-GP's causal reasoning capabilities by visualising a selected set of estimated posterior interventional distributions $p(X_i\given do(\Wb = \wb), \Dcal)$ in \cref{fig:main_sachsgraph_idist}.
To have access to ground truth interventional distributions, we simulate a non-linear additive noise model on the consensus protein interaction graph reported by \citet{Sachs2005} (see Appendix, \cref{fig:sachs-graph}).
We generate samples from the inferred interventional distributions with the procedure laid out in \cref{alg:bci,sec:bci-via-cos}, smoothing the empirical distribution with a kernel density estimate (see \cref{app:details} for details).
Importantly, the multi-modality of the inferred distributions illustrates the benefits of a Bayesian approach to causal inference, as we can represent structural uncertainty via full distributions instead of single point estimates (see especially \cref{fig:main_sachsgraph_idist}, lower left).

To quantify performance, we use our ground truth model to simulate five random interventional distributions $p(\Xb\given \text{do}(X_i = t))$ per node with $t\in[-1, 1]$, drawing $1000$ samples per intervention.
We then compute averages of the Maximum Mean Discrepancy (MMD) \citep{Gretton2012} and L1 and L2 distances between the distribution means of inferred vs. true interventional distributions across all interventions. 
We compare against two BCI baselines, i.e., DIBS-GP \citep{Toth2022,Lorch2021} and BEEPS \citep{Viinikka2020}.
To emulate a baseline using the traditional two-stage approach, we first infer a causal graph using RESIT, that we then use to fit a GP model and infer interventional distributions (RESIT-GP).
Note that our other baselines used in Task I do not infer interventional distributions, as they are geared towards structure learning.
The results in \cref{tab:dist-scores} show that ARCO-GP outperforms the baselines with a significant gap to the emulated two-stage baseline RESIT-GP.

\begin{table}[t]
\centering
\small
\caption{\textbf{Inferring Interventional Distributions.} 
We report the MMD, as well as L1 and L2 distances between the distribution means of inferred and true interventional distributions.
We generate samples from five interventional distributions $p(\Xb\given \text{do}(X_i = t))$ per node with random intervention values $t\sim\Ucal(-1, 1)$.
The metrics are averaged across all interventions.
The reported numbers are averages and $95\%$ confidence intervals (CIs) obtained from simulations on $10$ different ground truth models with fixed graph (see \cref{fig:sachs-graph}) and simulated nonlinear mechanisms. 
}
\label{tab:dist-scores}
\begin{tabular}{@{}lccccccccccc@{}}
    \toprule
    & MMD	& L1	& L2 \\ \midrule
ARCO-GP	 & 0.15 $\pm$ 0.04	 & 0.93 $\pm$ 0.16	 & 0.37 $\pm$ 0.06	 \\
BEEPS	 &  - 	             & 1.43 $\pm$ 0.16	 & 0.62 $\pm$ 0.07	 \\
DIBS-GP	 & 0.17 $\pm$ 0.05	 & 1.13 $\pm$ 0.26	 & 0.47 $\pm$ 0.11	 \\
RESIT-GP	 & 0.31 $\pm$ 0.12	 & 1.57 $\pm$ 0.19	 & 0.68 $\pm$ 0.07	 \\
    \bottomrule
\end{tabular}%
\end{table}

%% file: 06_discussion.tex
\section{DISCUSSION}\label{sec:discussion}
We demonstrated that our proposed ARCO-GP method for BCI, leveraging structural marginalisation, yields superior structure learning performance on non-linear additive noise models against a set of ten state-of-the-art baseline methods.
Moreover, we illustrate the capability of ARCO-GP to accurately infer posterior interventional distributions and average causal effects.
The capabilities our method relies upon the following assumptions and limitations.

\textbf{Assumptions.}
Our assumptions on the data generating process include causal sufficiency and additive, homoscedastic Gaussian noise.
An extension to heteroscedastic (non-Gaussian) noise could be achieved by utilising GP extensions as, e.g., proposed by \citet{Wang2012,Dutordoir2018}, without making substantial changes to the ARCO-GP framework.

We further assume a limited maximum parent set size.
Presumably, for very dense graphs the performance of ARCO-GP will deteriorate in comparison to other methods.
However, our ablations in \cref{app:ablations} show that ARCO-GP is still superior when violations are only moderate.

While our framework and implementation can handle training from interventional data, we do not evaluate this scenario experimentally because not all baselines support interventional input data, and the observational case is the more difficult problem from the perspective of model identifiability. 

\textbf{Scalability and Computation.}
\looseness-1
The main driver of complexity is the exact inference using GPs, which grows with $N^3$ in the number of available data points.
Although we used only CPUs for running our experiments, scaleable GPU inference techniques for GPs were proposed, e.g., by \citet{Gardner2018,Pleiss2018}.
Additionally, the training of the GPs could be straightforwardly parallelised.
Conceptually, our framework is flexible and modular, allowing to use alternative mechanism models like normalising flows as in \cite{Brouillard2020,Pawlowski2020a}.
A second driver of computational complexity is the exhaustive enumeration of parent sets, which may be prohibitive on larger problem instances and bigger parent sets.
Note, however, that the individual parent set contributions necessary to compute the importance weights in \cref{eq:importance-weight} could be pre-computed in parallel.

%% file: A0_checklist.tex
\section*{Checklist}

 \begin{enumerate}

 \item For all models and algorithms presented, check if you include:
 \begin{enumerate}
   \item A clear description of the mathematical setting, assumptions, algorithm, and/or model. [Yes]
   \item An analysis of the properties and complexity (time, space, sample size) of any algorithm. [Yes]
   \item (Optional) Anonymized source code, with specification of all dependencies, including external libraries. [Yes]
 \end{enumerate}

 \item For any theoretical claim, check if you include:
 \begin{enumerate}
   \item Statements of the full set of assumptions of all theoretical results. [Yes]
   \item Complete proofs of all theoretical results. [Yes]
   \item Clear explanations of any assumptions. [Yes]     
 \end{enumerate}

 \item For all figures and tables that present empirical results, check if you include:
 \begin{enumerate}
   \item The code, data, and instructions needed to reproduce the main experimental results (either in the supplemental material or as a URL). [Yes]
   \item All the training details (e.g., data splits, hyperparameters, how they were chosen). [Yes]
         \item A clear definition of the specific measure or statistics and error bars (e.g., with respect to the random seed after running experiments multiple times). [Yes]
         \item A description of the computing infrastructure used. (e.g., type of GPUs, internal cluster, or cloud provider). [Yes]
 \end{enumerate}

 \item If you are using existing assets (e.g., code, data, models) or curating/releasing new assets, check if you include:
 \begin{enumerate}
   \item Citations of the creator If your work uses existing assets. [Yes]
   \item The license information of the assets, if applicable. [Not Applicable]
   \item New assets either in the supplemental material or as a URL, if applicable. [Yes]
   \item Information about consent from data providers/curators. [Not Applicable]
   \item Discussion of sensible content if applicable, e.g., personally identifiable information or offensive content. [Not Applicable]
 \end{enumerate}

 \item If you used crowdsourcing or conducted research with human subjects, check if you include:
 \begin{enumerate}
   \item The full text of instructions given to participants and screenshots. [Not Applicable]
   \item Descriptions of potential participant risks, with links to Institutional Review Board (IRB) approvals if applicable. [Not Applicable]
   \item The estimated hourly wage paid to participants and the total amount spent on participant compensation. [Not Applicable]
 \end{enumerate}

 \end{enumerate}

%% file: A1_identifiability.tex
\section{IDENTIFIABILITY OF CAUSAL QUERIES}\label{app:identifiability}
In Bayesian causal inference, our goal is to infer a query posterior $p(Y\given\Dcal) = \EE_{\scm\given\Dcal}\sbr{p(Y\given\scm)}$ given collected (observational) data $\Dcal$ from the underlying true SCM $\scm^\true$ (see \cref{sec:bci-via-cos}).

As a consequence of this formulation, the query posterior $p(Y\given\Dcal)$ will match the true query distribution $p(Y\given\scm^\true)$, if the posterior over causal models $p(\scm\given\Dcal)$ converges to a point mass on the true SCM $\scm^\true$ in the limit on infinitely many data.%
\footnote{In case the query is fully determined by the SCM, then $p(Y\given\scm)$ has a point mass on $Y(\scm)$ and $p(Y\given\Dcal)$ will converge to a point mass on $Y(\scm^\true)$. For example, let $Y(\scm) = G$ be the causal graph, then $p(Y\given\Dcal)$ will asymptotically concentrate all its mass on the true graph $G^\true$.}
Therefore, whether a causal query $Y$ (or a query distribution $p(Y\given\scm)$) is identifiable (in a statistical rather than a causal sense), depends on our choices of likelihood $p(\Dcal\given\scm)$ and prior $p(\scm)$ over SCMs.

Relating to causal identifiability in our problem setup, recall that we focus on non-linear additive noise models, i.e., we place a GP prior with a Gaussian likelihood on the mechanisms.
Identifiability of non-linear additive noise models has been established under mild conditions in \citep{Hoyer2009,Peters2014,Buhlmann2014}.
Further note that, as we preclude unobserved confounding, causal effects are identifiable given the causal DAG. 
Thus, in the limit of infinitely many data, our posterior would collapse onto the true SCM and causal effects could be truthfully estimated.

%% file: A2_details.tex
\clearpage
\section{IMPLEMENTATION AND ESTIMATION DETAILS}\label{app:details}
Our model is implemented in Python mainly relying on PyTorch~\citep{Ansel2024}, and GPyTorch~\citep{Gardner2018} for GP inference.
Our code is available at \href{https://github.com/chritoth/bci-arco-gp}{https://github.com/chritoth/bci-arco-gp}.

\subsection{Mechanism Model}
We follow \citet{Toth2022} and use two types of models for our mechanism.

\textbf{For root nodes}, i.e., causal variables without parents, we place a conjugate normal-inverse-gamma prior N-$\Gamma^{-1}$ on the mean and variance of that node.
Specifically, 
\begin{align*}
    p(f_i, \sigma_i^2 \given \Pa_i=\varnothing) &= \text{N-}\Gamma^{-1}(f_i, \sigma_i^2\given \mu_0, \kappa^{-1}_0, \alpha_0, \beta_0)\\
                                    &= \text{N}(f_i\given \mu_0, \sigma_i^2 \cdot\kappa^{-1}_0) \cdot \Gamma^{-1}(\sigma_i^2\given \alpha_0, \beta_0)\\
\end{align*}
where $\mu_0, \kappa^{-1}_0, \alpha_0, \beta_0$ are fixed hyper-parameters.
When sampling ground-truth SCMs we set $\mu_0 = 0$, $\kappa_0 = 1$, $\alpha_0 = 5$ and $\beta_0 = 10$, sample a mean and variance from the prior and keep them fixed thereafter.
For the inference with ARCO-GP and DIBS-GP, we use $\mu_0 = 0$, $\kappa_0 = 1$, $\alpha_0 = 10$ and $\beta_0 = 10$. This yields a prior mean of $1$ for the variance $\sigma^2$, which is sufficiently broad considering that we standardise all training data to zero mean and unit variance.
Analytic expressions for the posterior marginal likelihood are found in~\citep{Murphy2007}.

\textbf{For non-root nodes} we place a GP prior on the mechanisms. Specifically, we use rational quadratic (RQ) kernel
\begin{align*}
  k_{\text{RQ}}(\mathbf{x_1}, \mathbf{x_2}) = \delta\cdot \left(1 + \frac{1}{2\gamma}
  (\mathbf{x_1} - \mathbf{x_2})^\top \lambda^{-2} (\mathbf{x_1} - \mathbf{x_2}) \right)^{-\gamma}
\end{align*}
with scaling parameter $\delta$, lengthscale parameter $\lambda$ and mixing parameter $\gamma$ to model non-linear mechanisms.

We model the additive noise with a Gaussian likelihood with variance $\sigma_i^2$.
We place Gamma priors $\Gamma(\delta\given \alpha_\delta, \beta_\delta)$, $\Gamma(\lambda\given \alpha_\lambda, \beta_\lambda)$, $\Gamma(\gamma\given \alpha_\gamma, \beta_\gamma)$ and $\Gamma(\sigma_i^2\given \alpha_\sigma, \beta_\sigma)$ on these parameters.

When generating ground-truth models we set $\alpha_\delta = 100$, $\beta_\delta=10$, $\alpha_\lambda=30\cdot |\Pa_i|$, $\beta_\lambda=30$, $\alpha_\gamma=20$, $\beta_\gamma=10$ and $\alpha_\sigma=50$, $\beta_\sigma=50$. For each GP we sample a set of hyper-parameters from their priors and keep them fixed thereafter. Additionally, we sample a function from the GP prior with $50$ support points sampled uniform random in range [-10, 10].
For the inference with ARCO-GP and DIBS-GP, we set $\alpha_\delta = 100$, $\beta_\delta=10$, $\alpha_\lambda=30\cdot |\Pa_i|$, $\beta_\lambda=30$, $\alpha_\gamma=20$, $\beta_\gamma=10$ and $\alpha_\sigma=2$, $\beta_\sigma=8$ again considering that we standardise all training data to zero mean and unit variance.
We train GP hyper-parameters for a maximum of $100$ steps with the RMSprop~\citep{Hinton2012} optimiser with learning rate $0.05$.

Alternatively, for ablations in \cref{app:ablations} we use invertible (sigmoidal) non-linear mechanisms to generate ground-truth models, as proposed by \citet{Buhlmann2014}.
Concretely, each mechanisms has the form
\begin{align*}
    f_i(\Pa_i) = \sum_{X_j\in\Pa_i} \delta_j \cdot \frac{\gamma_j (X_j + \mu_j)}{1 + |\gamma_j (X_j + \mu_j)|} + \epsilon,
\end{align*}
i.e., the mechanisms are additive and invertible over the individual parents.
In our experiments, we sample the parameters from $\delta_j\sim \Gamma(50, 10)$, $\gamma_j\sim \Ucal([-2, -0.5] \cup [0.5,2])$, $\mu_j\sim  \Ucal([-2,2])$ and $\epsilon\sim\Ncal(0,\sigma^2)$ where $\sigma^2\sim \Gamma(50, 50)$.

\subsection{ARCO Model}
We use a normal prior $\Ncal(\thetab\given 0, \sigma^2I)$ with $\sigma = 10$ over ARCO's neural network parameters $\thetab$. The neural network $g_{\thetab}$ uses a single hidden layer with $30$ neurons and ReLU activation functions (see \cref{sec:arco}). We train ARCO for a maximum of $400$ gradient steps, using the ADAM~\citep{Kingma2014} optimiser with learning rate of $0.01$. For the score-function gradient estimator in \cref{eq:arco-grad} we use an exponential moving average baseline with decay rate $0.9$. We limit the parent set size to a maximum of two parents per node unless stated otherwise.
To estimate causal queries (\cref{eq:scm_expectation}) or ARCO's gradients (\cref{eq:arco-grad}), we sample $100$ causal orders to approximate the expectation w.r.t. $p(L\given\thetab)$ and to compute the necessary importance weights respectively (see \cref{app:estimate_iw}).

\subsection{Estimation of Importance Weights $\bm{w^L}$}\label{app:estimate_iw}
To compute the importance weights (re-stated here for convenience; defined in \cref{eq:importance-weight})
\begin{align*}
    &w^L:= \frac{\EE_{G\given L}\sbr{p(\Dcal\given\psib, G) \cdot p(\psib\given G)}}{\EE_{L'\given\thetab}\sbr{\EE_{G'\given L'}\sbr{p(\Dcal\given\psib, G') \cdot p(\psib\given G')}}}
\end{align*}
needed in the estimation of queries and gradients (in \cref{eq:scm_expectation,eq:arco-grad}), note that $\EE_{G\given L}\sbr{p(\Dcal\given\psib, G) \cdot p(\psib\given G)}$ can be computed in closed-form by restricting the maximum parent set size (see \cref{sec:psm}).
As the expectation w.r.t. $p(L'\given\thetab)$ in the denominator is intractable, we use the asymptotically unbiased ratio estimator (see e.g., \citep{Robert2004}), i.e., we normalise the weights such that they sum to 1.

\subsection{Estimation of Causal Queries}\label{app:query-estimation}
To estimate causal queries $Y = Y(\scm)$%
\footnote{We write $Y(\scm)$ to make explicit that the query is a function of the SCM.}%
, we either
\begin{enumerate}
\item estimate the query posterior $p(Y\given\Dcal) = \EE_{\scm\given\Dcal}\sbr{p(Y\given\scm)}$ directly via \cref{eq:scm_expectation},
\item estimate the predictive posterior mean
\begin{align}\label{eq:query-mean}
    \EE_{Y\given\Dcal}\sbr{Y} &= \EE_{\scm\given\Dcal}\sbr{\EE_{Y\given\scm}\sbr{Y}} \notag\\
    &=  \EE_{\thetab,\psib\given\Dcal}\sbr{\EE_{L\given\thetab} \sbr{w^L\cdot\EE_{G\given L, \psib, \Dcal}\sbr{\EE_{\fb\given \psib, \Dcal}\sbr{\EE_{Y\given\scm}\sbr{Y}}}}},
\end{align}
\item or we sample from the query posterior $p(Y\given\Dcal)$ as outlined in \cref{sec:bci-via-cos} and compute a sample-based metric or estimate an empirical distribution as in \cref{fig:main_sachsgraph_idist}.
\end{enumerate}
Below, we provide concrete examples for all three cases for better comprehensibility.

\paragraph{Example 1: Marginal Edge Probability.}
To estimate the probability that the edge $X_i \to X_j$ is present in the true graph, our query $Y(\scm) = \indicator{\scm}{{X_i \to X_j}}$ indicates whether the edge is present $Y(\scm) = 1$ or absent $Y(\scm) = 0$ in the graph of $\scm$.
Consequently,
\begin{align*}
 p(Y = 1\given \scm) = \indicator{\scm}{{X_i \to X_j}} = \begin{cases} 1 \mtext{if} X_i\to X_j \in G \\ 0 \mtext{otherwise.}\end{cases}   
\end{align*}
and, starting from \cref{eq:scm_expectation}, we get
\begin{align*}  
    p(Y\given\Dcal) &= \EE_{\scm\given\Dcal}\sbr{p(Y\given\scm)} \\
    &= \EE_{\thetab,\psib\given\Dcal}\sbr{\EE_{L\given\thetab} \sbr{w^L\cdot\EE_{G\given L, \psib, \Dcal}\sbr{\EE_{\fb\given \psib, \Dcal}\sbr{p(Y\given\scm)}}}} \\
    &= \EE_{\thetab,\psib\given\Dcal}\sbr{\EE_{L\given\thetab} \sbr{w^L\cdot\EE_{G\given L, \psib, \Dcal}\sbr{\EE_{\fb\given \psib, \Dcal}\sbr{\indicator{\scm}{X_i \to X_j}}}}} \\
    \intertext{and since the query $\indicator{\scm}{X_i \to X_j}$ does not depend on the mechanisms $\fb$}
     &= \EE_{\thetab,\psib\given\Dcal}\sbr{\EE_{L\given\thetab} \sbr{w^L\cdot\EE_{G\given L, \psib, \Dcal}\sbr{\indicator{\scm}{X_i \to X_j}}}}.
\end{align*}
As $\indicator{\scm}{{X_i \to X_j}}$ clearly factorises over parent sets, \cref{prop:ex-factorising} applies to compute the expectation w.r.t. $G$ in closed-form.
To approximate the expectation w.r.t. $L$ we compute a Monte-Carlo estimate by sampling causal orders from ARCO with importance weights $w^L$ (computed as in \cref{app:estimate_iw}), given our learned MAP parameters $\thetab, \psib$.

\paragraph{Example 2: Expected Structural Hamming Distance.}
To compute the expected structural Hamming distance w.r.t. the true graph $G^\true$, the query is $Y(\scm) = \text{SHD}(G, G^\true)$ and thus
\begin{align*}
 p(Y = y\given\scm) = \begin{cases} 1 \mtext{if}y = \text{SHD}(G, G^\true) \\ 0 \mtext{otherwise.}\end{cases}   
\end{align*}
Note that in this case, $p(Y\given\scm)$ is a point mass, as $Y(\scm) = \text{SHD}(G, G^\true)$ is fully determined by the SCM.
Therefore, 
\begin{align*}
    \EE_{Y\given\scm}\sbr{Y} = Y(\scm) = \text{SHD}(G, G^\true)
\end{align*}
and the query posterior mean in \cref{eq:query-mean} reduces to
\begin{align*}  
    \EE_{Y\given\Dcal}\sbr{Y} 
     &= \EE_{\thetab,\psib\given\Dcal}\sbr{\EE_{L\given\thetab} \sbr{w^L\cdot\EE_{G\given L, \psib, \Dcal}\sbr{\EE_{\fb\given \psib, \Dcal}\sbr{\text{SHD}(G, G^\true)}}}} \\
     \intertext{and since again the query does not depend on the mechanisms $\fb$}
     &= \EE_{\thetab,\psib\given\Dcal}\sbr{\EE_{L\given\thetab} \sbr{w^L\cdot\EE_{G\given L, \psib, \Dcal}\sbr{\text{SHD}(G, G^\true)}}}.
\end{align*}
As the SHD additively decomposes over the parent sets, we can use \cref{prop:ex-summing} to compute the inner expectation w.r.t. $G$ in closed form.
We approximate the outer expectations as described in Example 1.
The above applies analogously for our other structure learning metrics (see \cref{app:metrics}), using either \cref{prop:ex-factorising}, \cref{prop:ex-summing} or a sampling-based approximation of the expectation w.r.t. $G$, depending on the given metric (see \cref{sec:psm}).

We use a sampling-based approximation w.r.t.  as described in Example 1.

\paragraph{Example 3: Maximum Mean Discrepancy.}
When evaluating the ability to infer interventional distributions, we compute an empirical estimate of the MMD (see \citep{Gretton2012}) between an inferred posterior interventional distribution $p(\Xb\given \doint{\wb}, \Dcal)$ and the corresponding ground truth $p(\Xb\given \doint{\wb}, \scm^*)$, i.e.,
\begin{align*}
    \mmd{p(\Xb\given \doint{\wb}, \Dcal), p(\Xb\given \doint{\wb}, \scm^*)} \approx \mmd{\xb, \xb^*},
\end{align*}
where $\xb \sim p(\Xb\given \doint{\wb}, \Dcal)$ and $\xb^* \sim p(\Xb\given \doint{\wb}, \scm^*)$ are samples drawn from the respective distributions (see \cref{app:metrics}).
In this case, the causal query $Y(\scm) = \Xb \given \doint{\Wb = \wb}$ is the set of endogenous (random) variables under intervention.
Accordingly, the query posterior reads
\begin{align*}
    p(Y\given\Dcal) &= p(\Xb\given \doint{\wb}, \Dcal) \\
     &= \EE_{\thetab,\psib\given\Dcal}\sbr{\EE_{L\given\thetab} \sbr{w^L\cdot\EE_{G\given L, \psib, \Dcal}\sbr{\EE_{\fb\given \psib, \Dcal}\sbr{p(\Xb\given \doint{\wb}, \fb, \psib, \Dcal)}}}} \\
     &= \EE_{\thetab,\psib\given\Dcal}\sbr{\EE_{L\given\thetab} \sbr{w^L\cdot\EE_{G\given L, \psib, \Dcal}\sbr{p(\Xb\given \doint{\wb}, \psib, \Dcal)}}}.
\end{align*}
We can draw samples from this expression as per \cref{alg:bci}, i.e., for our learned MAP parameters $\thetab, \psib$, we sample causal orders from $p(L\given\thetab)$, then graphs given orders from $p(G\given L, \psib, \Dcal)$ and finally samples from $p(\Xb\given \doint{\wb}, \psib, \Dcal)$ by sampling from the respective GP predictive posteriors (see \cref{sec:gps}).
The samples corresponding to each sampled order $L$ are weighted with their respective importance weights $w^L$, computed as in \cref{app:estimate_iw}.

\subsection{Kernel Density Estimates of Interventional Distributions.}\label{sec:kde}
To illustrate empirical distributions in \cref{fig:main_sachsgraph_idist}, we compute a kernel density estimate (KDE) given (weighted) samples drawn from the (learned) distributions. We draw $100$ causal orders, with $10$ graphs per order and $10$ samples per graph, equalling $10000$ samples from the posterior interventional distribution in total. From the ground-truth model, we draw $10000$ samples as well. We use the \href{https://scikit-learn.org/stable/modules/generated/sklearn.neighbors.KernelDensity.html}{KDE implementation} provided by scikit-learn \citep{Pedregosa2011}, with a Gaussian kernel and a bandwidth of $0.2$.

%% file: A3_exp_setup.tex
\clearpage
\section{EXPERIMENTAL SETUP}\label{app:exp-setup}

\subsection{Sampling of Ground-Truth SCMs.}
To sample ground truth SCMs we draw graph structures from either the Erdös-Rényi (ER) \citep{Erdos1959} or the scale-free (SF) \citep{Barabasi1999} random models commonly used in DAG structure learning benchmarks.
Per default, we follow the setup of \citet{Lorch2021} and generate scale-free and Erdös-Rényi graphs with an expected degree of $2$.
Specifically, this will yield SF graphs with a maximum parent set size of $2$, whereas no such restriction applies to ER graphs in general.
For ablations in \cref{app:ablations} we also generate graphs with expected degree in $\{4, 8\}$.
We instantiate the causal mechanisms for non-linear additive noise models using GPs as described in \cref{app:details}.

For each ground truth SCM, we sample a fixed set of $N$ training data from the observational distribution.
Importantly, \citet{Reisach2021} argue, that the causal order may strongly correlate with increasing marginal variance in simulated data, and therefore, benchmarks may be easy to game by exploiting this property. As this may be especially relevant to order-based inference methods, we follow their recommendation and normalise the training data for each endogenous variable to zero mean and unit variance.

\subsection{Metrics}\label{app:metrics}
As metrics for learning the causal structure we report the \emph{expected structural Hamming distance} (ESHD), as well as the \emph{area under the receiver operating characteristic} (AUROC) and the \emph{area under the precision recall curve} (AUPRC) w.r.t. posterior edge prediction as commonly reported metrics.
To provide additional insight into the methods' strengths and weaknesses, we also report the expected number of edges (\#Edges), the \emph{true positive rate} (TPR) and the \emph{true negative rate} (TNR) for the edge prediction task. We do \textbf{not} report the log-likelihood on held-out test data as is sometimes reported, because the evaluated methods implement different noise models and approximate inference schemes, which leads to the (marginal) log-likelihoods being uncalibrated and thus incomparable (cf. \citep{Murphy2023}[Sec. 7.5] and references therein).

To evaluate the inferred causal structures w.r.t. their causal implications, \citet{Henckel2024} recently proposed a family of causal distances called \emph{Adjustment Identification Distance} (AID).
Briefly summarised, the AID counts the number of correctly identified adjustment sets for pairwise causal effects w.r.t. a target graph, where the adjustment sets are determined according to a chosen identification strategy.
We adopt the three variants proposed by \citet{Henckel2024}, namely \emph{Ancestor} adjustment (A-AID), \emph{Parent} adjustment (P-AID), and \emph{Optimal} adjustment (OSET-AID).
These metrics can also be applied between different graph types, e.g., comparing the AID between a predicted CPDAG and a reference DAG.
Note that the (P-AID) computed between two DAGs is equivalent to the well-known Structural Identification Distance proposed by \citet{Peters2015}.

To evaluate the inference of interventional distributions, we report the \emph{maximum mean discrepancy} MMD \citep{Gretton2012} between an inferred posterior interventional distribution $p(\Xb\given \doint{\wb}, \Dcal)$ and the corresponding ground truth $p(\Xb\given \doint{\wb}, \scm^*)$.
As the MMD is intractable in general, we compute an empirical estimate
\begin{align*}
    \text{MMD}^2(\xb, \xb^*) =  \sum_{i}^N\sum_{j\neq i}^N \frac{w_i w_j}{\sum_{k\neq i} w_k} k(\xb_i, \xb_j) - \frac{2}{M}\sum_{i}^N\sum_{j}^M w_i \cdot k(\xb_i, \xb_j^*) + \frac{1}{M(M-1)}\sum_{i}^M\sum_{i\neq j}^M k(\xb_i^*, \xb_j^*)
\end{align*}
where $\xb = \{\xb_n \sim p(\Xb\given \doint{\wb}, \Dcal)\}_{n=1}^N$ are samples from our learned model with associated weights $w_n$ s.t. $\sum_n w_n = 1$, and $\xb^* = \{\xb_m \sim p(\Xb\given \doint{\wb}, \scm^*)\}_{m=1}^M$ are samples from the true model. We use a Gaussian kernel
\begin{align*}
    k(\xb,\xb') = \delta \cdot \exp\rbr{-\frac{(\xb - \xb')^T(\xb - \xb')}{2\gamma}}
\end{align*}
with scaling parameter $\delta = 1000$ and lengthscale $\gamma = 0.2$.
The parameters where choosen empirically, see also \cref{sec:kde}.

Additionally, we compute the L1 and L2 error norms
\begin{align*}
     || \EE_{\Xb\given \doint{\wb}, \Dcal}\sbr{\Xb} - \EE_{\Xb\given \doint{\wb}, \scm^*}\sbr{\Xb} ||_p \mtext{for} p\in\{1,2\},
\end{align*}
 between the inferred posterior mean and the true mean of an interventional distribution.

\begin{figure}[t]
    \centering 
    \includegraphics[width=0.5\textwidth]{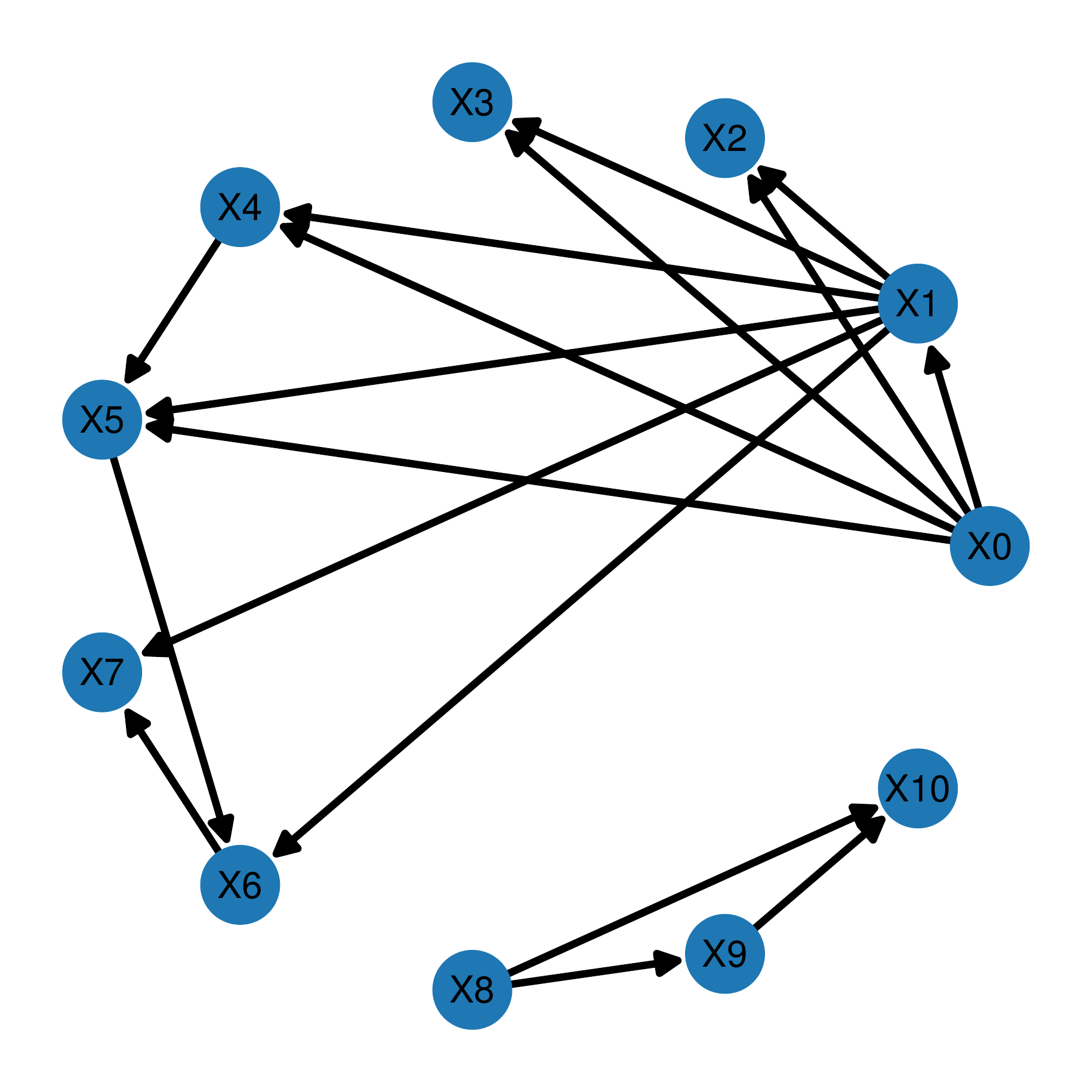}
    \caption{\textbf{Sachs Graph.}
    Consensus protein interaction graph from \citep{Sachs2005}. We relabeled nodes to avoid misinterpretation of our simulation results. Nodes X0 to X10 correspond to the original labels ['PKC', 'PKA', 'Jnk', 'P38', 'Raf', 'Mek', 'Erk', 'Akt', 'Plcg', 'PIP3', 'PIP2'].
    }
    \label{fig:sachs-graph}
\end{figure}

\subsection{Baselines.}

\paragraph{Structure learning baselines.}
In the past, a plethora of methods for \emph{causal structure learning} (a.k.a. \emph{causal discovery}) have been proposed \citep{Squires2022,Glymour2019, Vowels2022, Heinze-Deml2018}.
To evaluate our method in a wider context, we compare our inference model (ARCO-GP) to a diverse set of ten different structure learning methods from different streams of research.
We describe our baselines briefly in the following.
\begin{itemize}
    \item \textbf{BAYESDAG} \citep{Annadani2023}.
    BayesDAG utilises a mixture of MCMC to infer permutations and mechanism parameters, and Variational Inference (VI) to infer DAG edges given the permutations.
    We use the implementation provided by \citet{Annadani2023} at \href{https://github.com/microsoft/Project-BayesDAG}{https://github.com/microsoft/Project-BayesDAG}.
    We needed to adapt the sparsity regularisation hyper-parameter in order to get meaningful results and ran our experiments with the configurations in Listings~\ref{lst:bayesdag-config-sf}-\ref{lst:bayesdag-config-er}.    
    As the implementation runs a number of MCMC chains and only evaluates the best chain afterwards, we use only one MCMC chain to enable a fair comparison, as 
    multiple chains would correspond to multiple runs of the other methods.
    
    \item \textbf{DAG-GNN} \citep{Yu2019}.
    DAG-GNN is a gradient-based structure learning approach combining graph neural networks an acyclicity constraint similar to \citet{Zheng2018}.
    We use the implementation provided by \citet{Zhang2021a}[Version 1.0.3] using default settings.

    \item \textbf{DDS} \citep{Charpentier2022}.
    DDS builds upon the permutation-based approach of \cite{Cundy2021} and utilises differentiable permutation sampling and VI to infer a posterior over DAGs.
    We use the implementation provided by \citet{Charpentier2022} at \href{https://github.com/sharpenb/Differentiable-DAG-Sampling}{https://github.com/sharpenb/Differentiable-DAG-Sampling}.
    We needed to adapt the default hyper-parameters in order to get meaningful results and ran our experiments with the configuration displayed in Listing~\ref{lst:dds-config}.
    
    \item \textbf{DIBS-GP} \citep{Toth2022,Lorch2021}.
    DIBS-GP is a Bayesian causal inference framework recently proposed by \citet{Toth2022} utilising the differential structure learning method by \citet{Lorch2021} employing a soft acyclicity constraint in line with \citet{Zheng2018, Yu2019}.
    We use the implementation provided by \citet{Toth2022} at \href{https://github.com/chritoth/active-bayesian-causal-inference}{https://github.com/chritoth/active-bayesian-causal-inference}.
    We use their standard parameters with $10$ latent particles and constant hyper-parameters $\alpha = \beta = 1$ as described in \citep{Lorch2021}.
    For each latent particle we sample $100$ graphs to estimate gradients and causal quantities.
    
    \item \textbf{GADGET} \citep{Viinikka2020}.
    GADGET is a Bayesian structure learning method for linear Gaussian models based on MCMC and structure marginalisation over capped-size parent sets.
    We use the implementation provided by \citep{Viinikka2020} at \href{https://github.com/Sums-of-Products/sumu}{https://github.com/Sums-of-Products/sumu} and parameters as shown in Listing~\ref{lst:gadget-config}.
    
    \item \textbf{GES} \citep{Chickering2003}.
    GES is a well-known greedy score-based method for causal discovery using the BIC score.
    We use the implementation provided by \citet{Zhang2021a}[Version 1.0.3] using default settings.
    
    \item \textbf{GOLEM} \citep{Ng2020}.
    GOLEM is a differentiable DAG structure learning similar to \citet{Zheng2018} but with a likelihood-based score function for linear models.
    We use the implementation provided by \citet{Zhang2021a}[Version 1.0.3] using default settings.
    
    \item \textbf{GRASP} \citep{Lam2022}.
    We compare against GRASP as a state-of-the-art representative of a recent line of work of permutation-based approaches to causal discovery building upon the sparsest permutation principle \citep{Raskutti2018,Solus2021}. 
    We use the implementation provided by \citet{Zheng2024} using default settings.

    \item \textbf{RESIT} \citep{Peters2014}. RESIT is designed for causal discovery on additive noise models. It combines regression with independence tests on the regression residuals to first construct a causal order resp. a fully connected DAG, which it then uses to interatively remove edges to obtain the causal graph. As a regression model for RESIT, we use GPs with the same set of hyper-parameters and training procedure as ARCO-GP to provide better comparability.
    
    \item \textbf{PC} \citep{Spirtes2000}.
    The PC algorithm is another standard causal discovery methods based on conditional independence tests.
    We use the implementation provided by \citet{Zhang2021a}[Version 1.0.3] using default settings.
    It happens quite frequently that PC return a possibly cyclic PDAG, for which the AID metrics cannot computed and are thus omited in our experimental results.

\end{itemize}

\paragraph{BCI baselines.}
There are few prior works on BCI and baselines for non-linear additive noise models that learn a full generative model (i.e., graph, mechanisms and distribution parameters of the exogenous variables) and that provide the ability to sample from interventional distributions or estimate average causal effects are scarce (see \cref{sec:related-work}).
As this is the case for all our structure learning baselines except DIBS-GP, we need to use a different set of baselines when evaluating ARCO-GPs ability to infer interventional distributions and (average) causal effects.

We compare to 
\begin{itemize}
    \item \textbf{DIBS-GP} as a baseline operating in the same problem scenario (see description above),
    \item \textbf{RESIT-GP}, as a baseline emulating a traditional two-stage approach to causal inference: we first infer a causal DAG using RESIT (see above) and then fit GP mechanisms with the same setup as for ARCO-GP to provide better comparability, and
    \item \textbf{GADET+BEEPS} (\citep{Viinikka2020}), that we describe below.
\end{itemize}

GADGET+BEEPS is a BCI method for linear Gaussian models that samples from the posterior over causal graphs using GADGET and infers causal effects as the path-wise accumulation of the linear mechanism weights (BEEPS).
As the implementation of GADGET+BEEPS does not naturally support sampling from interventional distributions or the estimation of interventional distribution means, we extended the implementation in the following way.

A linear SCM can be written as
\begin{align*}
    \Xb = (\Gb\odot \Bb)^T\cdot \Xb + \mub + \epsilonb,
\end{align*}
where $\Xb$ are the endogenous variables, $\Gb$ is the adjacency matrix of the causal DAG, $\Bb$ is the weight matrix of the linear edge weights, $\odot$ denotes the element-wise multiplication, $\mub$ denotes the mean vector of the variables and $\epsilonb$ denotes the exogenous Gaussian noise.
Solving for $\Xb$ yields
\begin{align*}
    \Xb = (\Ib - (\Gb\odot \Bb)^T)^{-1} (\mub + \epsilonb),
\end{align*}
and taking the expectation w.r.t. $\epsilon$ yields
\begin{align}\label{eq:beeps-ace}
    \EE_\epsilon[\Xb] = (\Ib - (\Gb\odot \Bb)^T)^{-1}\mub.
\end{align}
We then estimate the interventional distribution mean $\EE_{\Xb\given do(X_j=x_j)}\sbr{\Xb}$ by (i) drawing posterior DAGs $G$ and weight matrices $\Bb$ using GADGET+BEEPS, (ii) performing the intervention by removing all parents of $X_j$ in the sampled DAGs and setting the posterior mean $\mu_j = x_j$, and (iii) finally solving \cref{eq:beeps-ace} to get the desired estimate.
We do not extend GADGET+BEEPS to be able to sample from interventional distributions.

\newcounter{codecounter}
\begin{lstlisting}[backgroundcolor=\color{lightgray},language=Python, caption=DDS Parameter Set.]
    # Architecture parameters
    'seed_model': 123,  # Seed to init model. 
    'ma_hidden_dims': [32, 32, 32],  # Output dimension. 
    'ma_architecture': 'linear',  # Output dimension. 
    'ma_fast': False,  # Output dimension. 
    'pd_initial_adj': 'Learned',  # Output dimension. 
    'pd_temperature': 1.0,  # Output dimension. 
    'pd_hard': True,  # Output dimension. 
    'pd_order_type': 'topk',  # Output dimension. 
    'pd_noise_factor': 1.0,  # Hidden dimensions.

    # Training parameters
    'max_epochs': 500,  # Maximum number of epochs for training
    'patience': 150,  # Patience for early stopping. 
    'frequency': 2,  # Frequency for early stopping test. 
    'batch_size': 16,  # Batch size. 
    'ma_lr': 1e-3,  # Learning rate. 
    'pd_lr': 1e-2,  # Learning rate. 
    'loss': 'ELBO',  # Loss name. string
    'regr': 1e-1,  # Regularization factor in Bayesian loss. 
    'prior_p': 1e-6  # Regularization factor in Bayesian loss. 
\end{lstlisting}
\refstepcounter{codecounter}\label{lst:dds-config}

\clearpage
\begin{lstlisting}[backgroundcolor=\color{lightgray},language=Python, caption=BAYESDAG Non-Linear ER Parameter Set.]
    "model_hyperparams": {
        "num_chains": 1,
        "sinkhorn_n_iter": 3000,
        "scale_noise_p": 0.001,
        "scale_noise": 0.001,
        "VI_norm": true,
        "input_perm": false,
        "lambda_sparse": 10,
        "sparse_init": false
    },
    "training_hyperparams": {
        "learning_rate": 1e-3,
        "batch_size": 512,
        "stardardize_data_mean": false,
        "stardardize_data_std": false,
        "max_epochs": 500
    }
\end{lstlisting}
\refstepcounter{codecounter}\label{lst:bayesdag-config-sf}

\begin{lstlisting}[backgroundcolor=\color{lightgray},language=Python, caption=BAYESDAG Non-linear SF Parameter Set.]
    "model_hyperparams": {
        "num_chains": 1,
        "sinkhorn_n_iter": 3000,
        "scale_noise_p": 0.001,
        "scale_noise": 0.001,
        "VI_norm": true,
        "input_perm": false,
        "lambda_sparse": 10,
        "sparse_init": false
    },
    "training_hyperparams": {
        "learning_rate": 1e-3,
        "batch_size": 512,
        "stardardize_data_mean": false,
        "stardardize_data_std": false,
        "max_epochs": 500
    }
\end{lstlisting}
\refstepcounter{codecounter}\label{lst:bayesdag-config-er}

\begin{lstlisting}[backgroundcolor=\color{lightgray},language=Python, caption=GADGET Parameter Set.]
    "scoref": 'bge',
    "max_id": -1,
    "K": min(self.num_nodes - 1, 16),
    "d": 2,
    "cp_algo": 'greedy-lite',
    "mc3_chains": 16,
    "burn_in": 1000,
    "iterations": 1000,
    "thinning": 10,
\end{lstlisting}
\refstepcounter{codecounter}\label{lst:gadget-config}

%% file: A4_ablations.tex
\clearpage
\section{EXTENDED EXPERIMENTAL RESULTS}\label{app:ablations}
In this appendix, we present additional experimental results and ablations further evaluating our ARCO-GP method.
For these extended experiments we report an extended set of eight additional structure learning metrics (described in \cref{app:exp-setup}) compared to our results in \cref{sec:experiments}.
For details on our experimental setup, e.g., data generation, baselines, etc., we refer to \cref{app:details,app:exp-setup}.

\paragraph{Real-world dataset from \citet{Sachs2005}.}
The dataset consists of 853 observational data.
The target consensus graph has 11 nodes and 17 edges.
We compare our ARCO-GP method in variants with maximum parent set size $k\in{2,3}$.
As the results in \cref{tab:sachs} show, both ARCO-GP variants are competitive with the baselines.

\paragraph{Simulations on larger models with $d=50$ variables.}
We report structure learning results for non-linear models with ER and SF graphs over $d=50$ variables in \cref{tab:ablations_50_nodes_1}.
The results of ARCO-GP w.r.t. the baselines are qualitatively similar to the ones achieved on 20 node models, i.e., we outperform all baselines for both graph types.
Especially on scale-free graphs for that our assumption on a maximum parent set size of $K=2$ holds, we achieve almost flawless graph identification, resulting in a significant performance gap to the baselines.

\paragraph{Influence of the maximum parent set size.}
We first evaluate the influence of the maximum parent set size restriction on the performance of our ARCO-GP method in our default benchmark setup (models with $20$ nodes and expected degree of $2$).
We distinguish the models variants by labels ARCO-K{k}-GP for $k\in\{1, 2, 3, 4\}$, meaning we allow a maximum parent set size of $k$.
The results in  \cref{tab:ablations_max_pssize_1} show, that $k=1$ performs worst as expected, since each node can only have one parent.
Not surprisingly, $k=2$ performs best on scale-free graphs as for these graphs no node has more than two parents.
On Erdös-Renyi graphs, $k=4$ performs best, as the these graphs can have an arbitrary number of parents, which is best reflected by the largest $k$.

In a second experiment, we keep the maximum parent set size of ARCO-GP fixed to $k\in\{2,4\}$ and we generate ground-truth models with expected degree in $\{4,8\}$.
Note again, that nodes in a scale-free graph cannot have more than $4$ parents, whereas nodes in an Erdös-Renyi graph can have an arbitrary number of parents.
The results are listed in \cref{tab:ablations_avg_deg_4,tab:ablations_avg_deg_8}.
As expected, the performance of ARCO-GP deteriorates as the true graphs get denser and our assumptions are violated.
However, we still outperform our baselines on most metrics, albeit with smaller margin.
It is interesting to observe, that ARCO-GP maintains a high true negative rate in all cases. 
We conjecture, that due to the restriction on the number of parents, ARCO-GP infers sub-graphs of the true model.

\paragraph{Invertible non-linear models.}
We test ARCO-GP on invertible non-linear models with sigmoidal mechanisms (see \cref{app:exp-setup}), for which structure identification is more challenging \citep{Buhlmann2014}.
The results are listed in \cref{tab:ablations_invertible_nl}.
Although performance decreases compared to models with non-invertible GP mechanisms, ARCO-GP still outperforms the baselines with a clear margin.

\paragraph{Influence of the number of available training data.}
We investigate the influence of the number of training data $N\in\cbr{100,200,500,1000}$ for non-linear scale-free models in \cref{tab:ablations_n_data_sf_nl_1,tab:ablations_n_data_sf_nl_2} and for non-linear Erdös-Renyi models in \cref{tab:ablations_n_data_er_nl_1,tab:ablations_n_data_er_nl_2}.
For scale-free models, ARCO-GP outperforms all baselines on all metrics in all configurations.
For Erdös-Renyi models, ARCO-GP outperforms the baselines on most metrics across the configurations or is competitive.

\begin{table*}[t]
\centering
\caption{\textbf{Real-world dataset from \citet{Sachs2005}).} 
The target graph has 11 nodes and 17 edges.
We report means and $95\%$ confidence intervals (CIs) across $20$ different ground truth models. 
Arrows next to metrics indicate lower is better ($\downarrow$) and higher is better ($\uparrow$).
}
\label{tab:sachs}
\vfill
\begin{subtable}{1.0\textwidth}
\centering
    \begin{tabular}{@{}lrccc@{}}
    \toprule
    Model & \#Edges & $\downarrow$ A-AID & $\downarrow$ P-AID & $\downarrow$ OSET-AID \\ \midrule
    ARCO-K2-GP	 & 7 $\pm$ 0	 & 0.22 $\pm$ 0.01	 & 0.48 $\pm$ 0.02	 & 0.23 $\pm$ 0.01	 \\
    ARCO-K3-GP	 & 7 $\pm$ 1	 & 0.22 $\pm$ 0.00	 & 0.47 $\pm$ 0.02	 & 0.22 $\pm$ 0.00	 \\
    BAYESDAG	 & 24 $\pm$ 1	 & 0.33 $\pm$ 0.02	 & 0.48 $\pm$ 0.02	 & 0.33 $\pm$ 0.02	 \\
    DAG-GNN	 & 7 $\pm$ 0	 & 0.23 $\pm$ 0.00	 & 0.45 $\pm$ 0.00	 & 0.23 $\pm$ 0.00	 \\
    DDS	 & 39 $\pm$ 1	 & 0.36 $\pm$ 0.01	 & 0.41 $\pm$ 0.01	 & 0.36 $\pm$ 0.01	 \\
    DIBS-GP	 & 7 $\pm$ 1	 & 0.22 $\pm$ 0.01	 & 0.43 $\pm$ 0.04	 & 0.22 $\pm$ 0.01	 \\
    GADGET	 & 9 $\pm$ 0	 & 0.25 $\pm$ 0.00	 & 0.50 $\pm$ 0.01	 & 0.25 $\pm$ 0.00	 \\
    GES	 & 8 $\pm$ 0	 & 0.28 $\pm$ 0.00	 & 0.56 $\pm$ 0.00	 & 0.28 $\pm$ 0.00	 \\
    GOLEM	 & 17 $\pm$ 0	 & 0.40 $\pm$ 0.00	 & 0.49 $\pm$ 0.00	 & 0.39 $\pm$ 0.00	 \\
    GRASP	 & 8 $\pm$ 0	 & 0.28 $\pm$ 0.00	 & 0.56 $\pm$ 0.00	 & 0.28 $\pm$ 0.00	 \\
    PC	 & 8 $\pm$ 0	 & --& --& --\\
    \bottomrule
    \end{tabular}%
    \subcaption{\textbf{AID Metrics.}}
\end{subtable}
\begin{subtable}{1.\textwidth}
\centering
    \begin{tabular}{@{}lrcccc@{}}
    \toprule
    Model & $\downarrow$ ESHD & $\uparrow$ AUROC & $\uparrow$ AUPRC & $\uparrow$ TPR & $\uparrow$ TNR \\ \midrule
    ARCO-K2-GP	 & 17 $\pm$ 1	 & 0.56 $\pm$ 0.02	 & 0.30 $\pm$ 0.02	 & 0.20 $\pm$ 0.03	 & 0.96 $\pm$ 0.01	 \\
    ARCO-K3-GP	 & 17 $\pm$ 0	 & 0.55 $\pm$ 0.03	 & 0.31 $\pm$ 0.02	 & 0.21 $\pm$ 0.02	 & 0.96 $\pm$ 0.00	 \\
    BAYESDAG	 & 34 $\pm$ 2	 & 0.49 $\pm$ 0.04	 & 0.18 $\pm$ 0.03	 & 0.22 $\pm$ 0.06	 & 0.78 $\pm$ 0.01	 \\
    DAG-GNN	 & 18 $\pm$ 0	 & 0.57 $\pm$ 0.00	 & 0.37 $\pm$ 0.00	 & 0.18 $\pm$ 0.00	 & 0.96 $\pm$ 0.00	 \\
    DDS	 & 45 $\pm$ 2	 & 0.48 $\pm$ 0.01	 & 0.19 $\pm$ 0.02	 & 0.33 $\pm$ 0.00	 & 0.64 $\pm$ 0.02	 \\
    DIBS-GP	 & 16 $\pm$ 1	 & 0.61 $\pm$ 0.02	 & 0.36 $\pm$ 0.04	 & 0.24 $\pm$ 0.04	 & 0.97 $\pm$ 0.01	 \\
    GADGET	 & 20 $\pm$ 0	 & 0.63 $\pm$ 0.03	 & 0.27 $\pm$ 0.02	 & 0.19 $\pm$ 0.01	 & 0.94 $\pm$ 0.00	 \\
    GES	 & 19 $\pm$ 0	 & 0.63 $\pm$ 0.00	 & 0.44 $\pm$ 0.00	 & 0.35 $\pm$ 0.00	 & 0.91 $\pm$ 0.00	 \\
    GOLEM	 & 26 $\pm$ 0	 & 0.55 $\pm$ 0.00	 & 0.29 $\pm$ 0.00	 & 0.24 $\pm$ 0.00	 & 0.86 $\pm$ 0.00	 \\
    GRASP	 & 19 $\pm$ 0	 & 0.63 $\pm$ 0.00	 & 0.44 $\pm$ 0.00	 & 0.35 $\pm$ 0.00	 & 0.91 $\pm$ 0.00	 \\
    PC	 & 19 $\pm$ 0	 & 0.63 $\pm$ 0.00	 & 0.44 $\pm$ 0.00	 & 0.35 $\pm$ 0.00	 & 0.91 $\pm$ 0.00	 \\
    \bottomrule
    \end{tabular}%
    \subcaption{\textbf{Edge prediction metrics.}}
\end{subtable}
\end{table*}

\begin{table*}[t]
\small
\centering
\caption{\textbf{Benchmarks on systems with 50 variables.} Ablation studies on simulated non-linear ground truth models with $50$ nodes and different DAG structures.
We report means and $95\%$ confidence intervals (CIs) across $20$ different ground truth models.
Arrows next to metrics indicate lower is better ($\downarrow$) and higher is better ($\uparrow$).
}
\label{tab:ablations_50_nodes_1}
\vfill
\begin{subtable}{1.0\textwidth}
\centering
    \begin{tabular}{@{}lrccc@{}}
    \toprule
    Model & \#Edges & $\downarrow$ A-AID & $\downarrow$ P-AID & $\downarrow$ OSET-AID  \\ \midrule
    ARCO-GP	 & 91 $\pm$ 1	 & 0.03 $\pm$ 0.01	 & 0.06 $\pm$ 0.02	 & 0.04 $\pm$ 0.02	 \\
    BAYESDAG	 & 237 $\pm$ 5	 & 0.25 $\pm$ 0.02	 & 0.57 $\pm$ 0.05	 & 0.25 $\pm$ 0.02	 \\
    DAG-GNN	 & 37 $\pm$ 4	 & 0.14 $\pm$ 0.01	 & 0.94 $\pm$ 0.02	 & 0.14 $\pm$ 0.01	 \\
    DDS	 & 1082 $\pm$ 26	 & 0.39 $\pm$ 0.01	 & 0.42 $\pm$ 0.01	 & 0.39 $\pm$ 0.01	 \\
    DIBS-GP	 & 70 $\pm$ 7	 & 0.14 $\pm$ 0.01	 & 0.82 $\pm$ 0.04	 & 0.14 $\pm$ 0.01	 \\
    GADGET	 & 176 $\pm$ 5	 & 0.31 $\pm$ 0.02	 & 0.75 $\pm$ 0.02	 & 0.31 $\pm$ 0.02	 \\
    GES	 & 220 $\pm$ 11	 & 0.38 $\pm$ 0.05	 & 0.71 $\pm$ 0.05	 & 0.39 $\pm$ 0.05	 \\
    GOLEM	 & 56 $\pm$ 5	 & 0.16 $\pm$ 0.02	 & 0.88 $\pm$ 0.03	 & 0.17 $\pm$ 0.02	 \\
    GRASP	 & 154 $\pm$ 8	 & 0.30 $\pm$ 0.02	 & 0.71 $\pm$ 0.03	 & 0.31 $\pm$ 0.02	 \\
    PC	 & 108 $\pm$ 3	 & --& --& --\\
    \bottomrule
    \end{tabular}%
    \subcaption{\textbf{Scale-free Nonlinear (Part 1/2).}}
\end{subtable}
\begin{subtable}{1.0\textwidth}
\centering
    \begin{tabular}{@{}lrcccc@{}}
    \toprule
    Model & $\downarrow$ ESHD & $\uparrow$ AUROC & $\uparrow$ AUPRC & $\uparrow$ TPR & $\uparrow$ TNR \\ \midrule
    ARCO-GP	 & 6 $\pm$ 2	 & 0.98 $\pm$ 0.01	 & 0.95 $\pm$ 0.03	 & 0.94 $\pm$ 0.02	 & 1.00 $\pm$ 0.00	 \\
    BAYESDAG	 & 220 $\pm$ 11	 & 0.77 $\pm$ 0.03	 & 0.40 $\pm$ 0.03	 & 0.59 $\pm$ 0.04	 & 0.92 $\pm$ 0.00	 \\
    DAG-GNN	 & 90 $\pm$ 4	 & 0.61 $\pm$ 0.02	 & 0.42 $\pm$ 0.03	 & 0.22 $\pm$ 0.03	 & 0.99 $\pm$ 0.00	 \\
    DDS	 & 1037 $\pm$ 25	 & 0.79 $\pm$ 0.03	 & 0.52 $\pm$ 0.05	 & 0.73 $\pm$ 0.03	 & 0.57 $\pm$ 0.01	 \\
    DIBS-GP	 & 123 $\pm$ 7	 & 0.61 $\pm$ 0.02	 & 0.27 $\pm$ 0.03	 & 0.22 $\pm$ 0.03	 & 0.98 $\pm$ 0.00	 \\
    GADGET	 & 191 $\pm$ 7	 & 0.79 $\pm$ 0.02	 & 0.30 $\pm$ 0.03	 & 0.42 $\pm$ 0.02	 & 0.94 $\pm$ 0.00	 \\
    GES	 & 229 $\pm$ 13	 & 0.69 $\pm$ 0.02	 & 0.34 $\pm$ 0.03	 & 0.46 $\pm$ 0.05	 & 0.92 $\pm$ 0.00	 \\
    GOLEM	 & 94 $\pm$ 6	 & 0.65 $\pm$ 0.02	 & 0.43 $\pm$ 0.04	 & 0.30 $\pm$ 0.03	 & 0.99 $\pm$ 0.00	 \\
    GRASP	 & 160 $\pm$ 9	 & 0.72 $\pm$ 0.01	 & 0.41 $\pm$ 0.02	 & 0.49 $\pm$ 0.02	 & 0.95 $\pm$ 0.00	 \\
    PC	 & 156 $\pm$ 5	 & 0.61 $\pm$ 0.01	 & 0.25 $\pm$ 0.02	 & 0.25 $\pm$ 0.02	 & 0.96 $\pm$ 0.00	 \\
    \bottomrule
    \end{tabular}%
    \subcaption{\textbf{Scale-free Nonlinear (Part 2/2).}}
\end{subtable}
\vfill
\begin{subtable}{1.\textwidth}
\centering
    \begin{tabular}{@{}lrccc@{}}
    \toprule
    Model & \#Edges & $\downarrow$ A-AID & $\downarrow$ P-AID & $\downarrow$ OSET-AID  \\ \midrule
    ARCO-GP	 & 48 $\pm$ 3	 & 0.09 $\pm$ 0.01	 & 0.26 $\pm$ 0.02	 & 0.09 $\pm$ 0.01	 \\
    BAYESDAG	 & 195 $\pm$ 4	 & 0.17 $\pm$ 0.03	 & 0.48 $\pm$ 0.06	 & 0.18 $\pm$ 0.03	 \\
    DAG-GNN	 & 34 $\pm$ 3	 & 0.14 $\pm$ 0.01	 & 0.57 $\pm$ 0.04	 & 0.14 $\pm$ 0.01	 \\
    DDS	 & 928 $\pm$ 30	 & 0.37 $\pm$ 0.03	 & 0.45 $\pm$ 0.03	 & 0.38 $\pm$ 0.03	 \\
    DIBS-GP	 & 33 $\pm$ 8	 & 0.12 $\pm$ 0.01	 & 0.43 $\pm$ 0.04	 & 0.13 $\pm$ 0.01	 \\
    GADGET	 & 121 $\pm$ 4	 & 0.25 $\pm$ 0.02	 & 0.50 $\pm$ 0.05	 & 0.26 $\pm$ 0.02	 \\
    GES	 & 134 $\pm$ 7	 & 0.25 $\pm$ 0.03	 & 0.45 $\pm$ 0.04	 & 0.27 $\pm$ 0.03	 \\
    GOLEM	 & 52 $\pm$ 5	 & 0.16 $\pm$ 0.02	 & 0.54 $\pm$ 0.03	 & 0.16 $\pm$ 0.01	 \\
    GRASP	 & 92 $\pm$ 5	 & 0.18 $\pm$ 0.02	 & 0.40 $\pm$ 0.05	 & 0.19 $\pm$ 0.02	 \\
    PC	 & 92 $\pm$ 3	 & --& --& --\\
    \bottomrule
    \end{tabular}%
\subcaption{\textbf{Erdös-Rényi Nonlinear (Part 1/2).}}
\end{subtable}
\vfill
\begin{subtable}{1.\textwidth}
\centering
    \begin{tabular}{@{}lrcccc@{}}
    \toprule
    Model & $\downarrow$ ESHD & $\uparrow$ AUROC & $\uparrow$ AUPRC & $\uparrow$ TPR & $\uparrow$ TNR  \\ \midrule
    ARCO-GP	 & 56 $\pm$ 4	 & 0.77 $\pm$ 0.02	 & 0.57 $\pm$ 0.04	 & 0.46 $\pm$ 0.03	 & 1.00 $\pm$ 0.00	 \\
    BAYESDAG	 & 213 $\pm$ 9	 & 0.68 $\pm$ 0.02	 & 0.29 $\pm$ 0.02	 & 0.41 $\pm$ 0.04	 & 0.93 $\pm$ 0.00	 \\
    DAG-GNN	 & 110 $\pm$ 4	 & 0.55 $\pm$ 0.01	 & 0.25 $\pm$ 0.03	 & 0.12 $\pm$ 0.01	 & 0.99 $\pm$ 0.00	 \\
    DDS	 & 900 $\pm$ 26	 & 0.75 $\pm$ 0.03	 & 0.44 $\pm$ 0.05	 & 0.64 $\pm$ 0.04	 & 0.63 $\pm$ 0.01	 \\
    DIBS-GP	 & 105 $\pm$ 5	 & 0.56 $\pm$ 0.02	 & 0.30 $\pm$ 0.04	 & 0.13 $\pm$ 0.03	 & 0.99 $\pm$ 0.00	 \\
    GADGET	 & 130 $\pm$ 4	 & 0.84 $\pm$ 0.01	 & 0.46 $\pm$ 0.03	 & 0.45 $\pm$ 0.02	 & 0.97 $\pm$ 0.00	 \\
    GES	 & 122 $\pm$ 9	 & 0.77 $\pm$ 0.02	 & 0.51 $\pm$ 0.03	 & 0.58 $\pm$ 0.03	 & 0.97 $\pm$ 0.00	 \\
    GOLEM	 & 111 $\pm$ 4	 & 0.59 $\pm$ 0.01	 & 0.31 $\pm$ 0.03	 & 0.20 $\pm$ 0.02	 & 0.99 $\pm$ 0.00	 \\
    GRASP	 & 80 $\pm$ 6	 & 0.79 $\pm$ 0.01	 & 0.61 $\pm$ 0.03	 & 0.60 $\pm$ 0.03	 & 0.98 $\pm$ 0.00	 \\
    PC	 & 112 $\pm$ 5	 & 0.69 $\pm$ 0.01	 & 0.43 $\pm$ 0.02	 & 0.40 $\pm$ 0.02	 & 0.98 $\pm$ 0.00	 \\
    \bottomrule
    \end{tabular}%
\subcaption{\textbf{Erdös-Rényi Nonlinear (Part 2/2).}}
\end{subtable}
\end{table*}

\begin{table*}[t]
\centering
\caption{\textbf{Varying the maximal parent set size.} 
Ablation studies on simulated (non-)linear ground truth models with $20$ nodes.
We report means and $95\%$ confidence intervals (CIs) across $20$ different ground truth models. 
Arrows next to metrics indicate lower is better ($\downarrow$) and higher is better ($\uparrow$).
}
\label{tab:ablations_max_pssize_1}
\vfill
\begin{subtable}{1.0\textwidth}
\centering
    \begin{tabular}{@{}lrccc@{}}
    \toprule
    Model & \#Edges & $\downarrow$ A-AID & $\downarrow$ P-AID & $\downarrow$ OSET-AID \\ \midrule
    ARCO-K1-GP	 & 10 $\pm$ 1	 & 0.17 $\pm$ 0.02	 & 0.82 $\pm$ 0.02	 & 0.19 $\pm$ 0.02	 \\
    ARCO-K2-GP	 & 32 $\pm$ 1	 & 0.06 $\pm$ 0.02	 & 0.13 $\pm$ 0.04	 & 0.07 $\pm$ 0.02	 \\
    ARCO-K3-GP	 & 35 $\pm$ 1	 & 0.09 $\pm$ 0.04	 & 0.15 $\pm$ 0.04	 & 0.10 $\pm$ 0.04	 \\
    ARCO-K4-GP	 & 37 $\pm$ 1	 & 0.08 $\pm$ 0.03	 & 0.16 $\pm$ 0.05	 & 0.10 $\pm$ 0.03	 \\
    \bottomrule
    \end{tabular}%
    \subcaption{\textbf{Scale-free Nonlinear (Part 1/2).}}
\end{subtable}
\vfill
\begin{subtable}{1.0\textwidth}
\centering
    \begin{tabular}{@{}lrcccc@{}}
    \toprule
    Model & $\downarrow$ ESHD & $\uparrow$ AUROC & $\uparrow$ AUPRC & $\uparrow$ TPR & $\uparrow$ TNR \\ \midrule
    ARCO-K1-GP	 & 26 $\pm$ 1	 & 0.86 $\pm$ 0.02	 & 0.67 $\pm$ 0.02	 & 0.28 $\pm$ 0.03	 & 1.00 $\pm$ 0.00	 \\
    ARCO-K2-GP	 & 5 $\pm$ 2	 & 0.97 $\pm$ 0.02	 & 0.94 $\pm$ 0.03	 & 0.88 $\pm$ 0.04	 & 1.00 $\pm$ 0.00	 \\
    ARCO-K3-GP	 & 9 $\pm$ 2	 & 0.94 $\pm$ 0.03	 & 0.89 $\pm$ 0.04	 & 0.87 $\pm$ 0.04	 & 0.99 $\pm$ 0.00	 \\
    ARCO-K4-GP	 & 11 $\pm$ 2	 & 0.94 $\pm$ 0.02	 & 0.90 $\pm$ 0.04	 & 0.86 $\pm$ 0.04	 & 0.98 $\pm$ 0.00	 \\
    \bottomrule
    \end{tabular}%
    \subcaption{\textbf{Scale-free Nonlinear (Part 2/2).}}
\end{subtable}
\begin{subtable}{1.\textwidth}
\centering
    \begin{tabular}{@{}lrccc@{}}
    \toprule
    Model & \#Edges & $\downarrow$ A-AID & $\downarrow$ P-AID & $\downarrow$ OSET-AID \\ \midrule
    ARCO-K1-GP	 & 8 $\pm$ 1	 & 0.21 $\pm$ 0.03	 & 0.50 $\pm$ 0.03	 & 0.21 $\pm$ 0.02	 \\
    ARCO-K2-GP	 & 20 $\pm$ 2	 & 0.15 $\pm$ 0.02	 & 0.32 $\pm$ 0.04	 & 0.18 $\pm$ 0.03	 \\
    ARCO-K3-GP	 & 29 $\pm$ 2	 & 0.12 $\pm$ 0.02	 & 0.23 $\pm$ 0.03	 & 0.16 $\pm$ 0.03	 \\
    ARCO-K4-GP	 & 32 $\pm$ 3	 & 0.10 $\pm$ 0.03	 & 0.21 $\pm$ 0.04	 & 0.15 $\pm$ 0.04	 \\
    \bottomrule
    \end{tabular}%
\subcaption{\textbf{Erdös-Rényi Nonlinear (Part 1/2).}}
\end{subtable}
\begin{subtable}{1.\textwidth}
\centering
    \begin{tabular}{@{}lrcccc@{}}
    \toprule
    Model & $\downarrow$ ESHD & $\uparrow$ AUROC & $\uparrow$ AUPRC & $\uparrow$ TPR & $\uparrow$ TNR  \\ \midrule
    ARCO-K1-GP	 & 32 $\pm$ 3	 & 0.79 $\pm$ 0.03	 & 0.54 $\pm$ 0.04	 & 0.19 $\pm$ 0.03	 & 1.00 $\pm$ 0.00	 \\
    ARCO-K2-GP	 & 21 $\pm$ 3	 & 0.84 $\pm$ 0.03	 & 0.68 $\pm$ 0.05	 & 0.49 $\pm$ 0.05	 & 1.00 $\pm$ 0.00	 \\
    ARCO-K3-GP	 & 17 $\pm$ 2	 & 0.87 $\pm$ 0.02	 & 0.76 $\pm$ 0.04	 & 0.65 $\pm$ 0.05	 & 0.99 $\pm$ 0.00	 \\
    ARCO-K4-GP	 & 17 $\pm$ 3	 & 0.88 $\pm$ 0.03	 & 0.79 $\pm$ 0.04	 & 0.70 $\pm$ 0.05	 & 0.99 $\pm$ 0.00	 \\
    \bottomrule
    \end{tabular}%
\subcaption{\textbf{Erdös-Rényi Nonlinear (Part 2/2).}}
\end{subtable}
\end{table*}

\begin{table*}[t]
\small
\centering
\caption{\textbf{Benchmarks on models with average degree 4.}
Ablation studies on simulated non-linear ground truth models with $20$ nodes scale-free and Erdös-Renyi DAG structures with average degree 4.
We report means and $95\%$ confidence intervals (CIs) across $20$ different ground truth models.
Arrows next to metrics indicate lower is better ($\downarrow$) and higher is better ($\uparrow$).
}
\label{tab:ablations_avg_deg_4}
\vfill
\begin{subtable}{1.0\textwidth}
\centering
    \begin{tabular}{@{}lrccc@{}}
    \toprule
    Model & \#Edges & $\downarrow$ A-AID & $\downarrow$ P-AID & $\downarrow$ OSET-AID  \\ \midrule
    ARCO-K2-GP	 & 24 $\pm$ 1	 & 0.30 $\pm$ 0.02	 & 0.75 $\pm$ 0.03	 & 0.34 $\pm$ 0.02	 \\
    ARCO-K4-GP	 & 53 $\pm$ 1	 & 0.23 $\pm$ 0.04	 & 0.41 $\pm$ 0.06	 & 0.34 $\pm$ 0.03	 \\
    BAYESDAG	 & 65 $\pm$ 2	 & 0.48 $\pm$ 0.02	 & 0.81 $\pm$ 0.02	 & 0.48 $\pm$ 0.02	 \\
    DAG-GNN	 & 22 $\pm$ 4	 & 0.40 $\pm$ 0.02	 & 0.92 $\pm$ 0.01	 & 0.41 $\pm$ 0.02	 \\
    DDS	 & 165 $\pm$ 5	 & 0.63 $\pm$ 0.01	 & 0.68 $\pm$ 0.01	 & 0.63 $\pm$ 0.01	 \\
    DIBS-GP	 & 38 $\pm$ 5	 & 0.36 $\pm$ 0.02	 & 0.71 $\pm$ 0.04	 & 0.39 $\pm$ 0.02	 \\
    GADGET	 & 54 $\pm$ 1	 & 0.54 $\pm$ 0.02	 & 0.86 $\pm$ 0.01	 & 0.55 $\pm$ 0.02	 \\
    GES	 & 63 $\pm$ 2	 & 0.59 $\pm$ 0.03	 & 0.85 $\pm$ 0.02	 & 0.60 $\pm$ 0.03	 \\
    GOLEM	 & 32 $\pm$ 2	 & 0.45 $\pm$ 0.02	 & 0.90 $\pm$ 0.02	 & 0.45 $\pm$ 0.02	 \\
    GRASP	 & 44 $\pm$ 1	 & 0.54 $\pm$ 0.03	 & 0.89 $\pm$ 0.02	 & 0.55 $\pm$ 0.03	 \\
    RESIT	 & 21 $\pm$ 1	 & 0.48 $\pm$ 0.03	 & 0.96 $\pm$ 0.01	 & 0.47 $\pm$ 0.03	 \\
    PC	 & 31 $\pm$ 1	 & --& --& --\\
    \bottomrule
    \end{tabular}%
    \subcaption{\textbf{Scale-free Nonlinear (Part 1/2).}}
\end{subtable}
\begin{subtable}{1.0\textwidth}
\centering
    \begin{tabular}{@{}lrcccc@{}}
    \toprule
    Model & $\downarrow$ ESHD & $\uparrow$ AUROC & $\uparrow$ AUPRC & $\uparrow$ TPR & $\uparrow$ TNR \\ \midrule
    ARCO-K2-GP	 & 47 $\pm$ 2	 & 0.76 $\pm$ 0.02	 & 0.57 $\pm$ 0.04	 & 0.32 $\pm$ 0.02	 & 0.99 $\pm$ 0.00	 \\
    ARCO-K4-GP	 & 26 $\pm$ 4	 & 0.87 $\pm$ 0.02	 & 0.80 $\pm$ 0.04	 & 0.71 $\pm$ 0.04	 & 0.98 $\pm$ 0.01	 \\
    BAYESDAG	 & 77 $\pm$ 3	 & 0.67 $\pm$ 0.02	 & 0.41 $\pm$ 0.03	 & 0.41 $\pm$ 0.03	 & 0.88 $\pm$ 0.01	 \\
    DAG-GNN	 & 62 $\pm$ 2	 & 0.58 $\pm$ 0.01	 & 0.44 $\pm$ 0.03	 & 0.18 $\pm$ 0.02	 & 0.97 $\pm$ 0.01	 \\
    DDS	 & 159 $\pm$ 5	 & 0.64 $\pm$ 0.03	 & 0.37 $\pm$ 0.04	 & 0.55 $\pm$ 0.04	 & 0.59 $\pm$ 0.01	 \\
    DIBS-GP	 & 65 $\pm$ 4	 & 0.62 $\pm$ 0.03	 & 0.40 $\pm$ 0.04	 & 0.29 $\pm$ 0.05	 & 0.94 $\pm$ 0.01	 \\
    GADGET	 & 75 $\pm$ 3	 & 0.72 $\pm$ 0.02	 & 0.43 $\pm$ 0.03	 & 0.33 $\pm$ 0.02	 & 0.90 $\pm$ 0.01	 \\
    GES	 & 86 $\pm$ 5	 & 0.60 $\pm$ 0.02	 & 0.38 $\pm$ 0.03	 & 0.33 $\pm$ 0.03	 & 0.86 $\pm$ 0.01	 \\
    GOLEM	 & 61 $\pm$ 3	 & 0.61 $\pm$ 0.01	 & 0.47 $\pm$ 0.03	 & 0.27 $\pm$ 0.03	 & 0.95 $\pm$ 0.01	 \\
    GRASP	 & 68 $\pm$ 4	 & 0.63 $\pm$ 0.01	 & 0.45 $\pm$ 0.03	 & 0.33 $\pm$ 0.02	 & 0.92 $\pm$ 0.01	 \\
    RESIT	 & 80 $\pm$ 2	 & 0.49 $\pm$ 0.01	 & 0.16 $\pm$ 0.03	 & 0.04 $\pm$ 0.01	 & 0.94 $\pm$ 0.00	 \\
    PC	 & 73 $\pm$ 3	 & 0.56 $\pm$ 0.01	 & 0.34 $\pm$ 0.03	 & 0.18 $\pm$ 0.02	 & 0.94 $\pm$ 0.01	 \\
    \bottomrule
    \end{tabular}%
    \subcaption{\textbf{Scale-free Nonlinear (Part 2/2).}}
\end{subtable}
\vfill
\begin{subtable}{1.\textwidth}
\centering
    \begin{tabular}{@{}lrccc@{}}
    \toprule
    Model & \#Edges & $\downarrow$ A-AID & $\downarrow$ P-AID & $\downarrow$ OSET-AID  \\ \midrule
    ARCO-K2-GP	 & 20 $\pm$ 2	 & 0.33 $\pm$ 0.02	 & 0.65 $\pm$ 0.04	 & 0.37 $\pm$ 0.02	 \\
    ARCO-K4-GP	 & 44 $\pm$ 2	 & 0.29 $\pm$ 0.04	 & 0.52 $\pm$ 0.05	 & 0.39 $\pm$ 0.03	 \\
    BAYESDAG	 & 55 $\pm$ 2	 & 0.44 $\pm$ 0.04	 & 0.77 $\pm$ 0.04	 & 0.46 $\pm$ 0.03	 \\
    DAG-GNN	 & 19 $\pm$ 4	 & 0.46 $\pm$ 0.03	 & 0.93 $\pm$ 0.02	 & 0.46 $\pm$ 0.03	 \\
    DDS	 & 166 $\pm$ 5	 & 0.67 $\pm$ 0.01	 & 0.72 $\pm$ 0.01	 & 0.67 $\pm$ 0.01	 \\
    DIBS-GP	 & 30 $\pm$ 5	 & 0.42 $\pm$ 0.02	 & 0.75 $\pm$ 0.03	 & 0.45 $\pm$ 0.02	 \\
    GADGET	 & 53 $\pm$ 3	 & 0.55 $\pm$ 0.02	 & 0.85 $\pm$ 0.02	 & 0.56 $\pm$ 0.02	 \\
    GES	 & 62 $\pm$ 4	 & 0.60 $\pm$ 0.04	 & 0.85 $\pm$ 0.03	 & 0.61 $\pm$ 0.04	 \\
    GOLEM	 & 29 $\pm$ 3	 & 0.52 $\pm$ 0.02	 & 0.93 $\pm$ 0.02	 & 0.51 $\pm$ 0.02	 \\
    GRASP	 & 44 $\pm$ 3	 & 0.57 $\pm$ 0.04	 & 0.87 $\pm$ 0.02	 & 0.59 $\pm$ 0.03	 \\
    RESIT	 & 20 $\pm$ 1	 & 0.51 $\pm$ 0.02	 & 0.91 $\pm$ 0.03	 & 0.52 $\pm$ 0.02	 \\
    PC	 & 32 $\pm$ 1	 & --& --& --\\
    \bottomrule
    \end{tabular}%
\subcaption{\textbf{Erdös-Rényi Nonlinear (Part 1/2).}}
\end{subtable}
\vfill
\begin{subtable}{1.\textwidth}
\centering
    \begin{tabular}{@{}lrcccc@{}}
    \toprule
    Model & $\downarrow$ ESHD & $\uparrow$ AUROC & $\uparrow$ AUPRC & $\uparrow$ TPR & $\uparrow$ TNR  \\ \midrule
    ARCO-K2-GP	 & 65 $\pm$ 4	 & 0.72 $\pm$ 0.03	 & 0.54 $\pm$ 0.03	 & 0.23 $\pm$ 0.02	 & 0.99 $\pm$ 0.00	 \\
    ARCO-K4-GP	 & 50 $\pm$ 4	 & 0.79 $\pm$ 0.03	 & 0.69 $\pm$ 0.04	 & 0.46 $\pm$ 0.04	 & 0.98 $\pm$ 0.00	 \\
    BAYESDAG	 & 82 $\pm$ 5	 & 0.65 $\pm$ 0.02	 & 0.44 $\pm$ 0.03	 & 0.34 $\pm$ 0.03	 & 0.91 $\pm$ 0.01	 \\
    DAG-GNN	 & 84 $\pm$ 4	 & 0.53 $\pm$ 0.01	 & 0.34 $\pm$ 0.04	 & 0.10 $\pm$ 0.03	 & 0.96 $\pm$ 0.01	 \\
    DDS	 & 157 $\pm$ 6	 & 0.63 $\pm$ 0.03	 & 0.39 $\pm$ 0.03	 & 0.55 $\pm$ 0.04	 & 0.60 $\pm$ 0.01	 \\
    DIBS-GP	 & 85 $\pm$ 6	 & 0.55 $\pm$ 0.02	 & 0.33 $\pm$ 0.04	 & 0.16 $\pm$ 0.03	 & 0.94 $\pm$ 0.01	 \\
    GADGET	 & 81 $\pm$ 5	 & 0.72 $\pm$ 0.02	 & 0.50 $\pm$ 0.04	 & 0.33 $\pm$ 0.03	 & 0.91 $\pm$ 0.01	 \\
    GES	 & 86 $\pm$ 6	 & 0.62 $\pm$ 0.02	 & 0.48 $\pm$ 0.03	 & 0.36 $\pm$ 0.03	 & 0.89 $\pm$ 0.01	 \\
    GOLEM	 & 84 $\pm$ 4	 & 0.55 $\pm$ 0.01	 & 0.39 $\pm$ 0.03	 & 0.16 $\pm$ 0.02	 & 0.95 $\pm$ 0.01	 \\
    GRASP	 & 79 $\pm$ 6	 & 0.62 $\pm$ 0.02	 & 0.49 $\pm$ 0.03	 & 0.31 $\pm$ 0.03	 & 0.93 $\pm$ 0.01	 \\
    RESIT	 & 97 $\pm$ 3	 & 0.48 $\pm$ 0.01	 & 0.17 $\pm$ 0.02	 & 0.03 $\pm$ 0.01	 & 0.94 $\pm$ 0.00	 \\
    PC	 & 84 $\pm$ 4	 & 0.56 $\pm$ 0.01	 & 0.41 $\pm$ 0.03	 & 0.19 $\pm$ 0.02	 & 0.94 $\pm$ 0.01	 \\
    \bottomrule
    \end{tabular}%
\subcaption{\textbf{Erdös-Rényi Nonlinear (Part 2/2).}}
\end{subtable}
\end{table*}

\begin{table*}[t]
\small
\centering
\caption{\textbf{Benchmarks on models with average degree 8.}
Ablation studies on simulated non-linear ground truth models with $20$ nodes scale-free and Erdös-Renyi DAG structures with average degree 8.
We report means and $95\%$ confidence intervals (CIs) across $20$ different ground truth models.
Arrows next to metrics indicate lower is better ($\downarrow$) and higher is better ($\uparrow$).
}
\label{tab:ablations_avg_deg_8}
\vfill
\begin{subtable}{1.0\textwidth}
\centering
    \begin{tabular}{@{}lrccc@{}}
    \toprule
    Model & \#Edges & $\downarrow$ A-AID & $\downarrow$ P-AID & $\downarrow$ OSET-AID  \\ \midrule
    ARCO-K2-GP	 & 20 $\pm$ 2	 & 0.37 $\pm$ 0.02	 & 0.61 $\pm$ 0.04	 & 0.38 $\pm$ 0.02	 \\
    ARCO-K4-GP	 & 42 $\pm$ 3	 & 0.32 $\pm$ 0.02	 & 0.58 $\pm$ 0.03	 & 0.41 $\pm$ 0.03	 \\
    BAYESDAG	 & 62 $\pm$ 3	 & 0.54 $\pm$ 0.02	 & 0.83 $\pm$ 0.02	 & 0.54 $\pm$ 0.02	 \\
    DAG-GNN	 & 21 $\pm$ 4	 & 0.43 $\pm$ 0.02	 & 0.87 $\pm$ 0.03	 & 0.43 $\pm$ 0.01	 \\
    DDS	 & 169 $\pm$ 6	 & 0.69 $\pm$ 0.01	 & 0.72 $\pm$ 0.01	 & 0.69 $\pm$ 0.01	 \\
    DIBS-GP	 & 45 $\pm$ 7	 & 0.36 $\pm$ 0.02	 & 0.55 $\pm$ 0.04	 & 0.42 $\pm$ 0.01	 \\
    GADGET	 & 61 $\pm$ 3	 & 0.62 $\pm$ 0.02	 & 0.87 $\pm$ 0.01	 & 0.62 $\pm$ 0.02	 \\
    GES	 & 71 $\pm$ 3	 & 0.65 $\pm$ 0.02	 & 0.85 $\pm$ 0.02	 & 0.65 $\pm$ 0.02	 \\
    GOLEM	 & 33 $\pm$ 3	 & 0.47 $\pm$ 0.03	 & 0.85 $\pm$ 0.03	 & 0.46 $\pm$ 0.02	 \\
    GRASP	 & 52 $\pm$ 3	 & 0.61 $\pm$ 0.02	 & 0.87 $\pm$ 0.02	 & 0.61 $\pm$ 0.02	 \\
    RESIT	 & 22 $\pm$ 1	 & 0.54 $\pm$ 0.02	 & 0.96 $\pm$ 0.01	 & 0.54 $\pm$ 0.02	 \\
    PC	 & 31 $\pm$ 2	 & --& --& --\\
    \bottomrule
    \end{tabular}%
    \subcaption{\textbf{Scale-free Nonlinear (Part 1/2).}}
\end{subtable}
\begin{subtable}{1.0\textwidth}
\centering
    \begin{tabular}{@{}lrcccc@{}}
    \toprule
    Model & $\downarrow$ ESHD & $\uparrow$ AUROC & $\uparrow$ AUPRC & $\uparrow$ TPR & $\uparrow$ TNR \\ \midrule
    ARCO-K2-GP	 & 83 $\pm$ 2	 & 0.62 $\pm$ 0.05	 & 0.47 $\pm$ 0.04	 & 0.17 $\pm$ 0.02	 & 0.99 $\pm$ 0.00	 \\
    ARCO-K4-GP	 & 78 $\pm$ 4	 & 0.72 $\pm$ 0.04	 & 0.57 $\pm$ 0.05	 & 0.31 $\pm$ 0.03	 & 0.96 $\pm$ 0.01	 \\
    BAYESDAG	 & 103 $\pm$ 4	 & 0.59 $\pm$ 0.02	 & 0.41 $\pm$ 0.02	 & 0.28 $\pm$ 0.02	 & 0.88 $\pm$ 0.01	 \\
    DAG-GNN	 & 95 $\pm$ 2	 & 0.54 $\pm$ 0.01	 & 0.42 $\pm$ 0.03	 & 0.11 $\pm$ 0.03	 & 0.96 $\pm$ 0.01	 \\
    DDS	 & 173 $\pm$ 8	 & 0.56 $\pm$ 0.03	 & 0.35 $\pm$ 0.03	 & 0.48 $\pm$ 0.04	 & 0.57 $\pm$ 0.02	 \\
    DIBS-GP	 & 87 $\pm$ 4	 & 0.64 $\pm$ 0.03	 & 0.50 $\pm$ 0.04	 & 0.28 $\pm$ 0.05	 & 0.94 $\pm$ 0.01	 \\
    GADGET	 & 111 $\pm$ 4	 & 0.62 $\pm$ 0.02	 & 0.37 $\pm$ 0.03	 & 0.24 $\pm$ 0.02	 & 0.87 $\pm$ 0.01	 \\
    GES	 & 117 $\pm$ 5	 & 0.55 $\pm$ 0.01	 & 0.41 $\pm$ 0.02	 & 0.27 $\pm$ 0.02	 & 0.84 $\pm$ 0.01	 \\
    GOLEM	 & 95 $\pm$ 2	 & 0.56 $\pm$ 0.01	 & 0.45 $\pm$ 0.02	 & 0.18 $\pm$ 0.02	 & 0.94 $\pm$ 0.01	 \\
    GRASP	 & 104 $\pm$ 4	 & 0.56 $\pm$ 0.02	 & 0.43 $\pm$ 0.03	 & 0.24 $\pm$ 0.03	 & 0.89 $\pm$ 0.01	 \\
    RESIT	 & 112 $\pm$ 2	 & 0.48 $\pm$ 0.01	 & 0.20 $\pm$ 0.02	 & 0.03 $\pm$ 0.01	 & 0.93 $\pm$ 0.00	 \\
    PC	 & 106 $\pm$ 2	 & 0.52 $\pm$ 0.01	 & 0.34 $\pm$ 0.02	 & 0.12 $\pm$ 0.01	 & 0.93 $\pm$ 0.00	 \\
    \bottomrule
    \end{tabular}%
    \subcaption{\textbf{Scale-free Nonlinear (Part 2/2).}}
\end{subtable}
\vfill
\begin{subtable}{1.\textwidth}
\centering
    \begin{tabular}{@{}lrccc@{}}
    \toprule
    Model & \#Edges & $\downarrow$ A-AID & $\downarrow$ P-AID & $\downarrow$ OSET-AID  \\ \midrule
    ARCO-K2-GP	 & 20 $\pm$ 2	 & 0.45 $\pm$ 0.02	 & 0.81 $\pm$ 0.02	 & 0.48 $\pm$ 0.02	 \\
    ARCO-K4-GP	 & 46 $\pm$ 4	 & 0.46 $\pm$ 0.03	 & 0.73 $\pm$ 0.04	 & 0.53 $\pm$ 0.03	 \\
    BAYESDAG	 & 60 $\pm$ 2	 & 0.59 $\pm$ 0.02	 & 0.87 $\pm$ 0.01	 & 0.60 $\pm$ 0.02	 \\
    DAG-GNN	 & 22 $\pm$ 4	 & 0.53 $\pm$ 0.01	 & 0.94 $\pm$ 0.02	 & 0.53 $\pm$ 0.01	 \\
    DDS	 & 49 $\pm$ 7	 & 0.58 $\pm$ 0.03	 & 0.85 $\pm$ 0.02	 & 0.60 $\pm$ 0.02	 \\
    DIBS-GP	 & 161 $\pm$ 7	 & 0.73 $\pm$ 0.01	 & 0.78 $\pm$ 0.01	 & 0.73 $\pm$ 0.00	 \\
    GADGET	 & 64 $\pm$ 3	 & 0.69 $\pm$ 0.01	 & 0.91 $\pm$ 0.01	 & 0.69 $\pm$ 0.01	 \\
    GES	 & 74 $\pm$ 2	 & 0.71 $\pm$ 0.02	 & 0.89 $\pm$ 0.01	 & 0.71 $\pm$ 0.02	 \\
    GOLEM	 & 34 $\pm$ 4	 & 0.56 $\pm$ 0.02	 & 0.93 $\pm$ 0.02	 & 0.56 $\pm$ 0.02	 \\
    GRASP	 & 54 $\pm$ 2	 & 0.70 $\pm$ 0.03	 & 0.92 $\pm$ 0.01	 & 0.70 $\pm$ 0.02	 \\
    RESIT	 & 20 $\pm$ 1	 & 0.58 $\pm$ 0.02	 & 0.97 $\pm$ 0.02	 & 0.58 $\pm$ 0.02	 \\
    PC	 & 33 $\pm$ 1	 & --& --& --\\
    \bottomrule
    \end{tabular}%
\subcaption{\textbf{Erdös-Rényi Nonlinear (Part 1/2).}}
\end{subtable}
\vfill
\begin{subtable}{1.\textwidth}
\centering
    \begin{tabular}{@{}lrcccc@{}}
    \toprule
    Model & $\downarrow$ ESHD & $\uparrow$ AUROC & $\uparrow$ AUPRC & $\uparrow$ TPR & $\uparrow$ TNR  \\ \midrule
    ARCO-K2-GP	 & 147 $\pm$ 3	 & 0.61 $\pm$ 0.04	 & 0.58 $\pm$ 0.04	 & 0.10 $\pm$ 0.01	 & 0.98 $\pm$ 0.00	 \\
    ARCO-K4-GP	 & 137 $\pm$ 5	 & 0.64 $\pm$ 0.04	 & 0.61 $\pm$ 0.04	 & 0.22 $\pm$ 0.03	 & 0.95 $\pm$ 0.01	 \\
    BAYESDAG	 & 149 $\pm$ 5	 & 0.58 $\pm$ 0.02	 & 0.54 $\pm$ 0.03	 & 0.22 $\pm$ 0.02	 & 0.89 $\pm$ 0.01	 \\
    DAG-GNN	 & 156 $\pm$ 3	 & 0.52 $\pm$ 0.01	 & 0.53 $\pm$ 0.04	 & 0.08 $\pm$ 0.02	 & 0.96 $\pm$ 0.01	 \\
    DDS	 & 166 $\pm$ 5	 & 0.47 $\pm$ 0.03	 & 0.43 $\pm$ 0.04	 & 0.14 $\pm$ 0.03	 & 0.88 $\pm$ 0.02	 \\
    DIBS-GP	 & 172 $\pm$ 11	 & 0.56 $\pm$ 0.04	 & 0.50 $\pm$ 0.04	 & 0.47 $\pm$ 0.04	 & 0.61 $\pm$ 0.03	 \\
    GADGET	 & 162 $\pm$ 4	 & 0.55 $\pm$ 0.02	 & 0.48 $\pm$ 0.02	 & 0.19 $\pm$ 0.01	 & 0.85 $\pm$ 0.01	 \\
    GES	 & 165 $\pm$ 7	 & 0.52 $\pm$ 0.02	 & 0.51 $\pm$ 0.03	 & 0.22 $\pm$ 0.02	 & 0.82 $\pm$ 0.02	 \\
    GOLEM	 & 152 $\pm$ 4	 & 0.54 $\pm$ 0.01	 & 0.56 $\pm$ 0.02	 & 0.13 $\pm$ 0.02	 & 0.94 $\pm$ 0.01	 \\
    GRASP	 & 163 $\pm$ 4	 & 0.52 $\pm$ 0.01	 & 0.50 $\pm$ 0.02	 & 0.17 $\pm$ 0.01	 & 0.87 $\pm$ 0.01	 \\
    RESIT	 & 169 $\pm$ 5	 & 0.49 $\pm$ 0.01	 & 0.37 $\pm$ 0.05	 & 0.04 $\pm$ 0.01	 & 0.94 $\pm$ 0.01	 \\
    PC	 & 164 $\pm$ 4	 & 0.50 $\pm$ 0.01	 & 0.46 $\pm$ 0.03	 & 0.09 $\pm$ 0.01	 & 0.92 $\pm$ 0.01	 \\
    \bottomrule
    \end{tabular}%
\subcaption{\textbf{Erdös-Rényi Nonlinear (Part 2/2).}}
\end{subtable}
\end{table*}

\begin{table*}[t]
\small
\centering
\caption{\textbf{Benchmarks on invertible non-linear models.}
Ablation studies on simulated non-linear ground truth models with $20$ nodes scale-free and Erdös-Renyi DAG structures with average degree 2 and invertible (sigmoidal) non-linear mechanisms.
We report means and $95\%$ confidence intervals (CIs) across $20$ different ground truth models.
Arrows next to metrics indicate lower is better ($\downarrow$) and higher is better ($\uparrow$).
}
\label{tab:ablations_invertible_nl}
\vfill
\begin{subtable}{1.0\textwidth}
\centering
    \begin{tabular}{@{}lrccc@{}}
    \toprule
    Model & \#Edges & $\downarrow$ A-AID & $\downarrow$ P-AID & $\downarrow$ OSET-AID  \\ \midrule
    ARCO-GP	 & 32 $\pm$ 1	 & 0.11 $\pm$ 0.04	 & 0.40 $\pm$ 0.07	 & 0.15 $\pm$ 0.04	 \\
    BAYESDAG	 & 64 $\pm$ 3	 & 0.38 $\pm$ 0.04	 & 0.62 $\pm$ 0.05	 & 0.38 $\pm$ 0.04	 \\
    DAG-GNN	 & 24 $\pm$ 3	 & 0.32 $\pm$ 0.03	 & 0.88 $\pm$ 0.03	 & 0.31 $\pm$ 0.03	 \\
    DDS	 & 166 $\pm$ 5	 & 0.50 $\pm$ 0.02	 & 0.54 $\pm$ 0.02	 & 0.50 $\pm$ 0.02	 \\
    DIBS-GP	 & 60 $\pm$ 5	 & 0.31 $\pm$ 0.05	 & 0.61 $\pm$ 0.06	 & 0.32 $\pm$ 0.05	 \\
    GADGET	 & 66 $\pm$ 4	 & 0.44 $\pm$ 0.03	 & 0.70 $\pm$ 0.03	 & 0.45 $\pm$ 0.03	 \\
    GES	 & 67 $\pm$ 3	 & 0.49 $\pm$ 0.05	 & 0.74 $\pm$ 0.04	 & 0.51 $\pm$ 0.05	 \\
    GOLEM	 & 41 $\pm$ 5	 & 0.39 $\pm$ 0.04	 & 0.78 $\pm$ 0.04	 & 0.41 $\pm$ 0.04	 \\
    GRASP	 & 49 $\pm$ 2	 & 0.47 $\pm$ 0.05	 & 0.77 $\pm$ 0.04	 & 0.48 $\pm$ 0.04	 \\
    RESIT	 & 30 $\pm$ 3	 & 0.36 $\pm$ 0.03	 & 0.89 $\pm$ 0.02	 & 0.33 $\pm$ 0.04	 \\
    PC	 & 28 $\pm$ 1	 & --& --& --\\ 
    \bottomrule
    \end{tabular}%
    \subcaption{\textbf{Scale-free Nonlinear (Part 1/2).}}
\end{subtable}
\begin{subtable}{1.0\textwidth}
\centering
    \begin{tabular}{@{}lrcccc@{}}
    \toprule
    Model & $\downarrow$ ESHD & $\uparrow$ AUROC & $\uparrow$ AUPRC & $\uparrow$ TPR & $\uparrow$ TNR \\ \midrule
    ARCO-GP	 & 16 $\pm$ 4	 & 0.91 $\pm$ 0.03	 & 0.79 $\pm$ 0.07	 & 0.72 $\pm$ 0.06	 & 0.98 $\pm$ 0.00	 \\
    BAYESDAG	 & 65 $\pm$ 5	 & 0.71 $\pm$ 0.04	 & 0.36 $\pm$ 0.04	 & 0.49 $\pm$ 0.04	 & 0.86 $\pm$ 0.01	 \\
    DAG-GNN	 & 45 $\pm$ 2	 & 0.58 $\pm$ 0.02	 & 0.29 $\pm$ 0.03	 & 0.20 $\pm$ 0.04	 & 0.95 $\pm$ 0.01	 \\
    DDS	 & 164 $\pm$ 5	 & 0.61 $\pm$ 0.04	 & 0.26 $\pm$ 0.05	 & 0.53 $\pm$ 0.05	 & 0.57 $\pm$ 0.01	 \\
    DIBS-GP	 & 60 $\pm$ 6	 & 0.77 $\pm$ 0.04	 & 0.42 $\pm$ 0.06	 & 0.50 $\pm$ 0.05	 & 0.88 $\pm$ 0.02	 \\
    GADGET	 & 72 $\pm$ 4	 & 0.76 $\pm$ 0.03	 & 0.30 $\pm$ 0.04	 & 0.41 $\pm$ 0.03	 & 0.85 $\pm$ 0.01	 \\
    GES	 & 77 $\pm$ 5	 & 0.61 $\pm$ 0.03	 & 0.32 $\pm$ 0.04	 & 0.38 $\pm$ 0.05	 & 0.84 $\pm$ 0.01	 \\
    GOLEM	 & 54 $\pm$ 5	 & 0.62 $\pm$ 0.03	 & 0.34 $\pm$ 0.04	 & 0.32 $\pm$ 0.05	 & 0.91 $\pm$ 0.01	 \\
    GRASP	 & 60 $\pm$ 5	 & 0.64 $\pm$ 0.03	 & 0.36 $\pm$ 0.05	 & 0.39 $\pm$ 0.06	 & 0.89 $\pm$ 0.01	 \\
    RESIT	 & 60 $\pm$ 3	 & 0.51 $\pm$ 0.01	 & 0.14 $\pm$ 0.02	 & 0.09 $\pm$ 0.02	 & 0.92 $\pm$ 0.01	 \\
    PC	 & 51 $\pm$ 2	 & 0.57 $\pm$ 0.01	 & 0.27 $\pm$ 0.02	 & 0.21 $\pm$ 0.02	 & 0.93 $\pm$ 0.00	 \\
    \bottomrule
    \end{tabular}%
    \subcaption{\textbf{Scale-free Nonlinear (Part 2/2).}}
\end{subtable}
\vfill
\begin{subtable}{1.\textwidth}
\centering
    \begin{tabular}{@{}lrccc@{}}
    \toprule
    Model & \#Edges & $\downarrow$ A-AID & $\downarrow$ P-AID & $\downarrow$ OSET-AID  \\ \midrule
    ARCO-GP	 & 27 $\pm$ 2	 & 0.16 $\pm$ 0.05	 & 0.39 $\pm$ 0.08	 & 0.18 $\pm$ 0.05	 \\
    BAYESDAG	 & 50 $\pm$ 2	 & 0.28 $\pm$ 0.04	 & 0.53 $\pm$ 0.08	 & 0.29 $\pm$ 0.05	 \\
    DAG-GNN	 & 28 $\pm$ 4	 & 0.32 $\pm$ 0.04	 & 0.71 $\pm$ 0.06	 & 0.34 $\pm$ 0.04	 \\
    DDS	 & 56 $\pm$ 6	 & 0.27 $\pm$ 0.04	 & 0.46 $\pm$ 0.07	 & 0.33 $\pm$ 0.04	 \\
    DIBS-GP	 & 158 $\pm$ 5	 & 0.46 $\pm$ 0.04	 & 0.51 $\pm$ 0.05	 & 0.46 $\pm$ 0.04	 \\
    GADGET	 & 59 $\pm$ 5	 & 0.28 $\pm$ 0.06	 & 0.46 $\pm$ 0.08	 & 0.30 $\pm$ 0.06	 \\
    GES	 & 61 $\pm$ 6	 & 0.34 $\pm$ 0.06	 & 0.49 $\pm$ 0.09	 & 0.35 $\pm$ 0.06	 \\
    GOLEM	 & 37 $\pm$ 7	 & 0.39 $\pm$ 0.05	 & 0.68 $\pm$ 0.06	 & 0.40 $\pm$ 0.04	 \\
    GRASP	 & 45 $\pm$ 4	 & 0.34 $\pm$ 0.07	 & 0.54 $\pm$ 0.09	 & 0.36 $\pm$ 0.07	 \\
    RESIT	 & 25 $\pm$ 2	 & 0.32 $\pm$ 0.04	 & 0.66 $\pm$ 0.07	 & 0.33 $\pm$ 0.04	 \\
    PC	 & 27 $\pm$ 1	 & --& --& --\\
    \bottomrule
    \end{tabular}%
\subcaption{\textbf{Erdös-Rényi Nonlinear (Part 1/2).}}
\end{subtable}
\vfill
\begin{subtable}{1.\textwidth}
\centering
    \begin{tabular}{@{}lrcccc@{}}
    \toprule
    Model & $\downarrow$ ESHD & $\uparrow$ AUROC & $\uparrow$ AUPRC & $\uparrow$ TPR & $\uparrow$ TNR  \\ \midrule
    ARCO-GP	 & 24 $\pm$ 5	 & 0.83 $\pm$ 0.04	 & 0.64 $\pm$ 0.06	 & 0.54 $\pm$ 0.05	 & 0.98 $\pm$ 0.01	 \\
    BAYESDAG	 & 56 $\pm$ 6	 & 0.71 $\pm$ 0.04	 & 0.35 $\pm$ 0.05	 & 0.43 $\pm$ 0.05	 & 0.90 $\pm$ 0.01	 \\
    DAG-GNN	 & 56 $\pm$ 4	 & 0.54 $\pm$ 0.01	 & 0.23 $\pm$ 0.02	 & 0.15 $\pm$ 0.02	 & 0.94 $\pm$ 0.01	 \\
    DDS	 & 59 $\pm$ 5	 & 0.74 $\pm$ 0.04	 & 0.41 $\pm$ 0.05	 & 0.45 $\pm$ 0.07	 & 0.89 $\pm$ 0.01	 \\
    DIBS-GP	 & 154 $\pm$ 6	 & 0.64 $\pm$ 0.05	 & 0.31 $\pm$ 0.06	 & 0.55 $\pm$ 0.06	 & 0.60 $\pm$ 0.01	 \\
    GADGET	 & 54 $\pm$ 7	 & 0.84 $\pm$ 0.04	 & 0.54 $\pm$ 0.06	 & 0.56 $\pm$ 0.06	 & 0.89 $\pm$ 0.01	 \\
    GES	 & 58 $\pm$ 9	 & 0.73 $\pm$ 0.04	 & 0.49 $\pm$ 0.05	 & 0.57 $\pm$ 0.07	 & 0.88 $\pm$ 0.02	 \\
    GOLEM	 & 59 $\pm$ 6	 & 0.57 $\pm$ 0.02	 & 0.28 $\pm$ 0.03	 & 0.22 $\pm$ 0.04	 & 0.92 $\pm$ 0.02	 \\
    GRASP	 & 47 $\pm$ 8	 & 0.73 $\pm$ 0.04	 & 0.52 $\pm$ 0.06	 & 0.54 $\pm$ 0.07	 & 0.92 $\pm$ 0.01	 \\
    RESIT	 & 61 $\pm$ 3	 & 0.49 $\pm$ 0.01	 & 0.12 $\pm$ 0.02	 & 0.06 $\pm$ 0.02	 & 0.93 $\pm$ 0.00	 \\
    PC	 & 45 $\pm$ 5	 & 0.63 $\pm$ 0.03	 & 0.40 $\pm$ 0.05	 & 0.31 $\pm$ 0.06	 & 0.95 $\pm$ 0.00	 \\
    \bottomrule
    \end{tabular}%
\subcaption{\textbf{Erdös-Rényi Nonlinear (Part 2/2).}}
\end{subtable}
\end{table*}

\begin{table*}[t]
\small
\centering
\caption{\textbf{Varying the number of data (scale-free non-linear models, table 1/2).} 
Ablation studies on simulated non-linear ground truth models with $20$ nodes.
We report means and $95\%$ confidence intervals (CIs) across $20$ different ground truth models. 
Arrows next to metrics indicate lower is better ($\downarrow$) and higher is better ($\uparrow$).
}
\label{tab:ablations_n_data_sf_nl_1}
\vfill
\begin{subtable}{1.0\textwidth}
\centering
    \begin{tabular}{@{}lrccc@{}}
    \toprule
    Model & \#Edges & $\downarrow$ A-AID & $\downarrow$ P-AID & $\downarrow$ OSET-AID \\ \midrule
    ARCO-GP	 & 32 $\pm$ 1	 & 0.06 $\pm$ 0.02	 & 0.13 $\pm$ 0.04	 & 0.07 $\pm$ 0.02	 \\
    BAYESDAG	 & 55 $\pm$ 2	 & 0.28 $\pm$ 0.02	 & 0.50 $\pm$ 0.03	 & 0.28 $\pm$ 0.02	 \\
    DAG-GNN	 & 15 $\pm$ 2	 & 0.23 $\pm$ 0.02	 & 0.87 $\pm$ 0.03	 & 0.23 $\pm$ 0.02	 \\
    DDS	 & 167 $\pm$ 4	 & 0.48 $\pm$ 0.02	 & 0.52 $\pm$ 0.02	 & 0.47 $\pm$ 0.02	 \\
    DIBS-GP	 & 26 $\pm$ 4	 & 0.23 $\pm$ 0.03	 & 0.65 $\pm$ 0.06	 & 0.23 $\pm$ 0.03	 \\
    GADGET	 & 35 $\pm$ 2	 & 0.37 $\pm$ 0.02	 & 0.82 $\pm$ 0.03	 & 0.37 $\pm$ 0.02	 \\
    GES	 & 44 $\pm$ 3	 & 0.44 $\pm$ 0.03	 & 0.78 $\pm$ 0.03	 & 0.45 $\pm$ 0.02	 \\
    GOLEM	 & 30 $\pm$ 2	 & 0.27 $\pm$ 0.03	 & 0.77 $\pm$ 0.04	 & 0.29 $\pm$ 0.03	 \\
    GRASP	 & 29 $\pm$ 2	 & 0.38 $\pm$ 0.02	 & 0.82 $\pm$ 0.04	 & 0.39 $\pm$ 0.02	 \\
    PC	 & 22 $\pm$ 1	 & --& --& --\\
    \bottomrule
    \end{tabular}%
    \subcaption{\textbf{N = 100 data.}}
\end{subtable}
\begin{subtable}{1.\textwidth}
\centering
    \begin{tabular}{@{}lrccc@{}}
    \toprule
    Model & \#Edges & $\downarrow$ A-AID & $\downarrow$ P-AID & $\downarrow$ OSET-AID  \\ \midrule
    ARCO-GP	 & 34 $\pm$ 1	 & 0.04 $\pm$ 0.03	 & 0.07 $\pm$ 0.04	 & 0.04 $\pm$ 0.03	 \\
    BAYESDAG	 & 62 $\pm$ 2	 & 0.27 $\pm$ 0.04	 & 0.39 $\pm$ 0.06	 & 0.27 $\pm$ 0.04	 \\
    DAG-GNN	 & 17 $\pm$ 4	 & 0.24 $\pm$ 0.02	 & 0.89 $\pm$ 0.03	 & 0.24 $\pm$ 0.02	 \\
    DDS	 & 169 $\pm$ 5	 & 0.46 $\pm$ 0.01	 & 0.49 $\pm$ 0.02	 & 0.46 $\pm$ 0.01	 \\
    DIBS-GP	 & 32 $\pm$ 4	 & 0.20 $\pm$ 0.04	 & 0.61 $\pm$ 0.07	 & 0.22 $\pm$ 0.03	 \\
    GADGET	 & 44 $\pm$ 2	 & 0.38 $\pm$ 0.02	 & 0.75 $\pm$ 0.03	 & 0.39 $\pm$ 0.02	 \\
    GES	 & 52 $\pm$ 3	 & 0.43 $\pm$ 0.03	 & 0.75 $\pm$ 0.03	 & 0.45 $\pm$ 0.03	 \\
    GOLEM	 & 26 $\pm$ 2	 & 0.28 $\pm$ 0.03	 & 0.85 $\pm$ 0.03	 & 0.28 $\pm$ 0.03	 \\
    GRASP	 & 37 $\pm$ 2	 & 0.41 $\pm$ 0.03	 & 0.79 $\pm$ 0.04	 & 0.43 $\pm$ 0.03	 \\
    PC	 & 29 $\pm$ 1	 & --& --& --\\
    \bottomrule
    \end{tabular}%
    \subcaption{\textbf{N = 200 data.}}
\end{subtable}
\vfill
\begin{subtable}{1.0\textwidth}
\centering
    \begin{tabular}{@{}lrccc@{}}
    \toprule
    Model & \#Edges & $\downarrow$ A-AID & $\downarrow$ P-AID & $\downarrow$ OSET-AID \\ \midrule
    ARCO-GP	 & 35 $\pm$ 1	 & 0.03 $\pm$ 0.02	 & 0.05 $\pm$ 0.03	 & 0.04 $\pm$ 0.02	 \\
    BAYESDAG	 & 77 $\pm$ 2	 & 0.33 $\pm$ 0.04	 & 0.41 $\pm$ 0.05	 & 0.33 $\pm$ 0.03	 \\
    DAG-GNN	 & 14 $\pm$ 2	 & 0.26 $\pm$ 0.03	 & 0.91 $\pm$ 0.02	 & 0.26 $\pm$ 0.03	 \\
    DDS	 & 178 $\pm$ 3	 & 0.46 $\pm$ 0.02	 & 0.48 $\pm$ 0.02	 & 0.46 $\pm$ 0.02	 \\
    DIBS-GP	 & 32 $\pm$ 3	 & 0.23 $\pm$ 0.02	 & 0.61 $\pm$ 0.05	 & 0.24 $\pm$ 0.02	 \\
    GADGET	 & 59 $\pm$ 3	 & 0.40 $\pm$ 0.03	 & 0.70 $\pm$ 0.03	 & 0.41 $\pm$ 0.03	 \\
    GES	 & 66 $\pm$ 4	 & 0.42 $\pm$ 0.04	 & 0.67 $\pm$ 0.03	 & 0.44 $\pm$ 0.03	 \\
    GOLEM	 & 23 $\pm$ 2	 & 0.28 $\pm$ 0.03	 & 0.85 $\pm$ 0.03	 & 0.29 $\pm$ 0.02	 \\
    GRASP	 & 50 $\pm$ 3	 & 0.45 $\pm$ 0.04	 & 0.73 $\pm$ 0.04	 & 0.46 $\pm$ 0.03	 \\
    PC	 & 37 $\pm$ 2	 & --& --& --\\
    \bottomrule
    \end{tabular}%
    \subcaption{\textbf{N = 500 data.}}
\end{subtable}
\vfill
\begin{subtable}{1.0\textwidth}
\centering
    \begin{tabular}{@{}lrccc@{}}
    \toprule
    Model & \#Edges & $\downarrow$ A-AID & $\downarrow$ P-AID & $\downarrow$ OSET-AID  \\ \midrule
    ARCO-GP	 & 35 $\pm$ 1	 & 0.03 $\pm$ 0.02	 & 0.05 $\pm$ 0.02	 & 0.04 $\pm$ 0.02	 \\
    BAYESDAG	 & 83 $\pm$ 4	 & 0.32 $\pm$ 0.03	 & 0.39 $\pm$ 0.05	 & 0.32 $\pm$ 0.03	 \\
    DAG-GNN	 & 18 $\pm$ 2	 & 0.26 $\pm$ 0.02	 & 0.88 $\pm$ 0.02	 & 0.26 $\pm$ 0.02	 \\
    DDS	 & 178 $\pm$ 3	 & 0.46 $\pm$ 0.02	 & 0.47 $\pm$ 0.02	 & 0.46 $\pm$ 0.02	 \\
    DIBS-GP	 & 37 $\pm$ 3	 & 0.24 $\pm$ 0.03	 & 0.59 $\pm$ 0.05	 & 0.26 $\pm$ 0.03	 \\
    GADGET	 & 73 $\pm$ 3	 & 0.41 $\pm$ 0.03	 & 0.62 $\pm$ 0.03	 & 0.41 $\pm$ 0.02	 \\
    GES	 & 81 $\pm$ 4	 & 0.43 $\pm$ 0.04	 & 0.59 $\pm$ 0.05	 & 0.44 $\pm$ 0.04	 \\
    GOLEM	 & 24 $\pm$ 2	 & 0.27 $\pm$ 0.03	 & 0.84 $\pm$ 0.03	 & 0.27 $\pm$ 0.02	 \\
    GRASP	 & 61 $\pm$ 2	 & 0.40 $\pm$ 0.04	 & 0.65 $\pm$ 0.05	 & 0.42 $\pm$ 0.03	 \\
    PC	 & 42 $\pm$ 2	 & --& --& --\\
    \bottomrule
    \end{tabular}%
    \subcaption{\textbf{N = 1000 data.}}
\end{subtable}
\end{table*}

\begin{table*}[t]
\small
\centering
\caption{\textbf{Varying the number of data (scale-free non-linear models, table 2/2).} 
Ablation studies on simulated non-linear ground truth models with $20$ nodes.
We report means and $95\%$ confidence intervals (CIs) across $20$ different ground truth models. 
Arrows next to metrics indicate lower is better ($\downarrow$) and higher is better ($\uparrow$).
}
\label{tab:ablations_n_data_sf_nl_2}
\vfill
\begin{subtable}{1.0\textwidth}
\centering
    \begin{tabular}{@{}lrcccc@{}}
    \toprule
    Model & $\downarrow$ ESHD & $\uparrow$ AUROC & $\uparrow$ AUPRC & $\uparrow$ TPR & $\uparrow$ TNR \\ \midrule
    ARCO-GP	 & 5 $\pm$ 2	 & 0.97 $\pm$ 0.02	 & 0.94 $\pm$ 0.03	 & 0.88 $\pm$ 0.04	 & 1.00 $\pm$ 0.00	 \\
    BAYESDAG	 & 46 $\pm$ 3	 & 0.81 $\pm$ 0.02	 & 0.50 $\pm$ 0.03	 & 0.61 $\pm$ 0.03	 & 0.91 $\pm$ 0.01	 \\
    DAG-GNN	 & 36 $\pm$ 2	 & 0.59 $\pm$ 0.02	 & 0.39 $\pm$ 0.05	 & 0.20 $\pm$ 0.04	 & 0.98 $\pm$ 0.01	 \\
    DDS	 & 157 $\pm$ 5	 & 0.71 $\pm$ 0.03	 & 0.37 $\pm$ 0.05	 & 0.65 $\pm$ 0.04	 & 0.58 $\pm$ 0.01	 \\
    DIBS-GP	 & 36 $\pm$ 4	 & 0.69 $\pm$ 0.03	 & 0.42 $\pm$ 0.05	 & 0.36 $\pm$ 0.05	 & 0.96 $\pm$ 0.01	 \\
    GADGET	 & 51 $\pm$ 3	 & 0.72 $\pm$ 0.02	 & 0.30 $\pm$ 0.04	 & 0.28 $\pm$ 0.03	 & 0.93 $\pm$ 0.01	 \\
    GES	 & 58 $\pm$ 4	 & 0.61 $\pm$ 0.02	 & 0.32 $\pm$ 0.03	 & 0.32 $\pm$ 0.03	 & 0.90 $\pm$ 0.01	 \\
    GOLEM	 & 38 $\pm$ 3	 & 0.67 $\pm$ 0.02	 & 0.45 $\pm$ 0.04	 & 0.38 $\pm$ 0.04	 & 0.95 $\pm$ 0.01	 \\
    GRASP	 & 45 $\pm$ 3	 & 0.63 $\pm$ 0.02	 & 0.37 $\pm$ 0.04	 & 0.32 $\pm$ 0.04	 & 0.94 $\pm$ 0.01	 \\
    PC	 & 46 $\pm$ 2	 & 0.57 $\pm$ 0.02	 & 0.27 $\pm$ 0.04	 & 0.18 $\pm$ 0.03	 & 0.95 $\pm$ 0.00	 \\
    \bottomrule
    \end{tabular}%
    \subcaption{\textbf{N = 100 data.}}
\end{subtable}
\begin{subtable}{1.\textwidth}
\centering
    \begin{tabular}{@{}lrcccc@{}}
    \toprule
    Model & $\downarrow$ ESHD & $\uparrow$ AUROC & $\uparrow$ AUPRC & $\uparrow$ TPR & $\uparrow$ TNR \\ \midrule
    ARCO-GP	 & 3 $\pm$ 2	 & 0.98 $\pm$ 0.02	 & 0.96 $\pm$ 0.03	 & 0.94 $\pm$ 0.04	 & 1.00 $\pm$ 0.00	 \\
    BAYESDAG	 & 48 $\pm$ 4	 & 0.85 $\pm$ 0.03	 & 0.52 $\pm$ 0.04	 & 0.70 $\pm$ 0.04	 & 0.89 $\pm$ 0.01	 \\
    DAG-GNN	 & 37 $\pm$ 3	 & 0.59 $\pm$ 0.03	 & 0.38 $\pm$ 0.06	 & 0.22 $\pm$ 0.05	 & 0.97 $\pm$ 0.01	 \\
    DDS	 & 158 $\pm$ 4	 & 0.73 $\pm$ 0.04	 & 0.46 $\pm$ 0.05	 & 0.66 $\pm$ 0.06	 & 0.58 $\pm$ 0.01	 \\
    DIBS-GP	 & 39 $\pm$ 4	 & 0.72 $\pm$ 0.04	 & 0.43 $\pm$ 0.05	 & 0.40 $\pm$ 0.06	 & 0.95 $\pm$ 0.01	 \\
    GADGET	 & 54 $\pm$ 3	 & 0.74 $\pm$ 0.03	 & 0.34 $\pm$ 0.04	 & 0.36 $\pm$ 0.03	 & 0.91 $\pm$ 0.01	 \\
    GES	 & 63 $\pm$ 4	 & 0.63 $\pm$ 0.02	 & 0.34 $\pm$ 0.03	 & 0.37 $\pm$ 0.04	 & 0.88 $\pm$ 0.01	 \\
    GOLEM	 & 40 $\pm$ 3	 & 0.63 $\pm$ 0.03	 & 0.39 $\pm$ 0.05	 & 0.30 $\pm$ 0.05	 & 0.96 $\pm$ 0.01	 \\
    GRASP	 & 49 $\pm$ 3	 & 0.65 $\pm$ 0.02	 & 0.38 $\pm$ 0.04	 & 0.37 $\pm$ 0.04	 & 0.92 $\pm$ 0.01	 \\
    PC	 & 50 $\pm$ 3	 & 0.57 $\pm$ 0.02	 & 0.27 $\pm$ 0.04	 & 0.21 $\pm$ 0.03	 & 0.94 $\pm$ 0.00	 \\
    \bottomrule
    \end{tabular}%
    \subcaption{\textbf{N = 200 data.}}
\end{subtable}
\vfill
\begin{subtable}{1.0\textwidth}
\centering
    \begin{tabular}{@{}lrcccc@{}}
    \toprule
    Model & $\downarrow$ ESHD & $\uparrow$ AUROC & $\uparrow$ AUPRC & $\uparrow$ TPR & $\uparrow$ TNR \\ \midrule
    ARCO-GP	 & 2 $\pm$ 1	 & 0.98 $\pm$ 0.01	 & 0.98 $\pm$ 0.01	 & 0.95 $\pm$ 0.03	 & 1.00 $\pm$ 0.00	 \\
    BAYESDAG	 & 61 $\pm$ 5	 & 0.84 $\pm$ 0.03	 & 0.45 $\pm$ 0.04	 & 0.71 $\pm$ 0.05	 & 0.85 $\pm$ 0.01	 \\
    DAG-GNN	 & 36 $\pm$ 2	 & 0.59 $\pm$ 0.02	 & 0.39 $\pm$ 0.05	 & 0.20 $\pm$ 0.04	 & 0.98 $\pm$ 0.00	 \\
    DDS	 & 164 $\pm$ 4	 & 0.76 $\pm$ 0.02	 & 0.52 $\pm$ 0.05	 & 0.70 $\pm$ 0.03	 & 0.56 $\pm$ 0.01	 \\
    DIBS-GP	 & 38 $\pm$ 3	 & 0.71 $\pm$ 0.03	 & 0.43 $\pm$ 0.04	 & 0.41 $\pm$ 0.04	 & 0.95 $\pm$ 0.01	 \\
    GADGET	 & 64 $\pm$ 5	 & 0.75 $\pm$ 0.03	 & 0.34 $\pm$ 0.05	 & 0.43 $\pm$ 0.03	 & 0.87 $\pm$ 0.01	 \\
    GES	 & 69 $\pm$ 5	 & 0.66 $\pm$ 0.02	 & 0.39 $\pm$ 0.03	 & 0.48 $\pm$ 0.03	 & 0.85 $\pm$ 0.01	 \\
    GOLEM	 & 37 $\pm$ 2	 & 0.64 $\pm$ 0.02	 & 0.44 $\pm$ 0.04	 & 0.32 $\pm$ 0.03	 & 0.97 $\pm$ 0.01	 \\
    GRASP	 & 57 $\pm$ 5	 & 0.67 $\pm$ 0.02	 & 0.41 $\pm$ 0.04	 & 0.45 $\pm$ 0.04	 & 0.89 $\pm$ 0.01	 \\
    PC	 & 54 $\pm$ 2	 & 0.60 $\pm$ 0.02	 & 0.30 $\pm$ 0.03	 & 0.27 $\pm$ 0.03	 & 0.92 $\pm$ 0.00	 \\
    \bottomrule
    \end{tabular}%
    \subcaption{\textbf{N = 500 data.}}
\end{subtable}
\vfill
\begin{subtable}{1.0\textwidth}
\centering
    \begin{tabular}{@{}lrcccc@{}}
    \toprule
    Model & $\downarrow$ ESHD & $\uparrow$ AUROC & $\uparrow$ AUPRC & $\uparrow$ TPR & $\uparrow$ TNR \\ \midrule
    ARCO-GP	 & 2 $\pm$ 1	 & 0.99 $\pm$ 0.01	 & 0.98 $\pm$ 0.01	 & 0.95 $\pm$ 0.02	 & 1.00 $\pm$ 0.00	 \\
    BAYESDAG	 & 68 $\pm$ 6	 & 0.84 $\pm$ 0.02	 & 0.43 $\pm$ 0.04	 & 0.72 $\pm$ 0.04	 & 0.83 $\pm$ 0.02	 \\
    DAG-GNN	 & 35 $\pm$ 1	 & 0.62 $\pm$ 0.01	 & 0.43 $\pm$ 0.03	 & 0.26 $\pm$ 0.03	 & 0.98 $\pm$ 0.00	 \\
    DDS	 & 163 $\pm$ 5	 & 0.76 $\pm$ 0.04	 & 0.52 $\pm$ 0.07	 & 0.71 $\pm$ 0.05	 & 0.56 $\pm$ 0.01	 \\
    DIBS-GP	 & 40 $\pm$ 3	 & 0.73 $\pm$ 0.02	 & 0.46 $\pm$ 0.03	 & 0.45 $\pm$ 0.04	 & 0.94 $\pm$ 0.01	 \\
    GADGET	 & 74 $\pm$ 4	 & 0.77 $\pm$ 0.03	 & 0.33 $\pm$ 0.04	 & 0.48 $\pm$ 0.03	 & 0.84 $\pm$ 0.01	 \\
    GES	 & 82 $\pm$ 7	 & 0.66 $\pm$ 0.04	 & 0.39 $\pm$ 0.04	 & 0.51 $\pm$ 0.06	 & 0.81 $\pm$ 0.01	 \\
    GOLEM	 & 36 $\pm$ 2	 & 0.65 $\pm$ 0.01	 & 0.46 $\pm$ 0.03	 & 0.34 $\pm$ 0.03	 & 0.97 $\pm$ 0.01	 \\
    GRASP	 & 63 $\pm$ 5	 & 0.69 $\pm$ 0.03	 & 0.42 $\pm$ 0.04	 & 0.51 $\pm$ 0.05	 & 0.87 $\pm$ 0.01	 \\
    PC	 & 59 $\pm$ 3	 & 0.59 $\pm$ 0.02	 & 0.28 $\pm$ 0.03	 & 0.27 $\pm$ 0.04	 & 0.91 $\pm$ 0.01	 \\
    \bottomrule
    \end{tabular}%
    \subcaption{\textbf{N = 1000 data.}}
\end{subtable}
\end{table*}

\begin{table*}[t]
\small
\centering
\caption{\textbf{Varying the number of data (Erdös-Renyi non-linear models, table 1/2).} 
Ablation studies on simulated non-linear ground truth models with $20$ nodes.
We report means and $95\%$ confidence intervals (CIs) across $20$ different ground truth models. 
Arrows next to metrics indicate lower is better ($\downarrow$) and higher is better ($\uparrow$).
}
\label{tab:ablations_n_data_er_nl_1}
\vfill
\begin{subtable}{1.0\textwidth}
\centering
    \begin{tabular}{@{}lrccc@{}}
    \toprule
    Model & \#Edges & $\downarrow$ A-AID & $\downarrow$ P-AID & $\downarrow$ OSET-AID \\ \midrule
    ARCO-GP	 & 20 $\pm$ 2	 & 0.15 $\pm$ 0.02	 & 0.32 $\pm$ 0.04	 & 0.18 $\pm$ 0.03	 \\
    BAYESDAG	 & 42 $\pm$ 1	 & 0.27 $\pm$ 0.04	 & 0.56 $\pm$ 0.06	 & 0.29 $\pm$ 0.04	 \\
    DAG-GNN	 & 13 $\pm$ 2	 & 0.27 $\pm$ 0.03	 & 0.71 $\pm$ 0.06	 & 0.26 $\pm$ 0.03	 \\
    DDS	 & 159 $\pm$ 7	 & 0.48 $\pm$ 0.03	 & 0.54 $\pm$ 0.04	 & 0.48 $\pm$ 0.04	 \\
    DIBS-GP	 & 18 $\pm$ 4	 & 0.23 $\pm$ 0.03	 & 0.48 $\pm$ 0.04	 & 0.25 $\pm$ 0.03	 \\
    GADGET	 & 28 $\pm$ 3	 & 0.29 $\pm$ 0.04	 & 0.62 $\pm$ 0.05	 & 0.31 $\pm$ 0.04	 \\
    GES	 & 36 $\pm$ 3	 & 0.39 $\pm$ 0.05	 & 0.67 $\pm$ 0.05	 & 0.42 $\pm$ 0.04	 \\
    GOLEM	 & 27 $\pm$ 2	 & 0.32 $\pm$ 0.04	 & 0.71 $\pm$ 0.05	 & 0.32 $\pm$ 0.04	 \\
    GRASP	 & 23 $\pm$ 2	 & 0.35 $\pm$ 0.05	 & 0.65 $\pm$ 0.07	 & 0.36 $\pm$ 0.05	 \\
    PC	 & 21 $\pm$ 1	 & --& --& --\\
    \bottomrule
    \end{tabular}%
    \subcaption{\textbf{N = 100 data.}}
\end{subtable}
\begin{subtable}{1.\textwidth}
\centering
    \begin{tabular}{@{}lrccc@{}}
    \toprule
    Model & \#Edges & $\downarrow$ A-AID & $\downarrow$ P-AID & $\downarrow$ OSET-AID  \\ \midrule
    ARCO-GP	 & 21 $\pm$ 2	 & 0.12 $\pm$ 0.02	 & 0.28 $\pm$ 0.04	 & 0.14 $\pm$ 0.02	 \\
    BAYESDAG	 & 48 $\pm$ 1	 & 0.18 $\pm$ 0.03	 & 0.38 $\pm$ 0.05	 & 0.20 $\pm$ 0.03	 \\
    DAG-GNN	 & 16 $\pm$ 3	 & 0.26 $\pm$ 0.03	 & 0.65 $\pm$ 0.05	 & 0.26 $\pm$ 0.03	 \\
    DDS	 & 159 $\pm$ 5	 & 0.43 $\pm$ 0.03	 & 0.48 $\pm$ 0.03	 & 0.43 $\pm$ 0.03	 \\
    DIBS-GP	 & 20 $\pm$ 3	 & 0.19 $\pm$ 0.02	 & 0.42 $\pm$ 0.05	 & 0.20 $\pm$ 0.02	 \\
    GADGET	 & 34 $\pm$ 2	 & 0.27 $\pm$ 0.03	 & 0.54 $\pm$ 0.05	 & 0.29 $\pm$ 0.03	 \\
    GES	 & 39 $\pm$ 3	 & 0.34 $\pm$ 0.04	 & 0.55 $\pm$ 0.05	 & 0.35 $\pm$ 0.03	 \\
    GOLEM	 & 23 $\pm$ 2	 & 0.27 $\pm$ 0.03	 & 0.62 $\pm$ 0.06	 & 0.27 $\pm$ 0.03	 \\
    GRASP	 & 28 $\pm$ 2	 & 0.31 $\pm$ 0.04	 & 0.55 $\pm$ 0.06	 & 0.32 $\pm$ 0.04	 \\
    PC	 & 26 $\pm$ 1	 & --& --& --\\
    \bottomrule
    \end{tabular}%
    \subcaption{\textbf{N = 200 data.}}
\end{subtable}
\vfill
\begin{subtable}{1.0\textwidth}
\centering
    \begin{tabular}{@{}lrccc@{}}
    \toprule
    Model & \#Edges & $\downarrow$ A-AID & $\downarrow$ P-AID & $\downarrow$ OSET-AID \\ \midrule
    ARCO-GP	 & 20 $\pm$ 2	 & 0.14 $\pm$ 0.02	 & 0.28 $\pm$ 0.03	 & 0.16 $\pm$ 0.02	 \\
    BAYESDAG	 & 61 $\pm$ 2	 & 0.22 $\pm$ 0.04	 & 0.38 $\pm$ 0.06	 & 0.24 $\pm$ 0.04	 \\
    DAG-GNN	 & 16 $\pm$ 3	 & 0.29 $\pm$ 0.03	 & 0.68 $\pm$ 0.06	 & 0.28 $\pm$ 0.03	 \\
    DDS	 & 161 $\pm$ 7	 & 0.43 $\pm$ 0.04	 & 0.47 $\pm$ 0.04	 & 0.43 $\pm$ 0.04	 \\
    DIBS-GP	 & 22 $\pm$ 5	 & 0.21 $\pm$ 0.03	 & 0.42 $\pm$ 0.07	 & 0.23 $\pm$ 0.03	 \\
    GADGET	 & 43 $\pm$ 4	 & 0.27 $\pm$ 0.04	 & 0.46 $\pm$ 0.07	 & 0.29 $\pm$ 0.04	 \\
    GES	 & 48 $\pm$ 5	 & 0.33 $\pm$ 0.05	 & 0.50 $\pm$ 0.07	 & 0.35 $\pm$ 0.05	 \\
    GOLEM	 & 21 $\pm$ 2	 & 0.30 $\pm$ 0.04	 & 0.65 $\pm$ 0.06	 & 0.30 $\pm$ 0.04	 \\
    GRASP	 & 36 $\pm$ 4	 & 0.32 $\pm$ 0.05	 & 0.50 $\pm$ 0.07	 & 0.33 $\pm$ 0.05	 \\
    PC	 & 32 $\pm$ 2	 & --& --& --\\
    \bottomrule
    \end{tabular}%
    \subcaption{\textbf{N = 500 data.}}
\end{subtable}
\vfill
\begin{subtable}{1.0\textwidth}
\centering
    \begin{tabular}{@{}lrccc@{}}
    \toprule
    Model & \#Edges & $\downarrow$ A-AID & $\downarrow$ P-AID & $\downarrow$ OSET-AID  \\ \midrule
    ARCO-GP	 & 22 $\pm$ 2	 & 0.13 $\pm$ 0.03	 & 0.31 $\pm$ 0.04	 & 0.16 $\pm$ 0.03	 \\
    BAYESDAG	 & 65 $\pm$ 2	 & 0.25 $\pm$ 0.05	 & 0.42 $\pm$ 0.06	 & 0.27 $\pm$ 0.05	 \\
    DAG-GNN	 & 18 $\pm$ 2	 & 0.29 $\pm$ 0.02	 & 0.73 $\pm$ 0.04	 & 0.29 $\pm$ 0.03	 \\
    DDS	 & 164 $\pm$ 5	 & 0.45 $\pm$ 0.03	 & 0.49 $\pm$ 0.03	 & 0.45 $\pm$ 0.03	 \\
    DIBS-GP	 & 22 $\pm$ 3	 & 0.22 $\pm$ 0.03	 & 0.47 $\pm$ 0.05	 & 0.25 $\pm$ 0.03	 \\
    GADGET	 & 53 $\pm$ 4	 & 0.30 $\pm$ 0.05	 & 0.50 $\pm$ 0.06	 & 0.32 $\pm$ 0.04	 \\
    GES	 & 58 $\pm$ 5	 & 0.36 $\pm$ 0.05	 & 0.54 $\pm$ 0.06	 & 0.38 $\pm$ 0.05	 \\
    GOLEM	 & 22 $\pm$ 2	 & 0.30 $\pm$ 0.03	 & 0.70 $\pm$ 0.03	 & 0.30 $\pm$ 0.03	 \\
    GRASP	 & 44 $\pm$ 3	 & 0.35 $\pm$ 0.04	 & 0.55 $\pm$ 0.05	 & 0.36 $\pm$ 0.04	 \\
    PC	 & 38 $\pm$ 2	 & --& --& --\\
    \bottomrule
    \end{tabular}%
    \subcaption{\textbf{N = 1000 data.}}
\end{subtable}
\end{table*}

\begin{table*}[t]
\small
\centering
\caption{\textbf{Varying the number of data (Erdös-Renyi non-linear models, table 2/2).} 
Ablation studies on simulated non-linear ground truth models with $20$ nodes.
We report means and $95\%$ confidence intervals (CIs) across $20$ different ground truth models. 
Arrows next to metrics indicate lower is better ($\downarrow$) and higher is better ($\uparrow$).
}
\label{tab:ablations_n_data_er_nl_2}
\vfill
\begin{subtable}{1.0\textwidth}
\centering
    \begin{tabular}{@{}lrcccc@{}}
    \toprule
    Model & $\downarrow$ ESHD & $\uparrow$ AUROC & $\uparrow$ AUPRC & $\uparrow$ TPR & $\uparrow$ TNR \\ \midrule
    ARCO-GP	 & 21 $\pm$ 3	 & 0.84 $\pm$ 0.03	 & 0.68 $\pm$ 0.05	 & 0.49 $\pm$ 0.05	 & 1.00 $\pm$ 0.00	 \\
    BAYESDAG	 & 51 $\pm$ 3	 & 0.71 $\pm$ 0.03	 & 0.34 $\pm$ 0.04	 & 0.39 $\pm$ 0.03	 & 0.92 $\pm$ 0.01	 \\
    DAG-GNN	 & 45 $\pm$ 3	 & 0.54 $\pm$ 0.01	 & 0.25 $\pm$ 0.03	 & 0.10 $\pm$ 0.02	 & 0.97 $\pm$ 0.01	 \\
    DDS	 & 157 $\pm$ 7	 & 0.63 $\pm$ 0.04	 & 0.29 $\pm$ 0.05	 & 0.53 $\pm$ 0.06	 & 0.59 $\pm$ 0.02	 \\
    DIBS-GP	 & 37 $\pm$ 5	 & 0.64 $\pm$ 0.03	 & 0.38 $\pm$ 0.05	 & 0.26 $\pm$ 0.05	 & 0.98 $\pm$ 0.01	 \\
    GADGET	 & 43 $\pm$ 4	 & 0.76 $\pm$ 0.02	 & 0.43 $\pm$ 0.05	 & 0.31 $\pm$ 0.03	 & 0.95 $\pm$ 0.01	 \\
    GES	 & 47 $\pm$ 5	 & 0.66 $\pm$ 0.03	 & 0.43 $\pm$ 0.05	 & 0.39 $\pm$ 0.05	 & 0.93 $\pm$ 0.01	 \\
    GOLEM	 & 48 $\pm$ 3	 & 0.59 $\pm$ 0.01	 & 0.32 $\pm$ 0.02	 & 0.23 $\pm$ 0.02	 & 0.95 $\pm$ 0.01	 \\
    GRASP	 & 40 $\pm$ 4	 & 0.64 $\pm$ 0.03	 & 0.44 $\pm$ 0.05	 & 0.33 $\pm$ 0.05	 & 0.96 $\pm$ 0.01	 \\
    PC	 & 42 $\pm$ 4	 & 0.61 $\pm$ 0.02	 & 0.39 $\pm$ 0.04	 & 0.25 $\pm$ 0.03	 & 0.96 $\pm$ 0.00	 \\
    \bottomrule
    \end{tabular}%
    \subcaption{\textbf{N = 100 data.}}
\end{subtable}
\begin{subtable}{1.\textwidth}
\centering
    \begin{tabular}{@{}lrcccc@{}}
    \toprule
    Model & $\downarrow$ ESHD & $\uparrow$ AUROC & $\uparrow$ AUPRC & $\uparrow$ TPR & $\uparrow$ TNR \\ \midrule
    ARCO-GP	 & 20 $\pm$ 2	 & 0.87 $\pm$ 0.03	 & 0.71 $\pm$ 0.05	 & 0.51 $\pm$ 0.04	 & 1.00 $\pm$ 0.00	 \\
    BAYESDAG	 & 46 $\pm$ 3	 & 0.76 $\pm$ 0.02	 & 0.45 $\pm$ 0.04	 & 0.52 $\pm$ 0.04	 & 0.92 $\pm$ 0.01	 \\
    DAG-GNN	 & 44 $\pm$ 3	 & 0.56 $\pm$ 0.01	 & 0.30 $\pm$ 0.04	 & 0.14 $\pm$ 0.02	 & 0.97 $\pm$ 0.01	 \\
    DDS	 & 152 $\pm$ 6	 & 0.68 $\pm$ 0.04	 & 0.40 $\pm$ 0.05	 & 0.59 $\pm$ 0.05	 & 0.60 $\pm$ 0.01	 \\
    DIBS-GP	 & 36 $\pm$ 3	 & 0.65 $\pm$ 0.02	 & 0.39 $\pm$ 0.03	 & 0.30 $\pm$ 0.04	 & 0.97 $\pm$ 0.01	 \\
    GADGET	 & 41 $\pm$ 3	 & 0.80 $\pm$ 0.02	 & 0.49 $\pm$ 0.06	 & 0.40 $\pm$ 0.04	 & 0.95 $\pm$ 0.01	 \\
    GES	 & 45 $\pm$ 4	 & 0.69 $\pm$ 0.02	 & 0.47 $\pm$ 0.03	 & 0.44 $\pm$ 0.03	 & 0.93 $\pm$ 0.01	 \\
    GOLEM	 & 44 $\pm$ 3	 & 0.59 $\pm$ 0.01	 & 0.35 $\pm$ 0.03	 & 0.23 $\pm$ 0.03	 & 0.96 $\pm$ 0.01	 \\
    GRASP	 & 37 $\pm$ 3	 & 0.69 $\pm$ 0.02	 & 0.51 $\pm$ 0.04	 & 0.43 $\pm$ 0.04	 & 0.96 $\pm$ 0.01	 \\
    PC	 & 41 $\pm$ 3	 & 0.65 $\pm$ 0.02	 & 0.43 $\pm$ 0.03	 & 0.33 $\pm$ 0.03	 & 0.96 $\pm$ 0.00	 \\
    \bottomrule
    \end{tabular}%
    \subcaption{\textbf{N = 200 data.}}
\end{subtable}
\vfill
\begin{subtable}{1.0\textwidth}
\centering
    \begin{tabular}{@{}lrcccc@{}}
    \toprule
    Model & $\downarrow$ ESHD & $\uparrow$ AUROC & $\uparrow$ AUPRC & $\uparrow$ TPR & $\uparrow$ TNR \\ \midrule
    ARCO-GP	 & 20 $\pm$ 2	 & 0.80 $\pm$ 0.03	 & 0.67 $\pm$ 0.05	 & 0.50 $\pm$ 0.04	 & 1.00 $\pm$ 0.00	 \\
    BAYESDAG	 & 50 $\pm$ 4	 & 0.80 $\pm$ 0.03	 & 0.48 $\pm$ 0.04	 & 0.64 $\pm$ 0.04	 & 0.89 $\pm$ 0.01	 \\
    DAG-GNN	 & 43 $\pm$ 3	 & 0.56 $\pm$ 0.02	 & 0.29 $\pm$ 0.05	 & 0.15 $\pm$ 0.04	 & 0.97 $\pm$ 0.01	 \\
    DDS	 & 150 $\pm$ 6	 & 0.73 $\pm$ 0.03	 & 0.50 $\pm$ 0.05	 & 0.65 $\pm$ 0.04	 & 0.60 $\pm$ 0.02	 \\
    DIBS-GP	 & 36 $\pm$ 5	 & 0.67 $\pm$ 0.03	 & 0.41 $\pm$ 0.06	 & 0.31 $\pm$ 0.06	 & 0.97 $\pm$ 0.01	 \\
    GADGET	 & 40 $\pm$ 4	 & 0.84 $\pm$ 0.03	 & 0.60 $\pm$ 0.06	 & 0.53 $\pm$ 0.05	 & 0.94 $\pm$ 0.01	 \\
    GES	 & 46 $\pm$ 7	 & 0.73 $\pm$ 0.03	 & 0.51 $\pm$ 0.05	 & 0.55 $\pm$ 0.06	 & 0.92 $\pm$ 0.01	 \\
    GOLEM	 & 43 $\pm$ 4	 & 0.59 $\pm$ 0.02	 & 0.34 $\pm$ 0.05	 & 0.22 $\pm$ 0.04	 & 0.96 $\pm$ 0.00	 \\
    GRASP	 & 37 $\pm$ 6	 & 0.74 $\pm$ 0.03	 & 0.56 $\pm$ 0.05	 & 0.54 $\pm$ 0.06	 & 0.94 $\pm$ 0.01	 \\
    PC	 & 40 $\pm$ 5	 & 0.68 $\pm$ 0.03	 & 0.48 $\pm$ 0.05	 & 0.41 $\pm$ 0.05	 & 0.95 $\pm$ 0.01	 \\
    \bottomrule
    \end{tabular}%
    \subcaption{\textbf{N = 500 data.}}
\end{subtable}
\vfill
\begin{subtable}{1.0\textwidth}
\centering
    \begin{tabular}{@{}lrcccc@{}}
    \toprule
    Model & $\downarrow$ ESHD & $\uparrow$ AUROC & $\uparrow$ AUPRC & $\uparrow$ TPR & $\uparrow$ TNR \\ \midrule
    ARCO-GP	 & 21 $\pm$ 3	 & 0.78 $\pm$ 0.03	 & 0.68 $\pm$ 0.05	 & 0.52 $\pm$ 0.06	 & 1.00 $\pm$ 0.00	 \\
    BAYESDAG	 & 51 $\pm$ 4	 & 0.85 $\pm$ 0.02	 & 0.51 $\pm$ 0.05	 & 0.68 $\pm$ 0.04	 & 0.89 $\pm$ 0.01	 \\
    DAG-GNN	 & 45 $\pm$ 3	 & 0.57 $\pm$ 0.02	 & 0.32 $\pm$ 0.05	 & 0.17 $\pm$ 0.04	 & 0.97 $\pm$ 0.00	 \\
    DDS	 & 150 $\pm$ 4	 & 0.75 $\pm$ 0.02	 & 0.52 $\pm$ 0.04	 & 0.68 $\pm$ 0.03	 & 0.60 $\pm$ 0.01	 \\
    DIBS-GP	 & 41 $\pm$ 5	 & 0.63 $\pm$ 0.03	 & 0.39 $\pm$ 0.06	 & 0.28 $\pm$ 0.05	 & 0.97 $\pm$ 0.01	 \\
    GADGET	 & 47 $\pm$ 5	 & 0.86 $\pm$ 0.03	 & 0.59 $\pm$ 0.05	 & 0.57 $\pm$ 0.04	 & 0.91 $\pm$ 0.01	 \\
    GES	 & 54 $\pm$ 8	 & 0.73 $\pm$ 0.03	 & 0.51 $\pm$ 0.04	 & 0.57 $\pm$ 0.04	 & 0.89 $\pm$ 0.02	 \\
    GOLEM	 & 44 $\pm$ 3	 & 0.59 $\pm$ 0.01	 & 0.37 $\pm$ 0.03	 & 0.22 $\pm$ 0.03	 & 0.96 $\pm$ 0.00	 \\
    GRASP	 & 42 $\pm$ 4	 & 0.75 $\pm$ 0.02	 & 0.55 $\pm$ 0.04	 & 0.56 $\pm$ 0.04	 & 0.93 $\pm$ 0.01	 \\
    PC	 & 47 $\pm$ 4	 & 0.66 $\pm$ 0.02	 & 0.44 $\pm$ 0.04	 & 0.39 $\pm$ 0.04	 & 0.94 $\pm$ 0.01	 \\
    \bottomrule
    \end{tabular}%
    \subcaption{\textbf{N = 1000 data.}}
\end{subtable}
\end{table*}

%% file: A5_proofs.tex
\FloatBarrier
\clearpage
\section{PROOFS AND DERIVATIONS}\label{app:proofs}

\subsection{Derivation of the Posterior Expectation w.r.t. SCMs}\label{app:scm-expectation}
This derivation is similar in spirit to the formulation in \citep{Lorch2021,Toth2022}, albeit for different generative models.
In the following, we derive the expectation w.r.t. SCMs in \cref{eq:scm_expectation} and the corresponding importance weights in \cref{eq:importance-weight}.
We start with the expectation w.r.t. SCMs $\scm = (G,\psib,\fb)$ that we parametrise as causal graph $G$, GP hyper-parameters $\psib$ and mechanisms $\fb$
\begin{align*}
    \EE_{\scm\given\Dcal}\sbr{p(Y\given\scm)} &= \EE_{G,\fb,\psib\given\Dcal}\sbr{p(Y\given\scm)} \\
    \intertext{and by marginalising over/exploiting conditional independences of our generative model in \cref{fig:scm}}
    &=\EE_{\thetab,\psib\given\Dcal}\sbr{ \EE_{G, \fb\given\thetab,\psib,\Dcal}\sbr{p(Y\given\scm)}} \\
    &=\EE_{\thetab,\psib\given\Dcal}\sbr{ \EE_{G\given\thetab,\psib,\Dcal}\sbr{\EE_{\fb\given\psib,\Dcal}\sbr{p(Y\given\scm)}}} \\
    &= \EE_{\thetab,\psib\given\Dcal}\sbr{\EE_{L\given\thetab,\psib,\Dcal}\sbr{\EE_{G\given L,\psib,\Dcal}\sbr{\EE_{\fb\given \psib, \Dcal}\sbr{p(Y\given\scm)}}}}\\
    \intertext{and by rewriting $p(L\given\thetab,\psib, \Dcal)$ using Bayes' law}
    &= \EE_{\thetab,\psib\given\Dcal}\sbr{\EE_{L\given\thetab}\sbr{\frac{p(\Dcal,\psib\given L)}{p(\Dcal,\psib\given \thetab)} \cdot \EE_{G\given L,\psib,\Dcal}\sbr{\EE_{\fb\given \psib, \Dcal}\sbr{p(Y\given\scm)}}}}\\
    &= \EE_{\thetab,\psib\given\Dcal}\sbr{\EE_{L\given\thetab}\sbr{w^L \cdot \EE_{G\given L,\psib,\Dcal}\sbr{\EE_{\fb\given \psib, \Dcal}\sbr{p(Y\given\scm)}}}}\\
\end{align*}
with 
\begin{align*}
    w^L &:= \frac{p(\Dcal,\psib\given L)}{p(\Dcal,\psib\given \thetab)} \\
    &= \frac{\EE_{G\given L}\sbr{p(\Dcal,\psib\given G)}}{p(\Dcal,\psib\given \thetab)} \\
    &= \frac{\EE_{G\given L}\sbr{p(\Dcal\given\psib, G)\cdot p(\psib\given G)}}{p(\Dcal,\psib\given \thetab)} \\
    &= \frac{\EE_{G\given L}\sbr{p(\Dcal\given\psib, G)\cdot p(\psib\given G)}}{\EE_{L'\given\thetab}\sbr{p(\Dcal,\psib\given L')}} \\
    &= \frac{\EE_{G\given L}\sbr{p(\Dcal\given\psib, G)\cdot p(\psib\given G)}}{\EE_{L'\given\thetab}\sbr{\EE_{G'\given L'}\sbr{p(\Dcal\given \psib, G') \cdot p(\psib\given G')}}}.
\end{align*}

\subsection{Derivation of the Gradient Estimators}\label{app:grads}
In the following, we derive the gradient estimators in \cref{eq:posterior_grad,eq:arco-grad,eq:psi-grad}. We denote by $\nabla = \nabla_{\thetab,  \psib}$ to avoid clutter.

\textbf{General Posterior Gradient.}
The general posterior gradient in \cref{eq:posterior_grad} reads as follows.
\begin{align*}
    \nabla \log p(\thetab,  \psib \given \Dcal) &= \nabla \log \frac{p(\Dcal, \psib \given \thetab) \cdot p(\thetab)}{p(\Dcal)} \\
    &= \nabla \log p(\thetab) + \nabla\log p(\Dcal,\psib \given \thetab)  \\
    &= \nabla \log p(\thetab) + \nabla\log\EE_{L\given\thetab}\sbr{ p(\Dcal,\psib \given L)}  \\
    &= \nabla \log p(\thetab) + \nabla\log\EE_{L\given\thetab}\sbr{\EE_{G\given L}\sbr{ p(\Dcal,\psib \given G)}} \\
    &= \nabla \log p(\thetab) + \nabla\log\EE_{L\given\thetab}\sbr{\EE_{G\given L}\sbr{ p(\Dcal\given\psib ,G) \cdot p(\psib \given G)}} \\
\end{align*}

\textbf{ARCO Gradient.}
Using the above as starting point for the gradient in \cref{eq:arco-grad}, we get
\begin{align*}
    \nabla_{\thetab} \log p(\thetab,  \psib \given \Dcal) &= \nabla_{\thetab} \log p(\thetab) + \nabla_{\thetab} \log\EE_{L\given\thetab}\sbr{\EE_{G\given L}\sbr{ p(\Dcal\given\psib ,G) \cdot p(\psib \given G)}} \\
    &= \nabla_{\thetab} \log p(\thetab) + \frac{\nabla_{\thetab} \EE_{L\given\thetab}\sbr{\EE_{G\given L}\sbr{ p(\Dcal\given\psib ,G) \cdot p(\psib \given G)}}}{\EE_{L'\given\thetab}\sbr{\EE_{G'\given L'}\sbr{p(\Dcal\given \psib, G') \cdot p(\psib \given G')}}} \\
    \intertext{and using $\nabla_{\thetab} p(L\given \thetab) = p(L\given \thetab)\cdot \nabla_{\thetab} \log p(L\given \thetab)$}
    &= \nabla_{\thetab} \log p(\thetab) + \frac{\EE_{L\given\thetab}\sbr{\EE_{G\given L}\sbr{ p(\Dcal\given\psib ,G) \cdot p(\psib \given G)} \cdot \nabla_{\thetab} \log p(L\given \thetab)}}{\EE_{L'\given\thetab}\sbr{\EE_{G'\given L'}\sbr{p(\Dcal\given \psib, G') \cdot p(\psib \given G')}}} \\
    &= \nabla_{\thetab} \log p(\thetab) + \EE_{L\given\thetab}\sbr{w^L \cdot \nabla_{\thetab}\log p(L \given \thetab)}
\end{align*}
with $w^L$ as defined in \cref{eq:importance-weight}.

\textbf{GP Hyper-parameter Gradient.}
For the gradient of a distinct GP modeling the mechanism from parents $\Pa_k$ to target node $X_k$ with corresponding hyper-parameters $\psib_k$ as in \cref{eq:psi-grad} we have
\begin{align*}
    \nabla_{\psib_k} \log p(\thetab,  \psib \given \Dcal) &= \nabla_{\psib_k} \log p(\thetab) + \nabla_{\psib_k} \log\EE_{L\given\thetab}\sbr{\EE_{G\given L}\sbr{p(\Dcal,\psib\given G)}} \\
    &= \frac{\EE_{L\given\thetab}\sbr{\EE_{G\given L}\sbr{ \nabla_{\psib_k}p(\Dcal,\psib\given G)}}}{\EE_{L\given\thetab}\sbr{\EE_{G\given L}\sbr{p(\Dcal,\psib\given G)}}}\\
    &= \frac{\EE_{L\given\thetab}\sbr{\EE_{G\given L}\sbr{ \nabla_{\psib_k}p(\Dcal,\psib\given G)}}}{p(\Dcal,\psib\given \thetab)}\\
    \intertext{and as the marginal likelihood and the prior over GP hyper-parameters factorise over parent sets we further get}
    &= \frac{\EE_{L\given\thetab}\sbr{\EE_{G\given L}\sbr{ \nabla_{\psib_k}\prod_{i=1}^d p(\Dcal_i, \psib_i \given \Pa_i^G)}}}{p(\Dcal,\psib\given \thetab)}\\
    &= \frac{\EE_{L\given\thetab}\sbr{\sum_G p(G\given L)\cdot \nabla_{\psib_k}\prod_{i=1}^d p(\Dcal_i, \psib_i \given \Pa_i^G)}}{p(\Dcal,\psib\given \thetab)}.\\
    \intertext{Now, note that for the summation over graphs $G$, the gradient is zero for all graphs that do not contain the parent set $\Pa_k$ corresponding to the GP with hyper-parameters $\psi_k$. Consequently, we get}
    &= \frac{\EE_{L\given\thetab}\sbr{\sum_{G\given \Pa_k\in G} p(G\given L)\cdot \nabla_{\psib_k}\prod_{i=1}^d p(\Dcal_i, \psib_i \given \Pa_i^G)}}{p(\Dcal,\psib\given \thetab)}\\
    \intertext{and since $\nabla_{\psib_k} p(\Dcal_k,\psib_k\given \Pa_k) =  p(\Dcal_k,\psib_k\given \Pa_k) \cdot \nabla_{\psib_k} \log p(\Dcal_k,\psib_k\given \Pa_k)$}
    &= \frac{\EE_{L\given\thetab}\sbr{\sum_{G\given \Pa_k\in G} p(G\given L)\cdot \prod_{i=1}^d p(\Dcal_i, \psib_i \given \Pa_i^G) \cdot \nabla_{\psib_k} \log p(\Dcal_k,\psib_k\given \Pa_k)}}{p(\Dcal,\psib\given \thetab)}\\
    &= \frac{\EE_{L\given\thetab}\sbr{\sum_{G\given \Pa_k\in G} p(G\given L)\cdot p(\Dcal, \psib \given G)}}{p(\Dcal,\psib\given \thetab)} \cdot \nabla_{\psib_k} \log p(\Dcal_k,\psib_k\given \Pa_k).\\
\end{align*}

Note that the term preceding the gradient $\nabla_{\psib_k} \log p(\Dcal_k,\psib_k\given \Pa_k)$ is a scalar factor that \emph{does not} influence the \emph{direction} of the gradient and can thus be practically omitted for gradient-based optimisation, as optimisation algorithms will scale the gradient depending on tune-able step size parameters anyways. Therefore, optimising the GP hyper-parameters w.r.t. the gradient in \cref{eq:posterior_grad} practically yields the same gradient direction as the common MAP type II gradient $\nabla_{\psib_k} \log p(\Dcal_k,\psib_k\given \Pa_k)$.
Arguably, for mechanism models that do not decompose over individual mechanisms for each parent set and target variable, naively estimating the gradient in \cref{eq:posterior_grad} may yield very noisy gradients, as the magnitude of the estimated gradient will depend on the sampled structures used for its estimation.

\clearpage
\subsection{Proofs regarding Exhaustive Parent Set Enumeration}\label{app:proposition-proofs}

\paragraph{Proof of \cref{prop:ex-factorising}}
\begin{proof}
    \begin{align*}
    \EE_{G\given L} \sbr{w(G) \cdot Y(G)} &=  \sum_G p(G\given L)\cdot w(G)\cdot Y(G) \\
    \intertext{Since we assume that $w(G)$ and $Y(G)$ factorise over the parent sets, we have}\\
    &=  \sum_G \prod_i p(\Pa_i^G\given L) \cdot w_i(\Pa_i^G) \cdot Y_i(\Pa_i^G) \\
    \intertext{The sum over all graphs can be represented as sum over all combinations of possible parent sets to get}\\
    &= \sum_{\Pa_1}\sum_{\Pa_2}\dots \sum_{\Pa_d} \prod_i p(\Pa_i\given L) \cdot w_i(\Pa_i) \cdot Y_i(\Pa_i) \\
    &= \sum_{\Pa_1} p(\Pa_1\given L) \cdot w_1(\Pa_1) \cdot Y_1(\Pa_1) \sum_{\Pa_2} p(\Pa_2\given L) \cdot w_2(\Pa_2) \cdot Y_2(\Pa_2) \dots\\
    \intertext{Since each summation over parent sets is independent of the others, we get the final result}\\
    &= \prod_i  \sum_{\Pa_i} p(\Pa_i\given L) \cdot w_i(\Pa_i) \cdot Y_i(\Pa_i)
\end{align*}
\end{proof}

\paragraph{Proof of \cref{prop:ex-summing}}
\begin{proof}
    \begin{align*}
    \EE_{G\given L} &\sbr{w(G) \cdot Y(G)} =  \sum_G p(G\given L)\cdot w(G)\cdot Y(G) \\
    \intertext{Since we assume that $w(G)$ factorises and $Y(G)$ sums over the parent sets, we have}\\
    &=  \sum_G \prod_i p(\Pa_i^G\given L) \cdot w_i(\Pa_i^G) \cdot \sum_j Y_j(\pa^G_j) \\
    \intertext{The sum over all graphs can be represented as sum over all combinations of possible parent sets to get}\\
    &= \sum_{\Pa_1}\sum_{\Pa_2}\dots \sum_{\Pa_d} \prod_i p(\Pa_i\given L) \cdot w_i(\Pa_i) \cdot \sum_j Y_j(\Pa_j) \\
    &= \sum_{\Pa_1}\sum_{\Pa_2}\dots \sum_{\Pa_d} \cdot \sum_j Y_j(\Pa_j) \prod_i p(\Pa_i\given L) \cdot w_i(\Pa_i) \\
    &= \sum_{\Pa_1}\sum_{\Pa_2}\dots \sum_{\Pa_d} Y_1(\Pa_1) \prod_i p(\Pa_i\given L) \cdot w_i(\Pa_i) +  \sum_{\Pa_1}\sum_{\Pa_2}\dots \sum_{\Pa_d}\sum_{j=2}^d \dots \\
    &= \sum_{\Pa_1} Y_1(\Pa_1)\cdot p(\Pa_1\given L) \cdot w_1(\Pa_1) \sum_{\Pa_2} p(\Pa_2\given L) \cdot w_2(\Pa_2) \dots +  \sum_{\Pa_1}\sum_{\Pa_2}\dots \sum_{\Pa_d}\sum_{j=2}^d \dots \\
    \intertext{By abbreviating $\alpha_i(L) = \sum_{\Pa_i} p(\Pa_i\given L) w_i(\Pa_i)$ we get}\\
    &= \sum_{\Pa_1} Y_1(\Pa_1)\cdot p(\Pa_1\given L) \cdot w_1(\Pa_1)\cdot \prod_{k\neq 1} \alpha_k(L) +  \sum_{\Pa_1}\sum_{\Pa_2}\dots \sum_{\Pa_d}\sum_{j=2}^d Y_j(\Pa_j) \prod_i p(\Pa_i\given L) \cdot w_i(\Pa_i) \\    
    \intertext{By repeating this procedure for the remaining summands $j$, we get the final result}\\
    &=\sum_i \rbr{\prod_{k\neq i} \alpha_k(L)}\cdot \sum_{\Pa_i} p(\Pa_i\given L)\cdot w_i(\Pa_i)\cdot Y_i(\Pa_i)
\end{align*}
\end{proof}